\documentclass[twoside,11pt]{article}

\usepackage{blindtext}

\usepackage{jmlr2e}

\usepackage[utf8]{inputenc} 

\usepackage{enumitem}

\usepackage[dvipsnames]{xcolor}
\usepackage{pifont}

\usepackage{wrapfig} \usepackage{tikz}
\usepackage{subcaption}

\usepackage{mathtools} \usepackage{amssymb} \usepackage{mathrsfs}
\usepackage{thmtools}
\usepackage{aligned-overset}

 \hypersetup{hypertexnames=false}

\makeatletter

\newcommand*{\dims}{N}
\newcommand*{\dimV}{d}
\newcommand*{\bigO}{\mathcal{O}}

\newcommand*{\ind}{\mathbf{1}}

\newcommand*{\cmark}{\ding{51}}\newcommand*{\xmark}{\ding{55}}

\newcommand*{\inter}{ \mathchoice{\mskip1.5mu{:}\mskip1.5mu}{\mskip1.5mu{:}\mskip1.5mu}{{:}}{{:}}
}

\newcommand*{\im}{i} \newcommand*{\Ft}{\mathcal{F}} \newcommand*{\jet}[1][k]{\mathrm{J}^{#1}}

\newcommand*{\transpose}{T}
\newcommand{\identity}{\mathbb{I}}
\newcommand{\basis}{b}
\newcommand{\Basis}{B}
\DeclareMathOperator{\Span}{span}

\newcommand{\E}{\mathbb{E}}
\renewcommand{\Pr}{\mathbb{P}}
\newcommand*{\Var}{\mathsf{Var}}
\newcommand*{\Cov}{\mathsf{Cov}}
\newcommand*{\normal}{\mathcal{N}}
\newcommand*{\filt}{\mathcal{F}}

\DeclareMathOperator{\iid}{iid}
\newcommand*{\density}{\varphi}
\newcommand*{\charfct}{\psi}

\DeclareMathOperator*{\support}{supp}

\newcommand*{\rv}{\mathbf{v}}
\newcommand*{\rw}{\mathbf{w}}

\newcommand*{\as}{a.s\@ifnextchar.{}{.\ }}

\newcommand*{\real}{\mathbb{R}}

\newcommand*{\nat}{\mathbb{N}}
\newcommand*{\integer}{\mathbb{Z}}

\newcommand*{\info}{I}
\newcommand*{\varinfo}{J}
\newcommand*{\liminfo}{\mathscr{I}}
\newcommand*{\Info}{\mathbf{I}}
\newcommand*{\VarInfo}{\hat{\mathbf{I}}}
\newcommand*{\VarLiminfo}{\hat{\mathscr{I}}}
\newcommand*{\gradients}{G}

\newcommand*{\gsa}{\mathfrak{G}}
\newcommand*{\param}{x}
\newcommand*{\Param}{X}
\newcommand*{\distance}{\Delta}

\newcommand*{\lr}{h}

\newcommand*{\timestep}{n}

\newcommand*{\C}{\mathcal{C}}
\newcommand*{\rf}{\mathbf{f}}
\newcommand*{\rg}{\mathbf{g}}
\newcommand*{\ikernel}{C}
\newcommand*{\kernel}{\kappa}
\newcommand*{\rcov}{\mathbf{\Sigma}}
\newcommand*{\rmean}{\mathbf{\mu}}

\newcommand*{\spectMeasure}{\sigma}
\newcommand*{\schoenbergMeas}{\nu}

\newcommand*{\dataMat}{\mathbf{A}}
\newcommand*{\signal}{\mathbf{\tilde{x}}}
\newcommand*{\noise}{\eta}

\newcommand*{\limf}{\mathfrak{f}}
\newcommand*{\limgdot}{\mathfrak{g}}
\newcommand*{\radius}{\lambda}

\newcommand*{\black}[1]{{\color{black} #1}}
\newcommand*{\red}[1]{{\color{red} #1}}
\newcommand*{\blue}[1]{{\color{cyan} #1}}
\newcommand*{\magenta}[1]{{\color{magenta} #1}}
\newcommand*{\green}[1]{{\color{LimeGreen} #1}}

\makeatother 
\newtheorem{assumption}[theorem]{Assumption}

\usepackage{lastpage}
\jmlrheading{27}{2026}{1-\pageref{LastPage}}{7/25; Revised 2/26}{4/26}{25-1651}{Felix Benning and Leif Döring}

\ShortHeadings{Predictable Progress}{Benning and Döring}
\firstpageno{1}

\begin{document}

\title{Gradient Span Algorithms Make Predictable Progress in High Dimension
}

\author{\name Felix Benning\thanks{Corresponding author, current affiliation: University of Luxembourg} \email felix.benning@gmail.com \\
  \name Leif Döring \email doering@uni-mannheim.de \\
  \addr Department of Mathematics\\
  University of Mannheim\\
  68159 Mannheim, Germany
}

\editor{Bryon Aragam}

\maketitle

\begin{abstract}We prove that all `gradient span algorithms' have asymptotically
deterministic behavior on scaled Gaussian random functions as the dimension tends to
infinity.
This is a functional generalization of similar results for
random quadratic functions and spin glasses. They explain
the counterintuitive phenomenon that different training
runs of many large machine learning models result in approximately equal
cost curves despite random initialization on a complicated non-convex landscape.
This `predictable progress' phenomenon is exploited by the AutoML community:
Since the optimization progress of a single run is already representative,
multiple retries with the same hyperparameters are not necessary.

 \end{abstract}


{ \hypersetup{hidelinks} \tableofcontents }

\section{Introduction}

\begin{figure}[h]
\begin{subfigure}[t]{0.42\linewidth}
            \def\svgwidth{\linewidth}
            \begingroup \makeatletter \providecommand\color[2][]{\errmessage{(Inkscape) Color is used for the text in Inkscape, but the package 'color.sty' is not loaded}\renewcommand\color[2][]{}}\providecommand\transparent[1]{\errmessage{(Inkscape) Transparency is used (non-zero) for the text in Inkscape, but the package 'transparent.sty' is not loaded}\renewcommand\transparent[1]{}}\providecommand\rotatebox[2]{#2}\newcommand*\fsize{\dimexpr\f@size pt\relax}\newcommand*\lineheight[1]{\fontsize{\fsize}{#1\fsize}\selectfont}\ifx\svgwidth\undefined \setlength{\unitlength}{262.5bp}\ifx\svgscale\undefined \relax \else \setlength{\unitlength}{\unitlength * \real{\svgscale}}\fi \else \setlength{\unitlength}{\svgwidth}\fi \global\let\svgwidth\undefined \global\let\svgscale\undefined \makeatother \begin{picture}(1,0.85714286)\lineheight{1}\setlength\tabcolsep{0pt}\put(0,0){\includegraphics[width=\unitlength,page=1]{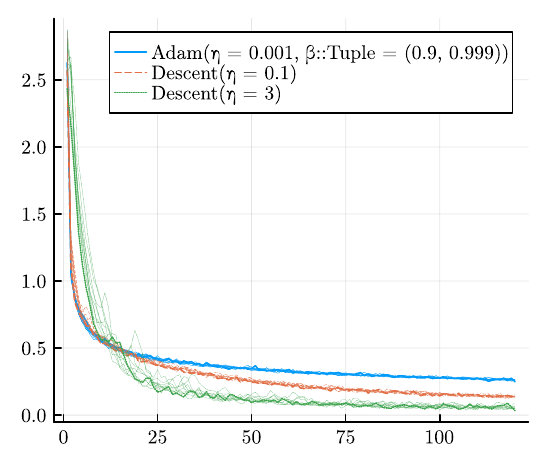}}\end{picture}\endgroup              \caption{
            The plot shows an empirical approximation of the cost sequence resulting
            from the training of a standard convolutional neural network on the MNIST dataset
            \citep{lecunMNISTDATABASEHandwritten2010}.
            We plot the values of \(f(\param_0^{(i)}),\dots, f(\param_{120}^{(i)})\) against the
            steps \(0,\dots,120\) on the \(x\)-axis. The minimization is performed with three optimization
            algorithms: Adam \citep{kingmaAdamMethodStochastic2015} (with learning rate \(\eta\) and
            momentum \(\beta\)) in blue and two version of gradient descent
            (learning rate \(\eta=0.1\) and \(\eta=3\)) in red and
            green. Each optimizer was run \(10\) times from randomly selected initializations 
            \(\param_0^{(i)}\) using the (random) default initialization procedure.
            }
            \label{fig: self loss plots}
        \end{subfigure}
        ~
        \begin{subfigure}[t]{0.50\linewidth}
            \includegraphics[width=\linewidth]{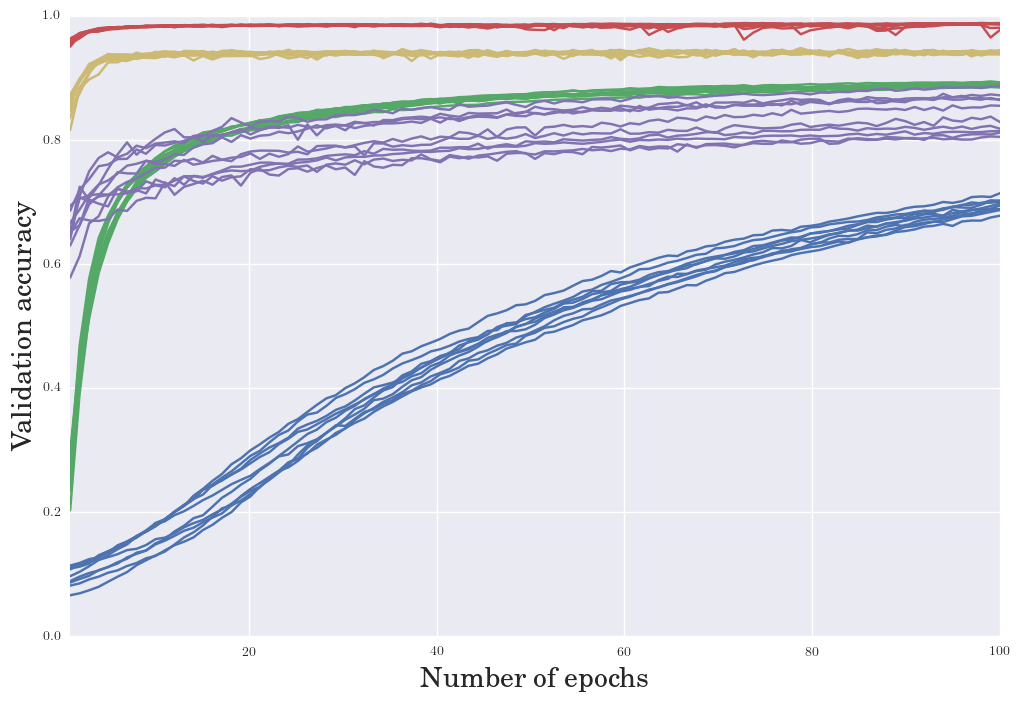}
            \caption{
                In the plot, taken from \citet[Figure~3]{kleinLearningCurvePrediction2017},
                optimization runs are grouped into 5 categories of hyperparameter settings
                represented by color. Each category
                contains 10 optimization runs using the same hyperparameter configuration.
                The predictable progress of the validation accuracy per
                configuration is used in \citet{kleinLearningCurvePrediction2017} to argue that it is
                sufficient to try a configuration once. The overall
                goal of \citet{kleinLearningCurvePrediction2017} is furthermore to fit a parametric models
                to the `learning curves' in order to stop training early and switch to a
                different configuration if the progression does not seem
                promising.
                \\
                The training heuristics that aspire to be `AutoML' (i.e.
                automatically fit data without human intervention) are therefore
                built on this phenomenon. And the fact that these training
                heuristics are so successful demonstrates how ubiquitous
                predictable progress is in practice.
            }
            \label{fig: other val plots}
        \end{subfigure}
\caption{Predictable progress in machine learning practice}
        \label{fig: loss plots}
    \end{figure}

\begin{figure}[h]
    \makebox[\textwidth]{
    \includegraphics[width=1.25\linewidth]{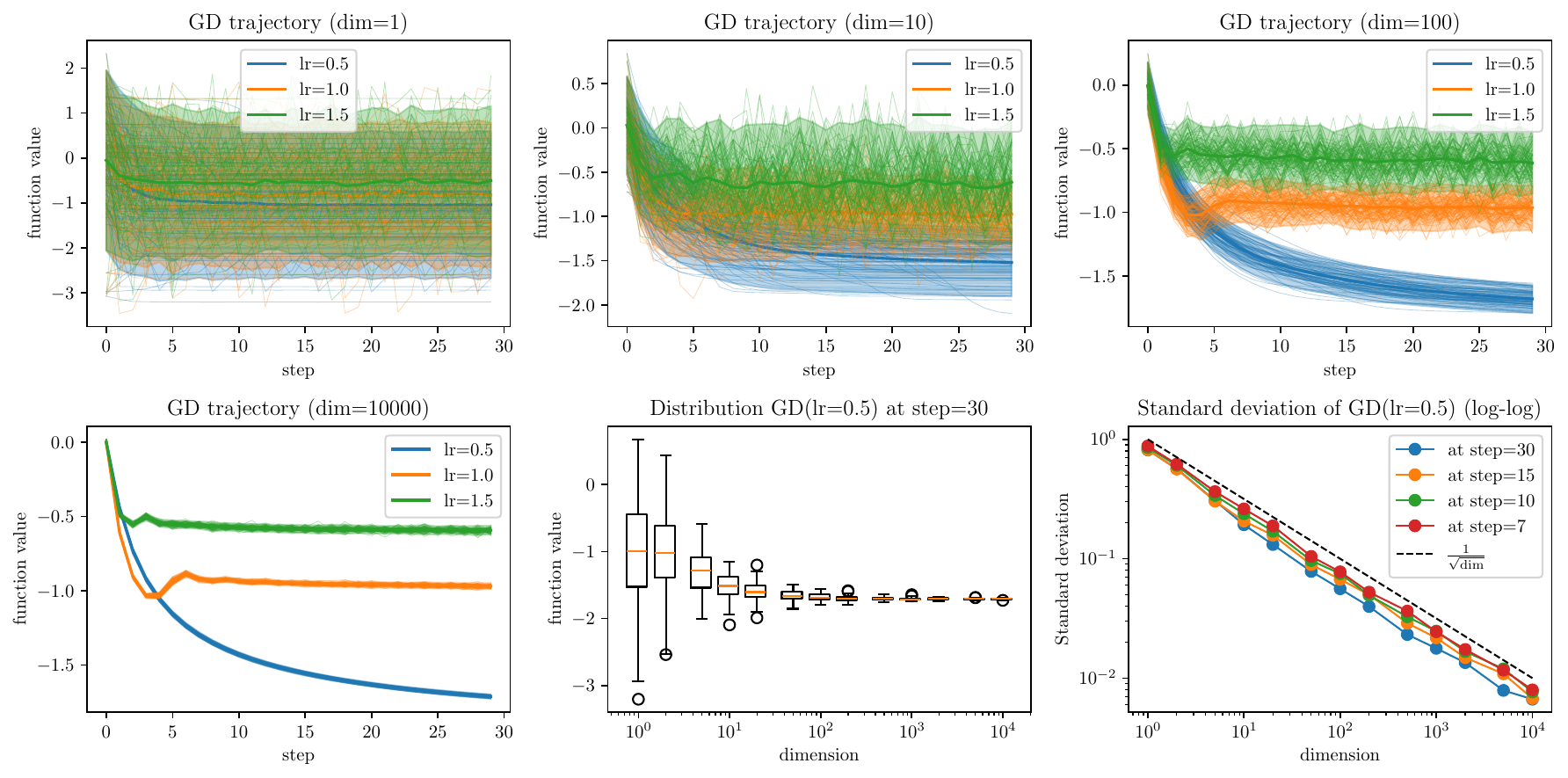}
    }
    \caption{
        For each learning rate hyperparameter we simulated \(100\) gradient
        descent trajectories on a centered Gaussian random function with covariance \(\C_\rf(x,y) = \frac1\dims\exp(-\frac{\|x-y\|^2}2)\).
        Plotted are the trajectories, their empirical mean and ribbons representing twice
        the empirical standard deviation. In the lower right we show how the variance
        of the trajectories decrease with the dimension. See Remark \ref{rem: computationally tracking the evolution}
        for more details.
    }
    \label{fig: simulated gradient descent}
\end{figure}

We present a theoretical explanation for a remarkable empirical
    phenomenon that occurs during the training of (large) machine learning models.
    `Training' is the process of minimizing a cost (also known as error or risk)
    function \(f\colon \real^\dims\to\real\) over parameter vectors \(\param\in
    \real^\dims\) by selecting successive parameter vectors
    \(\param_n\in\real^\dims\) based on noisy gradients
    at the previous parameter vectors \(\param_0,\dots \param_{n-1}\).
    The parameters are typically the weights
    of a neural network and $\dims$ is very large. The phenomenon is that over multiple
    initializations \(\param_0^{(1)},\param_0^{(2)},\dots \in \real^\dims\) the
    progress a particular optimizer makes during the optimization is approximately
    equal over different optimization runs:
    \[
        (f(\param_0^{(i)}), f(\param_1^{(i)}), \dots )
        \approx (f(\param_0^{(j)}), f(\param_1^{(j)}), \dots ).
    \]
    Figure~\ref{fig: loss plots} shows this `\textbf{predictable progress}' in practice. 
    Using a relatively high batch size (1000) to approximate the underlying noise-less cost,
    we demonstrate predictable progress ourselves for the training on the MNIST
    dataset using a standard convolutional neural network (cf. Figure~\ref{fig:
    self loss plots}). The key to predictable progress is high dimensionality of
    parameters (e.g.  the neural network of Figure~\ref{fig: self loss plots}
    has roughly \(\dims = 2.3\) million parameters, which is still small in comparison
    to neural networks used for tasks beyond the ``toy-problem'' MNIST).
    Figure~\ref{fig: other val plots} \citep{kleinLearningCurvePrediction2017}
    demonstrates that this phenomenon is well known and moreover relied upon for
    training heuristics. In Figure \ref{fig: simulated gradient descent} we demonstrate
    this phenomenon on synthetic random functions.

The predictable progress phenomenon is certainly surprising since there is no
reason to believe that function values along an optimization path should be approximately equal for different initializations. And, in view of handcrafted deterministic counterexamples,
predictable progress is obviously not provable for arbitrary high-dimensional functions.

But this phenomenon clearly exists empirically, and recently predictable
progress has been proven for \textbf{random quadratic functions} generated from
the mean squared error applied to linear models
\citep{paquetteHaltingTimePredictable2022,paquetteUniversalityConjugateGradient2022,deiftConjugateGradientAlgorithm2021,pedregosaAccelerationSpectralDensity2020,scieurUniversalAverageCaseOptimality2020}.
While quadratic functions are a simplified convex setting, these contributions
offered a first explanation for the observed phenomenon of predictable behavior
in high dimensions. We will extend these results to the much more general, non-convex
setting of Gaussian random functions.
This setting was already used in the machine learning literature for instance by  
\citet{pascanuSaddlePointProblem2014} and
\citet{dauphinIdentifyingAttackingSaddle2014} to explain why
the overwhelming share of critical points are saddle points in high dimension,
whereas the critical points of low dimensional GRFs are dominated by minima and
maxima \citep{rasmussenGaussianProcessesMachine2006}.
To lighten the technical difficulty that arises from this generalization, we do
not consider stochastic gradients of sample losses. Instead we assume the
algorithms operate with full gradients of the cost.

The specific modelling assumption of \citet{pascanuSaddlePointProblem2014} and
\citet{dauphinIdentifyingAttackingSaddle2014} were stationary isotropic GRFs,
which we will extend to (non-stationary) isotropic GRFs
to include the setting of random quadratic functions and more
generally dot product kernels (i.e.\ spin glasses) which are both non-stationary
(see Section \ref{sec: random quadratic functions}).\smallskip

Similar to the convention of capital letters for random variables, we use
bold font to denote random functions \(\rf\). Since our
results are limit theorems in the dimension \(\dims\), we mark this important
dependency in the index. While the parameter sequences would also have to
be indexed by the dimension \(\dims\) we omit this index for notational
clarity. In simplified terms we will prove the following theorem for predictable
optimization in high dimension, precise theorems are given in Section
\ref{sec:setting}:

\tikz[baseline]{\draw[thick] (0,1) -- (0,-1); \node[anchor=west] at (0,0) {\begin{minipage}[t]{\columnwidth-2\parindent} \textbf{
    If \(\rf_\dims\) is a (non-stationary) isotropic GRF on a 
    high-dimensional domain \(\real^\dims\), then the (random)  sequence 
    \(\rf_\dims(\Param_0),    \rf_\dims(\Param_1),...\) along the (random) parameter sequence \(\Param_0,
    \Param_1,...\) selected by a standard first order
    optimization algorithm are close to a deterministic sequence
    \(\limf_0,\limf_1,...\) with high probability. 
    }
\end{minipage}
};
}\smallskip

To be a bit more
precise, we will prove the following in Theorem~\ref{thm: asymptotically
deterministic behavior}: Suppose
\(\Param_0,\Param_1,...\) is the (random) sequence of parameter points
obtained by running an optimizer on the (random) function \(\rf_\dims\colon
\real^\dims \to \real\) initialized at an independent, possibly random point
\(\Param_0\). For `gradient span algorithms' \(\gsa\) (e.g. gradient
descent) and a sequence of (non-stationary) isotropic GRFs
\((\rf_\dims)_{\dims\in\nat}\) we construct a sequence of deterministic real numbers 
\(\limf_0, \limf_1,...\) such that, for $n\in\mathbb N$, 
\[
\lim_{\dims\to\infty}\Pr\Bigl(|\rf_\dims(\Param_\timestep) - \limf_\timestep| > \epsilon\Bigr) = 0,\quad \forall\epsilon>0.
\]
The proof is completely constructive (Remark \ref{rem: explicit evolution formula})
and the limiting values \(\limf_\timestep\) may be computed with complexity
\(\bigO(\timestep^6)\) given an algorithm \(\gsa\), the mean and covariance kernel of
the random functions \(\rf_\dims\) and the length of the initialization vector
\(\Param_0\) (Remark \ref{rem: computationally tracking the evolution}). In
Figure~\ref{fig: simulated gradient descent} we demonstrate convergence
empirically by applying gradient descent to simulations of a GRF in various dimensions.
In Corollary~\ref{cor: asymptotically identical cost curves} we
will show that an application of the stochastic
triangle inequality yields approximately equal
optimization progress given different initialization points (as can be seen in
Figure~\ref{fig: loss plots} and \ref{fig: simulated gradient descent}).

\paragraph*{Related work}

Beyond the work on random quadratic functions, our article is
closely related in spirit to work from \textbf{statistical mechanics} where similar high
dimensional limits are considered (referred to as the `thermodynamic limit').
Specifically, isotropic random functions restricted to the sphere coincide with
the ``spherical \textbf{spin-glasses}'' from statistical mechanics. This fact is not obvious,
as \(p\)-spin glasses are defined explicitly as random homogeneous
\(p\)-th order multivariate polynomials,
i.e.\footnote{
    Spin glasses were historically defined on the domain \(\{-1, +1\}^N\).
    Their spherical counterpart was therefore defined on the the sphere of
    radius \(\sqrt{N}\) which contains this set. However, with the definition of
    a rescaled inner product (the ``overlap''), they are effectively mapped
    to the unit sphere. Also note that the canonical ``Hamiltonian'' \(H_{\dims, p}\)
    is of the form \(H_{\dims, p}(x) = \dims\rf_{\dims,
    p}(\sqrt{\dims}x)\). The outer scaling is compatible with ours, since results
    in statistical mechanics are proven about \(H_{\dims,p}(x)/\dims\).
}
\begin{equation}
   \label{eq: p-spin glass} 
    \rf_{\dims,p}(x) = \frac1{\sqrt\dims} \sum_{i_1, \dots, i_p=1}^\dims J_{i_1,\dots, i_p} x_{i_1} \cdots  x_{i_p}
    = \frac1{\sqrt\dims}\langle J, x^{\otimes p}\rangle,
\end{equation}
with all entries of the tensor \(J=(J_{i_1,\dots, i_p})_{i_1,\dots, i_p}\) iid standard normal distributed.  However it is
straightforward to calculate their covariance function
\(\Cov(\rf_{\dims,p}(x), \rf_{\dims,p}(y))=\tfrac1\dims \langle x,y\rangle^p\),
which fully characterizes a centered Gaussian random function. And it turns out that the
mixtures of independent \(p\)-spin glasses
\[
    \rf_\dims(x) = \sum_{p=0}^\infty a_p \rf_{\dims, p}(x),
    \qquad a_n \in \real,
\]
have the covariance \(\C_{\rf_\dims}(x,y)= \sum_{p=0}^\infty a_p^2 \langle
x,y\rangle^p\), which exhaust all continuous isotropic covariance kernels on
the sphere that are valid in all dimensions
\citep{schoenbergPositiveDefiniteFunctions1942}.

Results about `spin glasses', such as the celebrated Parisi-formula are
therefore general results about isotropic random functions on the sphere.
A recent review of results relevant for optimization can be found in
\citep{auffingerOptimizationRandomHighDimensional2023}.  The Parisi-formula
\citep{parisiSequenceApproximatedSolutions1980,talagrandParisiFormula2006,panchenkoSherringtonKirkpatrickModel2013,huangConstructiveProofSpherical2024}
provides the limiting value of the global maximum/minimum
of spin glasses in the high dimensional limit. Beyond this static analysis,
optimization algorithms have 
recently been analyzed as well. In particular, there is a hardness result
\citep{huangTightLipschitzHardness2022}, which establishes an algorithmic barrier
that does not necessarily coincide with the global minimum. Even
more recently, \citet{sellkeThresholdEnergyLow2024} showed for pure \(p\)-spin
glasses that this algorithmic barrier may be reached by `Langevin-dynamics',
i.e.\ a continuous time model of stochastic gradient descent. 

The tensor form of spin glasses allows for the use of message passing
algorithms, specifically `Approximate Message Passing' (AMP) \citep{donohoMessagepassingAlgorithmsCompressed2009}.
AMP was originally developed to understand programs that repeatedly multiply
the same random matrix to non-linear functions of the iterate.
Since the same random matrix is used repeatedly, the iterate becomes correlated
with the matrix and it becomes necessary to control this `self-interaction' by
conditioning. This approach has been generalized to multiple matrices in the
`Tensor Program' series \citep[e.g.][]{yangTensorProgramsIII2021} and it has also
been generalized to spin glasses made up of Tensors
\citep{alaouiOptimizationMeanfieldSpin2021}. By translating first order
algorithms into AMP
\citep{celentanoEstimationErrorGeneral2020}, `predictable-progress' on spin
glasses may be recoverable from the general concept of
`state-evolution'
\citep{bayatiDynamicsMessagePassing2011,javanmardStateEvolutionGeneral2013,alaouiOptimizationMeanfieldSpin2021}.
`State-evolution', referred to as the `Master theorem' in \citep{yangTensorProgramsIII2021}
provides very general asymptotic statements about AMP algorithms.
However, the random function requires a tensor representation to
be expressed as AMP, which is currently unavailable for general
isotropic random functions on \(\real^\dims\). AMP is therefore not applicable.
This may change in the future as the recent characterization of
(non-stationary) isotropic covariance kernels
\citep{benningSchoenbergCharacterizationContinuous2025} represents
significant progress towards a tensor representation of general isotropic random functions.

In contrast to AMP, our proof technique does not rely on a Tensor representation and is entirely
functional in nature. We have to use a similar conditioning technique to
control the self-interaction, but our conditioning is closer in spirit to the one
that is used in \textbf{Bayesian optimization} (BO)
\citep[e.g.][]{kushnerNewMethodLocating1964,frazierBayesianOptimization2018}. This suggests
that it may be easier to analyze Bayesian optimization algorithms using our approach.
And the intuitiveness of our approach would only be obfuscated by a
translation of first order algorithms into AMP.

Our proof relies on two new techniques: custom coordinate systems and a functional
conditioning technique formalized by
\citet{benningMeasureTheoryConditionally2026}. These could be combined with
techniques from the AMP literature to significantly strengthen our results in future work.
For example, it should be possible to strengthen convergence in probability to almost sure
convergence.

\paragraph*{Organisation of the article}

In Section \ref{sec:setting} we formalize the setting
(gradient span optimization algorithms and (non-stationary) isotropic GRFs) and state the
main result. 
Our asymptotic formulation of the original problem requires a dimensional scaling of the random functions that
is also standard in statistical mechanics. This scaling, crucial to our approach, is
discussed in Section \ref{sec: discuss scaling}.
The proof of our main result is
given in Section \ref{sec:proof}, first as a sketch assuming stationary isotropy
and then in detail. In Appendix \ref{sec: random quadratic functions} we
show random quadratic functions to be a special case of isotropic Gaussian
random functions, in Appendix \ref{sec: conditional gaussian distribution} we provide a
refresher on conditional Gaussian distributions, as we rely heavily upon them, and
in Appendix \ref{sec: strictly pos definite derivatives} we prove results that allow
the strict positive definiteness assumption to be dropped in the case of
\emph{stationary} isotropy.

 \section{Setting and main results}\label{sec:setting}

To enable our asymptotic analysis, a careful setting of the scene is required.
Both, the optimization algorithms considered and the distribution of \(\rf_\dims\)
need a representation independent of the dimension \(\dims\) to allow for an analysis of
\(\dims\to\infty\).

\subsection{The class of optimization algorithms \texorpdfstring{\(\gsa\)}{}}\label{sec:algos}

Given a sufficiently smooth function \(f:\mathbb R^\dims\to \mathbb R\),
a naive optimization algorithm is given by gradient descent, whose evaluation points are
recursively defined as
\begin{align*}
	\param_{\timestep}=\param_{\timestep-1}- \alpha \nabla f (\param_{\timestep-1})
\end{align*}
with some initialization \(\param_0\in\real^\dims\) and a learning rate \(\alpha\).
The reader might want to keep gradient descent in mind as a toy example, but
everything we prove holds for a much larger class of first order algorithms that
contains many standard optimizers. Gradient span algorithms (GSA) are a very
general class of first order algorithms which pick the \(\timestep\)-th point
\(\param_\timestep\) from the previous span of gradients
\begin{equation}
	\label{eq: gsa property}
	\param_\timestep
	\in \Span\{
		\param_0, \nabla f(\param_0), \dots, \nabla f(\param_{\timestep-1})
	\}.
\end{equation}
GSAs contain classic gradient descent, momentum methods
such as heavy-ball momentum \citep{polyakMethodsSpeedingConvergence1964}
and Nesterov's momentum \citep[e.g.][]{nesterovLecturesConvexOptimization2018},
and also the conjugate
gradient method \citep{hestenesMethodsConjugateGradients1952}. GSAs do not only
encompass minimizers but also maximizers and all sorts of other algorithms. While the initial point \(\param_0\) is typically not included in the
span, we admit this generalization to allow for concepts such as
`weight normalization' \citep{salimansWeightNormalizationSimple2016}, which
project points back to the sphere (see Remark~\ref{rem: projection} for further
details). Since the defining property \eqref{eq: gsa property} of gradient span algorithms 
is not sufficient to identify a particular GSA, we define a very general
parametric family of GSAs in Definition~\ref{def: general gsa}. This family
includes all the algorithms mentioned above.
Importantly, the parametrization
chosen does not use dimension specific information and a fixed gradient span
algorithm (such as gradient descent) can therefore be used for all dimensions.
In Table \ref{table: characterization of gsa} we use a non-exhaustive list of
popular algorithms to illustrate the generality of gradient span
algorithms. The main limitation are preconditioning methods, which are
further discussed in Remark~\ref{rem: preconditioning}.

\begin{table}
	\centering
	\begin{tabular}{c| p{1.3cm} p{1.6cm} p{1.4cm} p{2.2cm} | c c}
		\multicolumn{1}{c}{}&\multicolumn{4}{c|}{} & \multicolumn{2}{c}{preconditioning}
		\\
		\hline
		Example & gradient descent & momentum methods & conjugate gradient descent & adaptive \underline{scalar} learning rate & Adam & Shampoo
		\\
		\hline
		is GSA? & \multicolumn{1}{c}{\cmark} &  \multicolumn{1}{c}{\cmark} & \multicolumn{1}{c}{\cmark} & \multicolumn{1}{c|}{\cmark} & \xmark & \xmark
	\end{tabular}
	\caption{What algorithms are gradient span algorithms (GSA)?}
	\label{table: characterization of gsa}
\end{table}

\begin{definition}[General gradient span algorithm]
	\label{def: general gsa}
    For a starting point \(\param_0\in \real^\dims\) and a function
    \(f:\real^\dims\to\real\), a \emph{gradient span algorithm} \(\gsa\)
    selects
    \begin{equation}
		\label{eq: definition gsa}
        \param_\timestep := \gsa(f, x_0, \timestep)
		= \lr^{(\param)}_{\timestep} \param_0
		+ \sum_{k=0}^{\timestep-1}\lr_{\timestep,k}^{(g)}\nabla f(\param_k),
	\end{equation}
	where we assume the prefactors \(\lr_\timestep = (\lr^{(\param)}_\timestep) \cup
	(\lr^{(g)}_{\timestep,k})_{k=0,\dots,\timestep-1}\), using the
	union \(\cup\) to indicate a concatenation of tuples,
	to be functions \(\lr_\timestep = \lr_\timestep(\info_{\timestep-1})\)
	of the previous dimensionless information \(\info_{\timestep-1}\) available at time \(\timestep\), that is
	\[
		\info_\timestep
		:= \Bigl(f(\param_k): k \le \timestep\Bigr)
		\cup \Bigl(
			\langle v, w\rangle :
			v,w \in (\param_0) \cup \gradients_\timestep
		\Bigr)
		\quad\text{with}\quad
		\gradients_\timestep := \bigl(\nabla f(\param_k): k \le \timestep\bigr).
	\]
    The algorithm \(\gsa\) is called \emph{\(\param_0\)-agnostic}, if 
    \begin{itemize}
        \item it remains in the gradient span shifted by \(\param_0\), i.e.
        \[
            \lr_\timestep^{(x)} = 1
            \qquad\forall \timestep\in \nat.
        \]
        \item The prefactors \(\lr_\timestep\) do not use the inner products
        with \(\param_0\), that is they are functions of the reduced information
        \[
            \info^{\setminus \param_0}_\timestep
            = \Bigl(f(\Param_k), \langle \nabla f(\param_k), \nabla f(\param_l)\rangle : k, l \le \timestep\Bigr).
        \]
    \end{itemize}
\end{definition}
As mentioned above the reader might just think about ordinary gradient descent,
which is also \(\param_0\)-agnostic.
We introduce the `\(\param_0\)-agnostic' property, to allow for arbitrary initialization
distributions in the stationary isotropic case. Without this property, the algorithm
could, for example, use the length \(\|\Param_0\|\) as an indicator to switch between optimizers and
therefore cause non-deterministic behavior.

\begin{remark}[Initial point is not special]
	It should perhaps be noted, that the initial point \(\param_0\) is not the only
	point that may be used by gradient span algorithms, since we have
	\[
		\Span\{ \param_0, \nabla f(\param_0), \dots, \nabla f(\param_{\timestep-1})\}
		= \Span\{ \param_0, \dots, \param_{\timestep-1}, \nabla f(\param_0), \dots, \nabla f(\param_{\timestep-1})\}
	\]
	by induction over \(\timestep\) as \(\param_\timestep\) is selected from this span.
\end{remark}

\begin{remark}[The same algorithm for all dimensions]
	Since the main objective of this article is a dimensional limit statement for
	fixed algorithms (i.e.\ fixed choice of prefactors \(\lr\)) it is important to
	note that every fixed gradient span algorithm \(\gsa\) is well-defined in all
	dimensions because the scalar product \(\langle\cdot,\cdot\rangle\) is well
	defined for every dimension \(\dims \in \nat\).\smallskip
\end{remark}

\begin{remark}[No preconditioning]
	\label{rem: preconditioning}
	Gradient span algorithms notably exclude preconditioning methods such as
	Adam \citep{kingmaAdamMethodStochastic2015} and Shampoo
	\citep{guptaShampooPreconditionedStochastic2018}, which apply a
	preconditioning matrix \(P_\timestep\) to the gradients and thereby
	leave the span of gradients. The reason is related to the previous remark:
	We want ``the same algorithm in every dimension'' and the GSA family only uses
	dimensionless information which can be shown to converge as the dimension
	tends to infinity. In contrast, preconditioning methods collect information about
	each coordinate separately and treat them differently accordingly. This makes it
	much more challenging to analyze their limiting behavior in the dimension as
	the limiting behavior can no longer be captured by a simple number.  Since
	Adam still seems to behave predictably in high dimension
	(Figure \ref{fig: loss plots}), it is likely possible, albeit much more
	challenging to analyze its limiting behavior (see also Section \ref{sec: outlook}, \ref{it: assumption gsa}).
\end{remark}

\begin{remark}[Projection]
	\label{rem: projection}
	We claimed that the inclusion of the initial point \(\param_0\) into the span,
	or equivalently the  prefactor \(\lr^{(\param)}_\timestep\), would allow for
	algorithms projecting back to the sphere or ball. In the following,
	we show that our  general gradient span algorithms also contain gradient span
	algorithms with spherical projections.

	Assume that the point \(\param_\timestep\) is defined by
	\[
		\param_\timestep := P \tilde{\param}_\timestep
		\quad\text{with}\quad
		\tilde{\param}_\timestep
		= \tilde{\lr}^{(\param)}_\timestep \param_0
		+ \sum_{k = 0}^{\timestep-1} \tilde{\lr}^{(g)}_{\timestep, k}\nabla f(\param_k),
	\]
	where \(P\) is either a projection to the sphere or the ball. Note that this implies
	either a division by \(\|\tilde{\param}_\timestep\|\) in case of the sphere,
	or a division by \(\max\{\|\tilde{\param_\timestep}\|,1\}\) in case of the
	ball. But the norm
	\[
		\|\tilde{\param}_\timestep\|^2
		=(\tilde{\lr}^{(\param)}_\timestep)^2
		\underbrace{\|\param_0\|^2}_{\in \info_\timestep}
		+ 2\tilde{\lr}^{(\param)}_\timestep \sum_{k=0}^{\timestep-1}
		\tilde{\lr}^{(g)}_{\timestep, k}\underbrace{
			\langle \nabla f(\param_k), \param_0\rangle
		}_{\in \info_\timestep}
		+ \sum_{k,l=0}^{\timestep-1}
		\tilde{\lr}^{(g)}_{\timestep, k}\tilde{\lr}^{(g)}_{\timestep, l}
		\underbrace{
			\langle \nabla f(\param_k),\nabla f(\param_l)\rangle
		}_{\in \info_{\timestep}}.
	\]
	is a function of the information in \(\info_{\timestep-1}\) and can therefore be
	used to define
	\[
		\lr^{(\param)}_\timestep := \tfrac{\tilde{\lr}^{(\param)}_\timestep}{\|\tilde{\param}_\timestep\|}
		\qquad
		\lr^{(g)}_{\timestep,k} := \tfrac{\tilde{\lr}^{(g)}_{\timestep,k}}{\|\tilde{\param}_\timestep\|}
	\]
	in the case of the projection to the sphere and similarly for the projection to the ball.
\end{remark}

\subsection{The class of random function distributions}
\label{sec: class of distributions}

For a random function\footnote{
	Recall that `random process', `stochastic process' and `random field' are
	used synonymously for `random function' in the literature, that their law is
	characterized by all finite dimensional marginals
	\citep[e.g.][Thm.~14.36]{klenkeProbabilityTheoryComprehensive2014} and that
	Gaussian random functions are fully determined by their mean and covariance.
} 
\(\rf_\dims:\real^\dims\to\real\) on some probability space $(\Omega, \mathcal A, \mathbb P)$ we denote the
mean and covariance functions by
\[
	\mu_{\rf_\dims}(x) := \E[\rf_\dims(x)]
	\quad\text{and}\quad 
	\C_{\rf_\dims}(x,y):= \Cov(\rf_\dims(x), \rf_\dims(y)),
	\qquad x,y\in \real^\dims.
\]
As we want to prove a limit theorem for \(\dims\to\infty\) to make predictions
about the high dimensional cost functions found in practice, the mean \(\mu_{\rf_\dims}\)
and covariance \(\C_{\rf_\dims}\) needs to be defined for every dimension. They
should also remain constant in some sense to avoid arbitrary sequences.
As the domain \(\real^\dims\) changes over \(\dims\) this requires a
parametric form provided by covariance kernels.
To obtain non-trivial results, some scaling is also required (discussed in Section~\ref{sec:
discuss scaling}).

\begin{definition}[Scaled sequence of isotropic Gaussian random functions (GRF)]
	\label{def: isotropic gaussian random function}	
	A scaled sequence \((\rf_\dims)_{\dims\in\nat}\) of GRFs \(\rf_\dims:\mathbb
	R^\dims \to \mathbb R\) is called
	\begin{enumerate}[label=(\roman*)]
	\item \textbf{(non-stationary) isotropic}, if
	\begin{equation}
		\label{eq: canonical parametrization}
		\mu_{\rf_\dims}(x) = \mu\bigl(\tfrac{\|x\|^2}2\bigr)
		\qquad\text{and}\qquad
		\C_{\rf_\dims}(x,y)
		= \red{\tfrac1{\dims}} \kernel\bigl(\tfrac{\|x\|^2}2,\tfrac{\|y\|^2}2, \langle x,y\rangle\bigr)
		,\quad x,y\in \real^{\red{\dims}},
	\end{equation}
	for continuous functions \(\mu\colon\real\to\real\) and 	 \(\kernel\colon D
	\to \real\) with \(
		D = \{ \lambda\in \real_{\ge 0}^2 \times \real : |\lambda_3| \le 2\sqrt{\lambda_1 \lambda_2}\}
		\subseteq\real^3.
	\)
	In that case we write \(\rf_\dims\sim\normal(\mu, \kernel)\).

	\item
	\textbf{stationary isotropic}, if
	\[
			\mu_{\rf_\dims}(x) = \mu \in \real
			\qquad\text{and}\qquad
			\C_{\rf_\dims}(x,y)
			= \red{\tfrac1{\dims}}\ikernel\bigl(\tfrac{\|x-y\|^2}2\bigr),
			\quad x,y\in \real^{\red\dims},
	\]
	for a continuous \(\ikernel\colon \real_{\ge 0}\to \real\).
		In that case we write \(\rf_\dims\sim\normal(\mu, \ikernel)\).
	\end{enumerate}	
\end{definition}

\begin{remark}[Motivation for isotropy]
    While we define isotropy as a functional form of the covariance
    function, isotropy is best understood axiomatically as an invariance to
    rotation and reflection
    \citep[e.g.][]{benningSchoenbergCharacterizationContinuous2025}. To motivate
    this invariance in the context of predictability let us highlight a case
    that must be avoided.

    Think about some one-dimensional function
    \(f_1\colon\real^1\to\real\) where the values along the
    optimization path (say of gradient descent) strongly depend on the
    initialization. Lift this one dimensional function into \(\real^\dims\) by
    \[
	f_\dims(x) := f_1(\langle v,x\rangle)
	\] for some direction vector \(v\). Then
    \(f_\dims\) certainly does not share the features of high dimensional cost
    functions encountered in machine learning, since the cost along the optimization
    path depends heavily on the initialization again. To explain the phenomenon
    of predictable cost sequences  (cf.~Figure~\ref{fig: loss plots})
    the model for cost functions therefore should not be reliant on particular directions
    \(v\). Thus, we assume that directions should be exchangeable, which suggests at least 
    rotation and reflection invariant random functions as a model.

	Non-stationary isotropy as a distributional assumption for cost functions was
	introduced by \citet{benningRandomFunctionDescent2024} as a generalization of the common
	stationary isotropy assumption
	\citep[e.g.][]{dauphinIdentifyingAttackingSaddle2014}. This was motivated by the
	fact that simple linear models already break the stationary isotropy assumptions.
	Linear models equipped with a mean squared loss can also be used to derive
	random quadratic functions (Section~\ref{sec: random quadratic functions}), which have been analyzed before and are isotropic.
\end{remark}

Before getting to the main results, we have to define one more concept. For
a given mean function \(\mu\) and covariance kernel \(\kernel\) we are interested in
a sequence of (non-stationary) isotropic random functions \(\rf_\dims\sim\normal(\mu,
\kernel)\) over the dimension \(\dims\). But it is not clear that a
function \(\kernel\) (resp. \(\ikernel\)) corresponds to a random function for every
dimension, as it may not always define a positive definite kernel. If it does, we speak of validity in all dimensions.
\begin{definition}[Valid in all dimensions]
	\label{def: valid in all dimensions}
	We say \(\kernel\) (resp. \(\ikernel\)) is valid in all
	dimensions if for any dimension \(\dims\) there exists a (non-stationary)
	isotropic random function \(\rf_\dims\sim \normal(\mu, \kernel)\) (respectively
	stationary isotropic \(\rf_\dims\sim \normal(\mu,
	\ikernel)\)).
\end{definition}

Since one can always restrict the domain of a random function
\(\rf_\dims\) to a lower dimensional subspace it should be clear that the only
possible type of restriction is an upper bound on the dimension.
We refer the reader to \citet[Section
3.8]{sasvariMultivariateCharacteristicCorrelation2013} for the ``Schoenberg''
characterization of stationary isotropic covariance kernels that are valid up to
dimension \(\dims\) \citep{schoenbergMetricSpacesPositive1938}. The stationary isotropic covariance functions which are valid
in all dimensions are given exactly by the kernels of the form
\begin{equation}
	\label{eq: schoenberg characterization I}
    \ikernel(r) = \int_{[0,\infty)} \exp(-t^2 r) \schoenbergMeas(dt),\qquad r\geq 0,
\end{equation}
for some finite measure \(\schoenbergMeas\) on \([0,\infty)\)
\citep[Theorem~3.8.5]{sasvariMultivariateCharacteristicCorrelation2013}.
The (non-stationary) isotropic kernels have only very recently been classified
\citep{benningSchoenbergCharacterizationContinuous2025}.
The continuous isotropic kernels valid in all dimensions are of the form
\[
	\kernel(\tfrac{\|x\|}2, \tfrac{\|y\|}2, \langle x,y\rangle)
	= \sum_{n=0}^\infty \alpha_n(\|x\|, \|y\|) \langle \tfrac{x}{\|x\|}, \tfrac{x}{\|x\|}\rangle^n,
\]
where the \(\alpha_n\) are continuous positive definite kernels on \([0, \infty)\).
With \(\alpha_n(x,y) = a_n \|x\|^n \|y\|^n\) for \(a_n\in \real\), dot product kernels
 represent a wide
class of (non-stationary) isotropic random functions that are not stationary
isotropic on \(\real^\dims\) but also valid in all dimensions. Restricted to
the sphere, dot product kernels are known as spin glasses in statistical
mechanics.

\subsection{Main result: predictable progress in high dimensions}\label{sec:mainresults}

Our main result proves that the optimization path
\(\rf_\dims(\Param_0),\rf_\dims(\Param_1),...\) is asymptotically deterministic for large dimension \(\dims\). In addition, we
prove a similar statement about the gradient norms
\(\|\nabla\rf_\dims(\Param_\timestep)\|^2\) and more
generally about their scalar products. As we assume continuity of the
prefactors \(\lr_\timestep\) for the gradient span algorithm \(\gsa\), they are
also asymptotically deterministic due to continuous mapping. To reduce the initial complexity,
we first state the simplified corollary for the stationary isotropic case.
In this case some of the technical assumptions can be removed which should allow
the reader to focus on the core message.

\begin{corollary}[Predictable progress, isotropic case]
	\label{cor: asymptotically deterministic behavior}
	Let \(\ikernel\) define a stationary isotropic kernel valid in all dimensions
	(Definition~\ref{def: valid in all dimensions}) and let
	\(\rf_\dims \sim \normal(\mu, \ikernel)\) be a sequence of scaled isotropic
	Gaussian random functions in \(\dims\) (Definition~\ref{def: isotropic gaussian random
	function}).
	Let \(\gsa\) be a general gradient span algorithm (Definition~\ref{def:
	general gsa}), where we assume that its prefactors
	\(\lr_\timestep=\lr_\timestep(\info_{\timestep-1})\) are continuous in the
	information \(\info_{\timestep-1}\), and utilize the most recent gradients, i.e.
	\(\lr^{(g)}_{\timestep,\timestep-1} \neq 0\) for all \(\timestep\in \nat\).
	Let the algorithm \(\gsa\) be applied to \(\rf_\dims\) with independent,
	possibly random starting point \(\Param_0\in \real^\dims\) and denote by
	\(\Param_\timestep\) the points resulting from \(\gsa\). 
	Finally, assume that \(\|\Param_0\|=\radius\) almost surely.
	Then there exist characteristic real numbers
	\begin{align*}
		\limf_\timestep &= \limf_\timestep(\gsa, \mu, \kernel, \radius) \in \real
		\quad \text{and}\quad
		\limgdot_{\timestep,k} = \limgdot_{\timestep,k}(\gsa, \mu, \kernel, \radius) \in \real,
	\end{align*}
	such that, for all \(\epsilon>0\), \(\timestep, k\in\nat\),
	\begin{align*}
		&\lim_{\dims\to \infty}\Pr\bigl(|\rf_\dims(X_\timestep) - \limf_\timestep| > \epsilon\bigr) = 0
		\quad\text{and}\quad
		\\
		&\lim_{\dims\to \infty}\Pr\Bigl(
			\bigl|\langle \nabla\rf_\dims(X_\timestep), \nabla\rf_\dims(X_k)\rangle - \limgdot_{\timestep,k}\bigr| > \epsilon
		\Bigr) = 0.
	\end{align*}
	If furthermore the algorithm is \(\param_0\)-agnostic (e.g. gradient descent), then the limiting values do not depend
	on \(\radius\) and the starting point \(\Param_0\) can have arbitrary distribution
	independent of \(\rf_\dims\).
\end{corollary}
	
Since most algorithms are \(\param_0\)-agnostic, this result explains the
approximately deterministic behavior in high dimension for stationary
isotropic GRFs. The general case is covered by the following remark. 

\begin{remark}[Initialization on the Sphere]
	Initialization procedures like Glorot initialization
	\citep{glorotUnderstandingDifficultyTraining2010} select the
	entries of
	\(\Param_0\) independent, essentially identically distributed, and scaled in
	such a way that the norm does not diverge.
	The norm therefore obeys a law of large numbers and is thus plausibly
	deterministic in high dimension. The assumption, \(\|\Param_0\|=\radius\)
	almost surely, is therefore realistic.
	This explains the phenomenon of (approximately) predictable progress of
	optimizers started in a randomly selected initial point using
	Glorot initialization, as observed in Figure~\ref{fig: loss plots}.
\end{remark}

The following proof of Corollary~\ref{cor: asymptotically deterministic behavior} should
also serve as a reading aid for Theorem~\ref{thm: asymptotically deterministic behavior},
which states the same result in greater generality.
The proof shows how the general statements of Theorem~\ref{thm: asymptotically deterministic behavior}
simplify to the more digestible statement for stationary isotropic GRFs.
\smallskip

\begin{proof}
	Observe that \(\rf_\dims(\Param_k)\) and \(\langle \nabla\rf_\dims(\Param_k), \nabla\rf_\dims(\Param_l)\rangle\)
	for \(k,l\le \timestep\) are members of the random information vector
	\(\Info_\timestep\) defined in Theorem~\ref{thm: asymptotically
	deterministic behavior}. Their convergence therefore follows immediately
	from the convergence of the information vector \(\Info_\timestep\) proven in
	Theorem~\ref{thm: asymptotically deterministic behavior}.
	Since all stationary isotropic random functions are (non-stationary) isotropic,
	this Corollary follows immediately from
	Theorem~\ref{thm: asymptotically deterministic behavior} once we verified the additional
	required assumptions.\smallskip

	The first assumption is the smoothness assumption on the covariance function  (Assumption~\ref{assmpt: smoothness}). It follows from the fact that
	\emph{all} stationary isotropic random function, which are valid in all dimensions,
	are infinitely differentiable by the Schoenberg characterization
	\citep[Theorem 3.8.5]{sasvariMultivariateCharacteristicCorrelation2013}, see also \eqref{eq: schoenberg characterization I}.
	The second assumption is the strict positive definiteness of
	\((\rf_\dims, \nabla\rf_\dims)\), which also holds for all stationary isotropic random
	functions valid in all dimensions by Corollary~\ref{cor: strict positive
	definite} below. Note that Corollary~\ref{cor: strict positive
	definite} requires that the random function \(\rf_\dims\) is not almost
	surely constant. But if \(\rf_\dims\) were almost surely constant, then we
	get asymptotically deterministic behavior from the fact that
	\(\rf_\dims(\param_0) \sim \normal(\mu, \frac1\dims\ikernel(0))\), i.e.
	\(\rf_\dims(\param_0)\to \mu\). Since \(\rf_\dims\) is almost surely constant,
	we thus obtain \(\rf_\dims \to \mu\) uniformly in probability.
	The \(\param_0\)-agnostic case follows from the stationary case of
	Proposition~\ref{prop: wlog deterministic starting points}.
\end{proof}
We now come to the main theorem of the paper, predictable progress of gradient span algorithms on (non-stationary) isotropic random functions in high dimensions. 
While kernels of stationary isotropic GRFs that are valid in all dimensions are always smooth, and the mean function is always constant,
the same may not be the case for the more general situation of (non-stationary) isotropy. We will therefore assume the following smoothness properties:
\begin{assumption}[Sufficiently smooth]
	\label{assmpt: smoothness}
	With \(\kernel_{i}(\lambda_1,\lambda_2,\lambda_3)
	:= \frac{d}{d\lambda_i}\kernel(\lambda_1,\lambda_2,\lambda_3)\),
	we assume the partial derivatives
	\begin{equation}
		\label{eq: necessary derivatives}
		\kernel_{12},\quad \kernel_{13},\quad \kernel_{23} \quad\text{and}\quad \kernel_{33}
	\end{equation}
	of the kernel \(\kernel\) exist and are continuous. Furthermore, we assume 
	the derivative of the mean \(\mu\) exists and is continuous.
\end{assumption}
In Equation~\eqref{eq: how to calculate the covariance of derivatives} we can see
	that the covariance of derivatives is directly related to the derivatives
	of the covariance function. For \(\nabla\rf_\dims\) to exist, it is therefore natural
	that \(\C_{\rf_\dims}\) has to be differentiable. The covariance has to be two times
	differentiable in a sense, as we require the following to be well defined
	\[
		\Cov(\partial_i \rf_\dims(x), \partial_j\rf_\dims(y)) = \partial_{x_i}\partial_{x_j}\C_{\rf_\dims}(x,y).
	\]
	In the case of (non-stationary) isotropic functions, this requires the existence
	of \eqref{eq: necessary derivatives} by Lemma~\ref{lem: cov of derivatives, non-stationary isotropy}.
	And the existence of the terms in \eqref{eq: necessary derivatives} 
	is in fact necessary and sufficient for \(\nabla\rf_\dims\)
	to exist in an \(L^2\)-sense \citep[e.g.][Ch.~IV, \S 3, Thm.~4]{gihmanTheoryStochasticProcesses1974}.
	Our additional assumption of continuity is slightly weaker than typical
	sufficient conditions for a point-wise defined, continuous version of
	\(\nabla\rf_\dims\) to exist \citep[e.g.][]{adlerRandomFieldsGeometry2007}.
	Intuitively, we therefore simply assume \(\nabla\rf_\dims\) to be continuous
	almost surely.\smallskip

Here is our main result:
\begin{theorem}[Predictable progress, general case]
	\label{thm: asymptotically deterministic behavior}
	Assume \(\kernel\) defines a kernel valid in all dimensions
	(Definition~\ref{def: valid in all dimensions}) and let \(\rf_\dims \sim \normal(\mu, \kernel)\) be a sequence of scaled (non-stationary) isotropic Gaussian random
	functions (Definition~\ref{def: isotropic gaussian random
	function}) in $\dims$. Assume that \(\mu\) and \(\kernel\)  are sufficiently smooth
	(Assumption~\ref{assmpt: smoothness}) and that the
	covariance of \((\rf_\dims,\nabla\rf_\dims)\) is strictly positive definite
	(Definition~\ref{def: strict positive definite random function}). Let \(\gsa\) be a general gradient span algorithm (Definition~\ref{def:
	general gsa}), where we assume that its prefactors
	\(\lr_\timestep=\lr_\timestep(\info_{\timestep-1})\) are continuous in the
	information \(\info_{\timestep-1}\), and utilize the most recent gradients, i.e.
	\(\lr^{(g)}_{\timestep,\timestep-1} \neq 0\) for all \(\timestep\in \nat\).
	Let the algorithm \(\gsa\) be applied to \(\rf_\dims\) with independent, possibly
	random starting point \(\Param_0\in \real^\dims\) and denote by
	\(\Param_\timestep\) the points resulting from \(\gsa\). Finally, assume that
	\(\|\Param_0\|=\radius\) almost surely.  Then, for any \(\timestep\in
	\nat\), the random information vector
	\[
		\Info_\timestep
		:= \Bigl(\rf_\dims(\Param_k): k \le \timestep\Bigr)
		\cup \Bigl(
			\langle v, w\rangle :
			v,w \in (\param_0) \cup \gradients_\timestep
		\Bigr)
		\quad\text{with}\quad
		\gradients_\timestep := \bigl(\nabla \rf_\dims(\Param_k): k \le \timestep\bigr).
	\]
	converges, as the dimension increases \((\dims\to\infty)\), in probability against a deterministic information vector
	\(\liminfo_\timestep=\liminfo_\timestep(\gsa, \mu, \kernel, \lambda)\).
\end{theorem}

An application of the stochastic triangle inequality yields the following
corollary, which perfectly describes Figure~\ref{fig: loss plots}.

\begin{corollary}[Asymptotically identical progress over initializations]
	\label{cor: asymptotically identical cost curves}
	Assume the setting of Theorem~\ref{thm: asymptotically deterministic
	behavior} and let \(\Param_0^{(1)}\), \(\Param_0^{(2)}\) be random initialization
	points selected independently from \(\rf_\dims\). 
	In the non-stationary case, we additionally assume that we we have almost
	surely \(\|\Param_0^{(1)}\| = \|\Param_0^{(2)}\| = \radius\in \real\). Let the sequences
	\((\Param_\timestep^{(i)})_{\timestep\in\nat}\) be generated from
	the initialization \(\Param_0^{(i)}\) by the same general gradient
	span algorithm \(\gsa\).
	Then we have, for all \(\timestep\in \nat\) and all \(\epsilon>0\),
	\[
		\lim_{\dims\to\infty}\Pr\Bigl(
			\max_{k\le \timestep}\Bigl|
				\rf_\dims(\Param_k^{(1)})-\rf_\dims(\Param_k^{(2)})
			\Bigr| > \epsilon
		\Bigr) = 0.
	\]
\end{corollary}
\begin{proof}
	The proof is essentially an application of the stochastic triangle inequality
	\[
		\Bigl\{ |X - Z| > \epsilon\Bigr\}
		\subseteq 
		\Bigl\{ |X - Y| > \frac{\epsilon}2\Bigr\}
		\cup \Bigl\{ |Y - Z| > \frac{\epsilon}2\Bigr\},
	\]
	which yields by Theorem~\ref{thm: asymptotically deterministic behavior}
	\begin{align*}
		&\lim_{\dims\to\infty}\Pr\Bigl(
			\max_{k\le \timestep}\Bigl|
				\rf_\dims(\Param_k^{(1)})-\rf_\dims(\Param_k^{(2)})
			\Bigr| > \epsilon
		\Bigr)
		\\
		&\le \lim_{\dims\to\infty}\sum_{k\le \timestep}
		\Bigl[\Pr\Bigl(
			\Bigl|
				\rf_\dims(\Param_k^{(1)})-\limf_k
			\Bigr| > \frac{\epsilon}2
		\Bigr)
		+ \Pr\Bigl(
			\Bigl|
				\rf_\dims(\Param_k^{(2)})-\limf_k
			\Bigr| > \frac{\epsilon}2
		\Bigr)
		\Bigr]
		\\
		&=0.
	\end{align*}
\end{proof}

Similar to \citet{paquetteHaltingTimePredictable2022}, we obtain so-called
asymptotically deterministic halting times as a corollary. 
The \(\epsilon\)-halting is defined as
\[
	T_\epsilon := \inf\{\timestep > 0: \|\nabla\rf_\dims(X_\timestep)\|^2 \le \epsilon\}.
\]
We are going to prove it is asymptotically equal to the
asymptotic \(\epsilon\)-halting time
\[
	\tau_\epsilon := \inf\{\timestep > 0: \limgdot_\timestep \le \epsilon\} \in \nat,
\]
where  \(\limgdot_\timestep := \limgdot_{\timestep,\timestep}\) is the stochastic limit of
\(\|\nabla\rf_\dims(X_\timestep)\|^2\) in the dimension.
In practical terms this means that optimization always stops at roughly
the same time in high dimension.

\begin{corollary}[Asymptotically deterministic halting times]
	If \(\epsilon\notin (\limgdot_\timestep)_{\timestep\in\nat}\), then
	the halting time is asymptotically deterministic
	\[
		\lim_{\dims\to\infty} \Pr(T_\epsilon = \tau_\epsilon) = 1.
	\]
	If \(\epsilon= \limgdot_\timestep\) for some \(\timestep\), then
	\[
		\lim_{\dims\to\infty} \Pr(T_\epsilon \in [\tau_\epsilon, \tau^+_\epsilon]) = 1
		\quad \text{with}\quad
		\tau^+_\epsilon := \inf\{\timestep > 0 : \limgdot_\timestep < \epsilon\}.
	\]
\end{corollary}
\begin{proof}
	The proof is identical to the proof of Theorem~4 of \citet{paquetteHaltingTimePredictable2022}.
\end{proof}

\begin{remark}[Explicit evolution formula]
	\label{rem: explicit evolution formula}
	In the proof of Theorem~\ref{thm: asymptotically deterministic behavior} we
	construct explicit formulas for the conditional distribution of \((\rf_\dims(X_n),
	\nabla\rf_\dims(X_n))\). This remark summarizes these formulas.
	Due to its terse nature as a summary it may be difficult to read without
	the context of the proof and may be more helpful as an overview after the reader
	has familiarized themselves with the proof. In this summary we also ignore
	mathematical conditioning challenges.

	Let \(\rf_\dims \sim \normal(\mu,
	\kernel)\), let the filtration be given by
	\[
		\filt_n \coloneq \sigma\Bigl(
			\Param_0, \rf_\dims(X_k), \nabla\rf_\dims(X_k) : k \le n
		\Bigr)
	\]
	and let \(\rv_{[0\inter \dimV_n)}=(\rv_0, \dots, \rv_{\dimV_n-1})\) (notation defined in \eqref{eq: discrete range notation})  be the orthonormal
	basis of
	\[
		V_n \coloneq \Span\Bigl\{ \param_0, \nabla\rf_\dims(X_k) : k \mathrel{\red{<}} n\Bigr\}
	\]
	obtained using the Gram-Schmidt process and \(\rw_{[\dimV_n \inter \dims)}\)
	any orthonormal basis of \(V_n^\perp\). Then for \(Y_0, \dots, Y_\dims \overset{\iid}\sim \normal(0,1)\)
	independent of \(\filt_{n-1}\) the following equality holds in distribution:
	\vspace{1em}
	\begin{align}
		\left[
		\begin{matrix}
		\begin{pmatrix}
			\rf_\dims(X_n) \\
			D_{\rv_0}\rf_\dims(X_n)
			\\
			\vdots \\
			D_{\rv_{\dimV_n-1}}\rf_\dims(X_n)
		\end{pmatrix}
		\\
		\begin{pmatrix}
		D_{\rw_{\dimV_n}}\rf_\dims(X_n)
		\\
		\vdots
		\\
		D_{\rw_{\dims-1}}\rf_\dims(X_n)
		\end{pmatrix}
		\end{matrix}
		\middle| \filt_{n-1}
		\right]
		\overset{d}&= \begin{bmatrix}
			\mathbf{m}_n
			+ \vphantom{\begin{pmatrix}
				Y_0 \\
				Y_1\\
				\vdots \\
				Y_{\dimV_\timestep}
			\end{pmatrix}}
			\smash{\overbrace{\tfrac1{\sqrt{\dims}} [\mathbf{C}_n]^{\frac12} \begin{pmatrix}
				Y_0 \\
				Y_1\\
				\vdots \\
				Y_{\dimV_\timestep}
			\end{pmatrix}}^{\to 0\mathrlap{\quad (\dimV_n \le n \ll \dims)}},
			}
			\\
			\frac{\sigma_n}{\sqrt{\dims}}\begin{pmatrix}
				Y_{\dimV_n} \\
				\vdots \\
				Y_{\dims-1}
			\end{pmatrix} 
		\end{bmatrix}
	\end{align}
	The terms \(\mathbf{m}_n\) and \(\mathbf{C}_n\) and \(\sigma_n^2\) all
	converge as \(N\to\infty\). Since there are only finitely many \((Y_0, \dots, Y_{\dimV_n})\),
	the term around \(\mathbf{C}_n\) becomes asymptotically irrelevant due to scaling.
	On the contrary \(\sigma_n\) remains important because there are \(\dims-\dimV_n\) many
	\(Y_i\) left such that their squared sum is of constant order.
	By \eqref{eq: residual variance} we have that this `residual variance' is given by
	\[
		\sigma_n^2
		\coloneq 
			\green{\kernel_3\bigl(\tfrac{\|\Param_\timestep\|^2}2,\tfrac{\|\Param_\timestep\|^2}2, \|\Param_\timestep\|^2\bigr)}
			- (\blue{\rcov^{\black{w},\dims}_{[0:\timestep),\timestep}})^T
			[\magenta{\rcov^{\black{w},\dims}_{[0:\timestep)}}]^{-1}
			\blue{\rcov^{\black{w},\dims}_{[0:\timestep),\timestep}} \in \real,
	\]
	with \(\kernel_m(\lambda_1,\lambda_2,\lambda_3) := \frac{d}{d\lambda_m}\kernel(\lambda_1,\lambda_2,\lambda_3)\) and
	\begin{align*}
		\blue{\rcov^{\black{w},\dims}_{[0:\timestep),\timestep}}
		&\coloneq \begin{pmatrix}
			\kernel_3\bigl(\tfrac{\|\Param_0\|^2}2,\tfrac{\|\Param_\timestep\|^2}2, \langle \Param_0, \Param_\timestep\rangle\bigr)
			\\
			\vdots
			\\
			\kernel_3\bigl(\tfrac{\|\Param_{n-1}\|^2}2,\tfrac{\|\Param_\timestep\|^2}2, \langle \Param_{n-1}, \Param_\timestep\rangle\bigr)
		\end{pmatrix}
		\\
		\magenta{\rcov^{\black{w},\dims}_{[0:\timestep)}}
		&\coloneq \begin{pmatrix}
			\kernel_{3}\bigl(\tfrac{\|\Param_0\|^2}2,\tfrac{\|\Param_0\|^2}2, \|\Param_0\|^2\bigr) & \cdots & \kernel_{3}\bigl(\tfrac{\|\Param_0\|^2}2,\tfrac{\|\Param_{n-1}\|^2}2, \langle \Param_0, \Param_{n-1}\rangle\bigr)
			\\
			\vdots & \ddots & \vdots
			\\
			\kernel_{3}\bigl(\tfrac{\|\Param_{n-1}\|^2}2,\tfrac{\|\Param_0\|^2}2, \langle \Param_{n-1}, \Param_0\rangle\bigr) & \cdots & \kernel_{3}\bigl(\tfrac{\|\Param_{n-1}\|^2}2,\tfrac{\|\Param_{n-1}\|^2}2, \|\Param_{n-1}\|^2\bigr)
		\end{pmatrix}.
	\end{align*}
	Note that all the \(\langle X_i, X_j\rangle\) terms converge \ref{ind:
	representation} as they are made up of components from the information
	vector. Consequently all the terms above converge in \(N\) such that
	\(\sigma_n^2\) converges in \(N\).
	Moreover
	\[
		\mathbf{m}_n = \begin{pmatrix}
			\mu\bigl(\tfrac{\|\Param_\timestep\|^2}2\bigr) \\
			\mu'\bigl(\tfrac{\|\Param_\timestep\|^2}2\bigr) \langle \Param_\timestep, \rv_0\bigl\rangle
			\\
			\vdots \\
			\mu'\bigl(\tfrac{\|\Param_\timestep\|^2}2\bigr) \langle \Param_\timestep, \rv_{\dimV_n-1}\bigl\rangle
		\end{pmatrix}
    +\blue{\rcov^{\black{v},\dims}_{[0:\timestep),\timestep}}^\transpose
    [\magenta{\rcov^{\black{v},\dims}_{[0:\timestep)}}]^{-1}
    \bigl(Z(\rv_{[0:\dimV_\timestep)}; \Param_{[0:\timestep)})-\mu_{[0:\timestep)}^{(v,\dims)}\bigr).
	\]
	where the following row-major matrices are flattened into column vectors
	\begin{align}
		\label{eq: state}
		Z(\rv_{[0:\dimV_\timestep)}; \Param_{[0:\timestep)})
		&= \mathrm{vec}\begin{pmatrix}
			\rf_\dims(\Param_0) & \cdots & \rf_\dims(\Param_{\timestep-1}) 
			\\
			D_{\rv_0}\rf_\dims(\Param_0) & \cdots & D_{\rv_0}\rf_\dims(\Param_{\timestep-1})
			\\
			\vdots & & \vdots
			\\
			D_{\rv_{\dimV_\timestep-1}}\rf_\dims(\Param_0) & \cdots & D_{\rv_{\dimV_\timestep-1}}\rf_\dims(\Param_{\timestep-1})
		\end{pmatrix}	
		\\
		\nonumber
		\mu_{[0:\timestep)}^{(v,\dims)}
		&= \mathrm{vec}\begin{pmatrix}
			\mu\bigl(\tfrac{\|\Param_0\|^2}2\bigr) & \cdots & \mu\bigl(\tfrac{\|\Param_{\timestep-1}\|^2}2\bigr) 
			\\
			\mu'\bigl(\tfrac{\|\Param_0\|^2}2\bigr) \langle \Param_0, \rv_0\bigl\rangle & \cdots & \mu'\bigl(\tfrac{\|\Param_{\timestep-1}\|^2}2\bigr) \langle \Param_{\timestep-1}, \rv_0\bigl\rangle
			\\
			\vdots & & \vdots
			\\
			\mu'\bigl(\tfrac{\|\Param_0\|^2}2\bigr) \langle \Param_0, \rv_{\dimV_\timestep-1}\bigl\rangle & \cdots & \mu'\bigl(\tfrac{\|\Param_{\timestep-1}\|^2}2\bigr) \langle \Param_{\timestep-1}, \rv_{\dimV_\timestep-1}\bigl\rangle
		\end{pmatrix}.
	\end{align}
	With \(D_\emptyset \rf_\dims = \rf_\dims\) observe that the matrices above
	may be reindexed by \(\{\emptyset, \rv_0, \dots, \rv_{\dimV_\timestep-1}\}\)
	and \(\{X_0,\dots, X_{\timestep-1}\}\) instead of \(\{0, \dots, \dimV_\timestep\}\) and
	\(\{0,\dots, n-1\}\). This makes it easier to state the corresponding
	covariance matrices
	\[\begin{aligned}
		[\magenta{\rcov^{\black{v},\dims}_{[0:\timestep)}}]_{(v,x, w, y)}
		&= \widetilde{\Cov}(D_v \rf_\dims(x), D_w \rf_\dims(y))
		\qquad v, w \in \{\emptyset, \rv_0, \dots, \rv_{\dimV_\timestep-1}\}, x,y\in X_{[0:\timestep)}
		\\
		[\blue{\rcov^{\black{v},\dims}_{[0:\timestep),\timestep}}]_{(v, x, w)}
		&= \widetilde{\Cov}(D_v \rf_\dims(x), D_w \rf_\dims(X_\timestep))
	\end{aligned}
	\]
	where \(\widetilde{\Cov}\) treats \(\rv_i\) and \(X_i\) like deterministic
	variables (formal details in the proof of Theorem~\ref{thm:
	asymptotically deterministic behavior}). These are flattened to match the flattened
	vectors above. Finally,
	\[
		\mathbf{C}_n = 
		\green{\rcov^{\black{v},\dims}_\timestep}
        - \blue{\rcov^{\black{v},\dims}_{[0:\timestep),\timestep}}^T
        [\magenta{\rcov^{\black{v},\dims}_{[0:\timestep)}}]^{-1}
        \blue{\rcov^{\black{v},\dims}_{[0:\timestep),\timestep}},
	\]
	where \(\green{\rcov^{\black{v},\dims}_\timestep}\) is the ``covariance''
	matrix of \((\rf_\dims(X_n), D_{\rv_0}\rf_\dims(X_n), \dots,
	D_{\rv_{\dimV_n-1}}\rf_\dims(X_n))\) treating \(\rv_i\) and \(X_n\) like
	deterministic variables, i.e.
	\[
		\green{\rcov^{\black{v},\dims}_\timestep}
		= \begin{pmatrix}
			\widetilde{\Cov}(\rf_\dims(X_\timestep), \rf_\dims(X_\timestep)) & \cdots & \widetilde{\Cov}(\rf_\dims(X_\timestep), D_{\rv_{\dimV_\timestep-1}}\rf_\dims(X_\timestep))
\\
			\vdots & & \vdots
			\\
			\widetilde{\Cov}(D_{\rv_{\dimV_\timestep-1}}\rf_\dims(X_\timestep), \rf_\dims(X_\timestep)) & \cdots & \widetilde{\Cov}(D_{\rv_{\dimV_\timestep-1}}\rf_\dims(X_\timestep), D_{\rv_{\dimV_\timestep-1}}\rf_\dims(X_\timestep))
		\end{pmatrix}.
	\]
\end{remark}

\begin{remark}[Computationally tracking the evolution]
	\label{rem: computationally tracking the evolution}
	Observe that the state space is essentially given by \eqref{eq: state},
	which contains \(n(\dimV_n+1) \sim n^2\) random variables. Its covariance matrix
	\(\magenta{\rcov^{\black{v},\dims}_{[0:\timestep)}}\)
	consequently has \(\bigO(n^4)\) entries and may be decomposed and inverted in
	\(\bigO(n^6)\) time. Since
	\(\magenta{\rcov^{\black{v},\dims}_{[0:\timestep)}}\)
	contains the entries of the previous covariance matrices, the Cholesky
	decomposition can be updated such that the total complexity over all
	iterations up to \(n\) remains at \(\bigO(n^6)\) instead of growing to
	\(\bigO(n^7)\).
	To avoid the complexity of \(\bigO(n^6 + \dims n^2)\) when \(\dims\) is really
	large it is important to never compute the basis vectors \(\rv_i\). They are
	simply an orthonormal basis and it suffices to compute the gradients and \(X_i\)
	in this coordinate system. Observe that by definition of \(\rv_{\dimV_n}\)
	\[
		D_{\rv_{\dimV_n}} \rf_\dims(X_n)
		= \Biggl\| \begin{pmatrix}
			D_{\rw_{\dimV_n}}\rf_\dims(X_n)
			\\
			\vdots \\
			D_{\rw_{\dims-1}}\rf_\dims(X_n)
		\end{pmatrix}
		\Biggr\|
		\overset{d}= \frac{\sigma_n}{\sqrt{\dims}} \sqrt{\sum_{i=\dimV_n}^{\dims-1} Y_i^2}
		\sim \sigma_n \frac{\chi_{\dims-\dimV_n}}{\sqrt{\dims}} \to \sigma_n,
	\]
	where \(\chi_k\) is the chi distribution with \(k\) degrees of
	freedom. The remaining components of \(\nabla \rf_\dims(X_n)\) are
	already expressed in terms of the basis \(\rv_i\).
	
	The implementation used to generate Figure~\ref{fig: simulated gradient
	descent} can be found at \url{https://github.com/FelixBenning/pyGRF} and we
	plan to publish a detailed explanation separately.
\end{remark}

 \section{A discussion of the dimensional scaling}
\label{sec: discuss scaling}

Readers unfamiliar with the \(\frac1\dims\) scaling of the covariance in
\eqref{eq: canonical parametrization} might hypothesize that this reduction of
the variance simply
collapses the random function \(\rf_\dims\) to
the mean \(\mu\) in the asymptotic limit. Since asymptotically deterministic
behavior would then follow trivially, it is important to understand why this is
not the case.

To simplify the argument consider a stationary isotropic GRF, i.e.
\(\rf_\dims\sim\normal(\mu,\ikernel)\), the arguments work similarly for any (non-stationary) isotropic GRF. For any \emph{fixed} parameter \(x\) we have \(\rf_\dims(\param)\sim\normal(\mu, \frac1\dims\ikernel(0))\),
so the function value \(\rf_\dims(\param)\) indeed collapses exponentially fast to the mean
by a standard Chernoff-bound
\begin{equation}
    \label{eq: Chernoff bound}
    \Pr\bigl(|\rf_\dims(x) - \mu| \ge t\bigr)
    \le 2\exp\bigl(-\dims\tfrac{t}{2\ikernel(0)}\bigr).
\end{equation}
While the function value \(\rf_\dims(\param)\) collapses to the mean, we will
proceed to show that the gradient \(\nabla\rf_\dims(\param)\) does not.
And if the gradient does not collapse, it is sufficient to follow the gradient
to find points where \(\rf_\dims\) stays away from \(\mu\). So the function \(\rf_\dims\)
does not \emph{uniformly} collapse to the mean.

First, let us sketch how the the derivatives of random functions are related to
the derivatives of the covariance, see for instance \citet[Sec.~1.4.2]{adlerRandomFieldsGeometry2007}
for a formal derivation. Assuming the mean to be zero, without
loss of generality, we move the derivatives outside of the integral
to obtain
\[
    \Cov(\partial_{x_i} \rf_\dims(x), \rf_\dims(y))
    = \partial_{x_i} \E[\rf_\dims(x)\rf_\dims(y)]
    = \partial_{x_i} \C_{\rf_\dims}(x,y).
\]
Iterating on this idea yields the covariance of derivatives
\begin{equation}
	\label{eq: how to calculate the covariance of derivatives}	
    \Cov(\partial_{x_i} \rf_\dims(x), \partial_{y_j}\rf_\dims(y))
    = \partial_{x_i}\partial_{y_j} \E[\rf_\dims(x)\rf_\dims(y)]
    = \partial_{x_i}\partial_{y_j} \C_{\rf_\dims}(x,y).
\end{equation}
In the case of a stationary isotropic covariance \(\C_{\rf_\dims}(x,y) =
\frac1\dims \ikernel\bigl(\frac{\|x-y\|^2}2\bigr)\) this implies 
\[
    \partial_{x_i}\partial_{y_j} \C_{\rf_\dims}(x,y)
    = \tfrac{1}{\dims}\Bigl[
        \underbrace{\ikernel''\bigl(\tfrac{\|x-y\|^2}2\bigr)(x_i-y_i)(y_j-x_j)}_{\text{(I)}}
        - \underbrace{\ikernel'\bigl(\tfrac{\|x-y\|^2}2\bigr) \delta_{ij}}_{\text{(II)}}
    \Bigr],
\]
where \(\delta_{ij}\) is the Kronecker delta. Part (I) is zero if \(x=y\).
The Kronecker delta in part (II) then implies that the entries of \(\nabla\rf(x)\)
are uncorrelated and therefore independent by the
Gaussian assumption. Specifically, we have
\[
    \partial_{x_i} \rf_\dims(x)
    \overset{\iid}\sim \normal\bigl(0, -\tfrac1\dims\ikernel'(0)\bigr).
\]
The gradient norm therefore experiences a law of large numbers
\begin{equation}
	\label{eq: slope}
    \|\nabla \rf_\dims(x)\|^2
    = \sum_{i=1}^\dims (\partial_{\param_i}\rf_\dims(x))^2
    \underset{\dims\to\infty}{\overset{p}\to} -\ikernel'(0).
\end{equation}
Thus, while the function value \(\rf_\dims(x)\) collapses to the mean exponentially
fast \eqref{eq: Chernoff bound}, the slope of the function
(counterintuitively) does not. Any other type of scaling would result in vanishing or exploding gradients. Since the supremum over gradient norms is exactly the Lipschitz constant of
\(\rf_\dims\), this scaling is therefore necessary to stay in a class of Lipschitz
functions, as is often assumed in optimization theory. And the Parisi formula \citep{parisiSequenceApproximatedSolutions1980,talagrandParisiFormula2006,panchenkoSherringtonKirkpatrickModel2013}
shows in the case of spin glasses, that this scaling also stabilizes the global
maximum/minimum.

\begin{remark}[Isoperimetry]
    In view of the counterintuitive observations above, one might ask 
    whether the assumption of isotropy results in very peculiar Lipschitz
    functions.  To understand why that is not the case, consider the concept of
    isoperimetry \citep[e.g.][]{bubeckUniversalLawRobustness2021}. A random
    variable \(X\) on \(\real^\dims\) is said to satisfy c-isoperimetry, if
    for any \(L\)-Lipschitz function \(f:\real^\dims\to\real\) and any \(t\ge
    0\)	an exponential concentration bound holds
    \[
        \Pr\Bigl(|f(X) - \E[f(X)]| \ge t\Bigr)
        \le 2 \exp\bigl(-\dims\tfrac{ t^2}{2cL}\bigr).
    \]
    Observe that this concentration bound is the mirror image of \eqref{eq: Chernoff bound}.
    While we consider random functions, isoperimetry is concerned with random input \(X\) to
    deterministic \(L\)-Lipschitz functions. In particular Gaussian or uniform input
    satisfies isoperimetry \citep{bubeckUniversalLawRobustness2021}. Intuitively, this paints the following picture of very high dimensional
    Lipschitz functions:
    Most of the function is equal to the mean except for a few peculiar points.
    In the case of a deterministic functions this requires the exclusion of specific
    deterministic points. In the case of random functions this implies any deterministic point is allowed
    but not the use of gradient information. In either case the point we pick must be `independent' 
    of the function. And we show in Theorem~\ref{thm: asymptotically deterministic
    behavior} that if both function and input is random, independence is
    essentially a sufficient condition.
\end{remark}

Finally, note that our definition of scaled sequences of random functions can be
reframed in terms of a single random function defined on \(\real^\infty\), which
is externally scaled.

\begin{remark}[External scaling]\label{remark:scaling}

One could define a a (non-stationary)
isotropic GRF \(\rf\) with covariance
\[
	\C_\rf(x,y) = \kernel\bigl(\tfrac{\|x\|^2}2,\tfrac{\|y\|^2}2, \langle x,y\rangle\bigr),
	\qquad x,y\in \real^\infty
\]
on the space of eventually-zero
sequences \(\real^\infty\), which can also be viewed as the union of
\(\real^\dims\)
\[
	\real^\infty := \bigcup_{\dims\in \nat}\real^\dims
	\quad\text{with}\quad
	\real^\dims := \{ (\param_i)_{i\in \nat}\subset\real : \param_i = 0 \quad \forall i > N\}.
\]
Since the scaling \(\frac1{\sqrt{\dims}}\) would scale away any mean
the function \(\rf\) might possess, we assume \(\rf\) to be \emph{centered} and
define
\[
	\rf_\dims(\param)
	:= \mu\bigl(\tfrac{\|\param\|^2}2\bigr) + \tfrac1{\sqrt{\dims}}\rf(\param),
	\quad \param\in\real^\dims.
\]
The distribution of \(\rf_\dims\) is then equivalent to the one in
Definition~\ref{def: isotropic gaussian random function}.
\end{remark}

 \section{Proof of Theorem~\ref{thm: asymptotically deterministic behavior}}\label{sec:proof}
We now present the proofs of our main result. First, we provide a sketch
to motivate the overall picture (Section~\ref{sec:sketch}). After we explain how the covariance of derivatives are determined
(Section~\ref{sec: covariance of derivatives}), we reforge
Theorem~\ref{thm: asymptotically deterministic behavior} into an even more general
Theorem~\ref{thm: asymptotically deterministic behavior variant}.
This theorem is more natural to prove but requires the definition of a special
orthonormal coordinate system (Definition~\ref{def: previsible orthonormal
coordinate system}).  Finally, we give the main proof of 
Theorem~\ref{thm: asymptotically deterministic behavior variant}. Proofs for some instrumental results are deferred to the
appendix.

\subsection{Sketch of the proof of Theorem~\ref{thm: asymptotically deterministic behavior}}
\label{sec:sketch}

To better guide the reader, we first sketch the idea behind the proof. For
simplicity, we will assume the random function to be centered and stationary isotropic
\(\rf_\dims \sim \normal(0, \ikernel)\).

\subsubsection*{Idea~1: Induction over Gaussian Conditionals}

We interpret the objects of interest
\(\rg = (\rf_\dims, \nabla \rf_\dims)\) (the heights joined with the gradients, sometimes
called the `jet') as an auxiliary discrete-time stochastic process
\((\rg(\Param_\timestep))_{\timestep\in\nat_0}\) and
prove the claimed convergence in \(\dims\) by induction over the steps \(\timestep\).
Please pretend for now that the inputs \(\Param_\timestep\) were deterministic \(\param_\timestep\).
We will address this problem in Idea~3 of the sketch. The process
\((\rg(\param_\timestep))_{\timestep\in \nat_0}\) is then a Gaussian process and
\[
    \rg(\param_{[0:\timestep]}) := (\rg(\param_0), \dots, \rg(\param_\timestep))
\]
is therefore a multivariate Gaussian vector. Here \(\rg\) is
applied entry-wise, i.e.\ \(\rg(x_I) = (\rg(x_i))_{i\in I}\)
with \(x_I=(x_i)_{i\in I}\), and we use the following notation for discrete ranges:
\begin{equation}
    \label{eq: discrete range notation}
    [n\inter m] := [n,m] \cap \integer, \qquad
    [n\inter m) := [n,m) \cap \integer,
    \qquad \text{etc.}
\end{equation}
The reader is encourage to remember them, as we will make use of them throughout the following sections. It is well known
that the conditional distribution \(\rg(\param_\timestep)\mid
\rg(\param_{[0:\timestep)})\) is then also normal distributed (cf.\
Theorem~\ref{thm: conditional gaussian distribution}).
By the induction hypothesis \(\rg(\param_{[0:\timestep)})\) is already converging in
probability to something deterministic, so it is perhaps
natural to decompose \(\rg(\param_\timestep)\) into
\begin{equation}
    \label{eq: the induction engine}
    \rg(\param_\timestep) = \E[\rg(\param_\timestep) \mid \rg(\param_{[0:\timestep)})]
    + \sqrt{\Cov[\rg(\param_\timestep)\mid \rg(\param_{[0:\timestep)})]}\begin{pmatrix}
        Y_0
        \\
        \vdots
        \\
        Y_{\dims}
    \end{pmatrix}
\end{equation}
for independent standard normal distributed \(Y_i\) independent of \(\rg(x_{[0:\timestep)})\).
The conditional expectation is then given (cf.\ Theorem~\ref{thm: conditional gaussian distribution}) by
\[
	\E[\rg(\param_\timestep) \mid \rg(\param_{[0:\timestep)})]
	= \blue{\Cov(\rg(\param_{[0:\timestep)}), \rg(\param_\timestep))}^T[\magenta{\Cov(\rg(\param_{[0:\timestep)}))}]^{-1} \rg(\param_{[0:\timestep)})
\]
and conditional variance by
\begin{align*}
	&\Cov[\rg(\param_\timestep) \mid \rg(\param_{[0:\timestep)})]
	\\
	&= \green{\Cov[\rg(\param_\timestep)]} - \blue{\Cov(\rg(\param_{[0:\timestep)}), \rg(\param_\timestep))}^T[\magenta{\Cov(\rg(\param_{[0:\timestep)}))}]^{-1}\blue{\Cov(\rg(\param_{[0:\timestep)}), \rg(\param_\timestep))}.
\end{align*}
The vector \(\rg(x_{[0:\timestep)})\) already converges by induction as the dimension
\(\dims\) increases. What is therefore left to study are the covariance matrices.

\subsubsection*{Idea~2: Finding sparsity in the covariance matrices with a custom coordinate system}

The covariance matrices which make up the conditional expectation and variance
are increasing in size as the dimension \(\dims\) grows, because the gradient
\(\nabla\rf_\dims\) contained in \(\rg\) increases in size. Additionally we somehow need
to group the independent \(Y_i\) into something which converges.
It turns out that the standard coordinate system is ill suited to achieve these
goals. To get sufficiently sparse (and therefore manageable) matrices
we will now phrase everything in terms of directional derivatives.

In Lemma~\ref{lem: cov of derivatives, stationary isotropy} we show how the covariances of the
directional derivatives of an isotropic \(\rf_\dims\sim\normal(0, \ikernel)\) can be
calculated explicitly. In particular we have for vectors \(v,w\in \real^\dims\) and points
\(x,y\in \real^\dims\) with distance \(\distance = x-y\) 
\begin{equation}
    \label{eq: the two sparsity requirements}
    \Cov(D_v\rf_\dims(x), D_w\rf_\dims(y))
    = - \tfrac1\dims\Bigl[
        \underbrace{\ikernel''\bigl(\tfrac{\|\distance\|^2}2\bigr)\langle \distance, w\rangle\langle \distance, v\rangle}_{\text{(I)}}
        + \underbrace{\ikernel'\bigl(\tfrac{\|\distance\|^2}2\bigr)\langle w, v\rangle}_{\text{(II)}}
    \Bigr].
\end{equation}
We begin by explaining how an orthogonal basis keeps (II) sparse before we explain
where the use of the standard basis breaks down for (I).
For simplicity, let us forget the height $\rf_\dims$ for now and pretend \(\rg\) consists only of the
gradient, i.e. \(\rg=\nabla\rf_\dims\). Consider the covariance matrix at a fixed point 
\(\nabla\rf_\dims(x_0)\) only. Since we only consider the derivatives
at a single point we always have \(x= y = x_0\) so the distance \(\distance\)
vanishes. This completely removes the first part (I) (which will later require a
special basis). The second part (II) is zero if \(v\) and \(w\) are orthogonal.
Assuming we use an orthonormal coordinate system (e.g. the standard basis), we
therefore get
\[
    \green{\Cov[\rg(x_0)]}= \Cov[ \nabla\rf_\dims(x_0) ] = \frac1\dims\begin{pmatrix}
        -\ikernel'(0) &
        \\
        & \ddots
        \\
        & & -\ikernel'(0)
    \end{pmatrix}
    = \tfrac{-\ikernel'(0)}\dims \identity.
\]
As \(\rg(\param_{[0:0)})\) is the empty set we therefore have in the first step of the induction in $n$ 
(cf.~Equation~\eqref{eq: the induction engine})
\begin{align}
    \nonumber
    \rg(\param_0)
    &= \E[\rg(\param_0) \mid \rg(\param_{[0:0)})] 
    + \sqrt{\Cov[\rg(\param_0)\mid \rg(\param_{[0:0)})]}\begin{pmatrix}
        Y_0
        \\
        \vdots
        \\
        Y_{\dims}
    \end{pmatrix}
    \\
    \label{eq: first law of large numbers}
    &= \underbrace{\E[\rg(\param_0)]}_{=0} + \sqrt{\green{\Cov[\rg(x_0)]}} \begin{pmatrix}
        Y_0
        \\
        \vdots
        \\
        Y_{\dims}
    \end{pmatrix}
    \\
    \nonumber
    &= \sqrt{\tfrac{-\ikernel'(0)}\dims}\begin{pmatrix}
        Y_0
        \\
        \vdots
        \\
        Y_{\dims}
    \end{pmatrix}.
\end{align}
In particular, we have by the law of large numbers,
\[
    \langle \nabla \rf_\dims(x_0), \nabla\rf_\dims(x_0) \rangle
    = \frac{-\ikernel'(0)}{\dims} \sum_{i=1}^\dims Y_i^2 \overset{p}{\underset{\dims\to \infty}\to} -\ikernel'(0).
\]
For the induction start and (II) there is therefore no need to change the coordinate system.
But for the induction step with \(\timestep>1\) the situation in (I) becomes more delicate.\smallskip

We now want to understand where the issues in (I) with the standard coordinate
system come form. Recall that we also want to determine
limiting values of \(\langle \nabla \rf_\dims(x_0), \nabla\rf_\dims(x_\red{1}) \rangle\).

\textbf{Problem:} Since \(x_1\) will be \(x_0\) plus a gradient step,
\(\distance=x_1 - x_0\) is therefore a multiple of the gradient. If we continue to use
the standard coordinate system, (I) causes the covariance matrix to be dense. This is because the
entries of \(\distance\) are proportional to \(\partial_i \rf_\dims(x_0)\), which are
multiples of the \(Y_i\) and therefore almost surely never zero.

\textbf{Solution:} We use an adapted coordinate system. For this we select
\(\tilde{\rv}_0 := \nabla \rf_\dims(\param_0)\) and normalize this vector \(\rv_0 =
\frac{\tilde{\rv}_0}{\|\tilde{\rv}_0\|}\). Again, we are going to pretend that this vector
was deterministic and mark this fact with the notation \(v_0\).
We then extend it by \(w_{[1:\dims)}\) to an orthonormal basis. Since the \(w_i\)
are orthogonal to the gradient span, they are orthogonal to the distances
\(\param_0 - \param_1\), i.e. \(\langle \distance, w_i\rangle = 0\) and therefore
(I) is zero for almost all directions (except for \(v_0\)). As the coordinate
system is orthonormal, (II) is sparse again. More generally, we chose a coordinate system
\(v_{[0:\timestep)}\) capturing the span of the first \(\timestep\) gradients \(\nabla\rf_\dims(\param_{[0:\timestep)})\) and
therefore also the span of the first \(\timestep\) parameters and extend this coordinate
system by \(w_{[\timestep:\dims)}\) to an orthonormal coordinate system such that
we have essentially
\[
    \rg(\param_{[0:\timestep)})
    = (\nabla\rf_\dims(\param_0),\dots, \nabla\rf_\dims(\param_{\timestep-1}))
    = U\begin{pmatrix}
        D_{v_0}\rf_\dims(\param_0) & \cdots & D_{v_0}\rf_\dims(\param_{\timestep-1})
        \\
        \vdots & & \vdots
        \\
        D_{v_{\timestep-1}}\rf_\dims(\param_0) & \cdots & D_{v_{\timestep-1}}\rf_\dims(\param_{\timestep-1})
        \\
        D_{w_\timestep}\rf_\dims(\param_0) & \cdots & D_{w_\timestep}\rf_\dims(\param_{\timestep-1})
        \\
        \vdots & & \vdots
        \\
        D_{w_{\dims-1}}\rf_\dims(\param_0) & \cdots & D_{w_{\dims-1}}\rf_\dims(\param_{\timestep-1})
    \end{pmatrix}U^T.
\]
Here \(U\) is a basis change matrix which we are going to suppress for simplicity in the following.
Note that the covariance matrix is dense for the vectors \(v_{[0:\timestep)}\) because they span the subspace
of the evaluation points \(x_k\) and therefore cause (I) to be non-zero. But (I)
remains zero for the vectors \(w_i\) and the covariance matrix is therefore sparse there.
We want to group the chaos together (i.e. group the directional
derivatives of \(v_{[0:\timestep)}\)), so we view \(\rg(\param_{[0:\timestep)})\)
as a row-major matrix, i.e.  whenever we treat \(\rg(\param_{[0:\timestep)})\) as a vector we
concatenate the rows. It then turns out that the covariance matrix is of the
form
\[
    \magenta{\Cov[\rg(\param_{[0:\timestep)})]}
    = \left(\begin{array}{c c c c c}
        \cline{1-1}
        \multicolumn{1}{|c|}{}
        \\
        \multicolumn{1}{|c|}{\Cov[D_{v_{[0:\timestep)}\rf_\dims(\param_{[0:\timestep)})}]} 
        \\
        \multicolumn{1}{|c|}{}
        \\
        \cline{1-2}
        &\multicolumn{1}{|c|}{\Sigma_\timestep}
        \\
        \cline{2-3}
        & &\multicolumn{1}{|c|}{\quad}
        \\
        \cline{3-3}
        & & & \ddots
        \\
        \cline{5-5}
        & & & &\multicolumn{1}{|c|}{\Sigma_{\dims-1}}
        \\
        \cline{5-5}
    \end{array}\right),
\]
where \(\Sigma_i = \Cov[D_{w_i}\rf_\dims(\param_{[0:\timestep)})]\) are \(\timestep\times \timestep\) matrices
and \(\Cov[D_{[0:\timestep)}\rf_\dims(\param_{[0:\timestep)})]\) is a dense \(\timestep^2\times \timestep^2\) block matrix.
We can assume \(\dims-1 > \timestep\) without loss of generality as we let \(\dims \to
\infty\).
In the proof we will make the observation that all the
\(\Sigma_i\) are in fact equal.
Note that the mixed covariance matrix \(\blue{\Cov(\rg(\param_{[0:\timestep)}),
\rg(\param_\timestep))}\) used in the conditional expectation and covariance have
similar block form and \(\green{\Cov[\rg(\param_\timestep)]}\)
is always diagonal.
The key is now to understand that
as \(\dims\) grows to infinity, the scaling \(\frac1\dims\) keeps the \(\timestep^2\times
\timestep^2\) chaos in check while we get a similar law of large numbers as in the first
step (cf.~\eqref{eq: first law of large numbers}) by grouping the identical
\(\Sigma_i\) to get the length of
the new gradient \(\nabla\rf_\dims(x_\timestep)\) projected to
the subspace of \(w_{[\timestep:\dims)}\) which is the orthogonal space of \(v_{[0:\timestep)}\).
This projection makes up the new coordinate \(v_\timestep\) and the induction continues.

\subsubsection*{Idea~3: Treating the random input with care}

At the beginning of the sketch we asked the reader to ignore the fact that the
evaluation points \((\Param_\timestep)_{\timestep\in \nat_0}\) are random themselves. Moreover,
we ignored that the adapted orthogonal basis \(\rv_{[0:\timestep)}\) was constructed from
random gradients \(\nabla\rf_\dims(\Param_k)\) which implies it is random as well.
Instead we denoted it by \(v_{[0:\timestep)}\) and pretended it was deterministic.
In the following, we sketch how to get rid of the randomness of
\(\Param_\timestep\) and the randomness of the  coordinate system \(\rv_{[0:\timestep)}\). It is
perhaps most obvious for the random coordinate system, that
\[
    D_{\rv_0}\rf_\dims(\param_0) = \langle \rv_0, \nabla\rf_\dims(x_0)\rangle = \|\nabla\rf_\dims(x_0)\|^2
\]
is certainly not Gaussian by construction (recall \(\rv_0 = \nabla\rf_\dims(x_0)\)).
Similarly, the random points \(\Param_\timestep\) will typically change the distribution
of \(\rf_\dims(\Param_\timestep)\) and taking (conditional)
expectations and covariances becomes quite delicate.

Intuitively, the key to solve this issue is to condition on all the previously seen
information (e.g.\ gradients), such that the evaluation location do become
`deterministic'. While this is already common practice in Bayesian optimization,
measure theoretically this is very delicate. But for the canonical conditional
Gaussian distribution (Theorem \ref{thm: conditional gaussian distribution}), this
has recently been proven to work \citep[Corollary 2.12]{benningMeasureTheoryConditionally2026}.
Essentially, if \((\Param_\timestep)_{\timestep \in \nat_0}\) is previsible,
i.e. \(\Param_{\timestep+1}\) is measurable with regard to the filtration
\(\filt_\timestep = \sigma(\rg(X_k): k \le \timestep)\) the following is true
\[
    \E[h(\rg(\Param_\timestep))\mid \filt_{\timestep-1}]
    = \Bigl(
        (x_{[0:\timestep]})\mapsto \E[h(\rg(x_\timestep)) \mid \rg(x_{[0:\timestep)})]
    \Bigr)(X_{[0:\timestep]}).
\]
Applying this lemma to the evaluations points
\(\Param_\timestep\) is relatively straightforward. But it is not obvious how to treat the
random coordinate system. The key is to turn the coordinate system into an
input. The definition
\[
    Z(v; \param) := D_v \rf_\dims(\param)
\]
allows us to apply Corollary 2.12 from \citet{benningMeasureTheoryConditionally2026} to both coordinate
system and points. The delicate part is that the input must be previsible with
respect to the filtration \(\filt_\timestep\). This implies the coordinate system must also be
previsible! We therefore cannot use future gradients for our coordinate
system. This forces us to define a custom coordinate system for every step. That
means after \(\timestep\) steps we have \(\timestep\) different coordinate
systems. We will carefully reduce this to just one coordinate system per
iteration in the formal proof. Turning this strategy into a rigorous proof is a
bit tedious and requires a careful induction including a few additional
technical induction hypothesis.
 \subsection{Covariances of derivatives of (smooth) random functions}\label{sec: covariance of derivatives}

It should be clear from the sketch given in Section \ref{sec:sketch} that we
will have to deal with covariances of derivatives of the random functions under
consideration. In this section we use known results to calculate the covariance
of derivatives.
Swapping integration and differentiation, we have for a centered random function
\(\rf_\dims\)
\begin{align*}
	\Cov(\partial_{x_i} \rf_\dims(x), \rf_\dims(y))
	&= \E[\partial_{x_i} \rf_\dims(x) \rf_\dims(y)]
	= \partial_{x_i} \E[\rf_\dims(x)\rf_\dims(y)]
	\\
	&= \partial_{x_i} \C_{\rf_\dims}(x,y)
\end{align*}
so the covariance of a derivative of \(\rf_\dims\) with \(\rf_\dims\) is equal to a partial
derivative of the covariance function. For the formal details consult
\citet[chapter~5]{scheurerComparisonModelsMethods2009}, \citet[Ch.~IV, \S 3, Def.~3 and
following]{gihmanTheoryStochasticProcesses1974} or
\citet[Sec.~1.4.2]{adlerRandomFieldsGeometry2007}. Similarly, other covariances
can be calculated, e.g.
\[
	\Cov(\partial_{x_i}\rf_\dims(x), \partial_{y_i}\rf_\dims(y))
	= \partial_{x_i}\partial_{y_i} \C_{\rf_\dims}(x,y),
\]
and similarly the expectation of the derivative of uncentered random function
can be calculated \( \E[\partial_{x_i}\rf_\dims(x)] = \partial_{x_i}\E[\rf_\dims(x)] \).
For this reason the derivatives of the covariance function are interesting as they represent
the covariance of derivatives. Due to bilinearity of the covariance it is also
possible to move directional derivatives out of the covariance.\smallskip

As a warm-up, let us calculate the covariance of derivatives for isotropic random
functions \(\rf_\dims\), which have covariance functions of the form \(\C_{\rf_\dims}(x,y) =
\tfrac1\dims \ikernel\bigl(\tfrac{\|x-y\|^2}2\bigr).\)
\begin{lemma}[Covariance of derivatives, stationary isotropy]
    \label{lem: cov of derivatives, stationary isotropy}
    Let \(\rf_\dims\sim\normal(\mu, \ikernel)\) be an isotropic random function and let \(\distance = x-y\), then
    \begin{equation*}
    \def\arraystretch{1.5}
    \begin{tabular}{c | c c}
        \(\Cov\) & \(\rf_\dims(y)\) & \(D_w\rf_\dims(y)\) \\	
        \hline
        \(\rf_\dims(x)\)
        &  \(\tfrac1\dims\ikernel\bigl(\frac{\|\distance\|^2}{2}\bigr)\)
        & \(-\tfrac1\dims\ikernel'\bigl(\frac{\|\distance\|^2}2\bigr)\langle \distance, w\rangle\)
        \\
        \(D_v\rf_\dims(x)\)
        &  \(\tfrac1\dims\ikernel'\bigl(\frac{\|\distance\|^2}{2}\bigr)\langle \distance, v\rangle\)
        & \(
            - \tfrac1\dims\Bigl[
                \underbrace{\ikernel''\bigl(\tfrac{\|\distance\|^2}2\bigr)\langle \distance, w\rangle\langle \distance, v\rangle}_{
                    \text{(I)}
                }
                + \underbrace{\ikernel'\bigl(\tfrac{\|\distance\|^2}2\bigr)\langle w, v\rangle}_{\text{(II)}}
            \Bigr]
        \)
    \end{tabular}
    \end{equation*}
\end{lemma}
\begin{proof}
    Straightforward application of chain and product rules with the considerations above. 
\end{proof}

Recall that we split the covariance of derivatives up into two parts (I) and (II)
in the proof sketch. We will now outline how the same can be done for (non-stationary) isotropic
kernels. Since the (non-stationary) isotropic kernel \(\kernel\) have multiple inputs, we
further define the notation
\[
    \kernel_i(\lambda_1,\lambda_2,\lambda_3)
    := \tfrac{d}{d\lambda_i} \kernel(\lambda_1, \lambda_2, \lambda_3)
\]

\begin{restatable}[Covariance of derivatives, non-stationary isotropy]{lemma}{covOfDerivatives}
    \label{lem: cov of derivatives, non-stationary isotropy}
    Let \(\rf_\dims\sim\normal(\mu, \kernel)\) be (non-stationary) isotropic
    (Definition~\ref{def: isotropic gaussian random function}). Then the
    expectation of a directional derivative is given by
    \begin{equation}
        \label{eq: expect df}
        \E[D_v\rf_\dims(x)] = \mu'\bigl(\tfrac{\|x\|^2}2\bigr) \langle x, v\rangle,
    \end{equation}
    whereas, using the notation \(\kernel :=
    \kernel\bigl(\tfrac{\|x\|^2}2,\tfrac{\|y\|^2}2, \langle x,y\rangle \bigr)\)
    to omit inputs, the covariance of directional derivatives is given by
    \begin{align}
        \label{eq: cov df, f}
        \Cov(D_v \rf_\dims(x), \rf_\dims(y))   
        &= \frac1\dims\Bigl[\kernel_1 \langle x, v\rangle + \kernel_3 \langle y, v\rangle\Bigr]
        \\
        \label{eq: cov df, df}
        \Cov(D_v \rf_\dims(x), D_w\rf_\dims(y))   
        &= \frac1\dims\Bigl[\underbrace{\begin{aligned}[t]
            &\kernel_{12} \langle x, v\rangle\langle y, w\rangle + \kernel_{13} \langle x, v\rangle\langle x, w\rangle
            \\
            & + \kernel_{32} \langle y, v\rangle\langle y, w\rangle + \kernel_{33}\langle y, v\rangle\langle y, w\rangle
            \\
        \end{aligned}}_{\text{(I)}}
        + \underbrace{\kernel_3 \langle v, w\rangle}_{\text{(II)}}\Bigr].
    \end{align}
    In particular, if the directions \(v,w\) are orthogonal to the points
    \(x,y\) then \eqref{eq: cov df, f} and (I) of \eqref{eq: cov df, df} are zero.
    (II) in turn is zero, if the directions \(v,w\) are orthogonal.
\end{restatable}
\begin{proof}
    Straightforward application of chain and product rules.
\end{proof}
 
\subsection{Proof of Theorem~\ref{thm: asymptotically deterministic behavior}}\label{sec:main_proof}

After the preparations of the previous sections we can now give the details of
the proof sketched in Section \ref{sec:sketch}. In the sketch of the proof,
we highlighted the importance of an adapted coordinate system in order to
keep the covariance matrices sparse. In particular, part (I) of \eqref{eq: the
two sparsity requirements} requires orthogonality to the walking directions.
We also needed the coordinate system to be orthogonal to keep (II) as sparse as
possible. In Idea 3, we sketched why the coordinate system had to be previsible.
That is, why we needed a different coordinate system in every step.
We therefore begin the proof by building these coordinate systems.\smallskip

Let us start by defining the natural filtration
\[
    \filt_\timestep
    := \sigma\bigl( \rf_\dims(\Param_k), \nabla\rf_\dims(\Param_k) : 0\le k \le \timestep\bigr).
\]
The following previsible \emph{vector space of evaluation points}
\begin{equation}
    \label{eq: vector space of evaluation points}    
    V_\timestep
    := \Span\Bigl(
        \{\param_0\} \cup \{\nabla\rf_\dims(\Param_k) : 0\le k \mathrel{\red{<}}\timestep\}
    \Bigr),
    \qquad \dimV_\timestep := \dim(V_\timestep),
\end{equation}
contains \(\Param_{[0:\timestep]}\) by definition of the generalized gradient
span algorithm (Definition~\ref{def: general gsa}) and is similarly measurable
with respect to \(\filt_{\timestep-1}\), i.e. previsible.
Also recall that \([0\inter \timestep]\) is notation for a discrete ranges
\begin{equation}
    [n\inter m] := [n,m] \cap \integer, \qquad
    [n\inter m) := [n,m) \cap \integer,
    \qquad \text{etc.}
\end{equation}
We will now use this chain of vector spaces to define previsible coordinate
systems for every step \(\timestep\).

\begin{definition}[Previsible orthonormal coordinate systems]
    \label{def: previsible orthonormal coordinate system}
    We inductively define an \(\filt_{\timestep-1}\)-measurable orthonormal basis
    \(\rv_{[0:\dimV_\timestep)}\) of \(V_\timestep\) such that
    \(\rv_{[0:\dimV_k)}\) is a basis of the space \(V_k\).
    
    \paragraph*{Induction start} Assuming \(\param_0\neq 0\) and thus \(\dimV_0 =
    1\), we define
    \[
        \rv_0 :=\frac{\param_0}{\|\param_0\|}.
    \]
    In any case this results in a basis \(\rv_{[0:\dimV_0)}\) of \(V_0\)
    since \([0:\dimV_0)=\emptyset\) if \(\dimV_0 = 0\).

    \paragraph*{Induction step (Gram-Schmidt)}
    Assuming we have a basis \(\rv_{[0:\dimV_\timestep)}\) of \(V_\timestep\), let
    us construct a basis for \(V_{\timestep+1}\). For this we define the
    following Gram-Schmidt procedure candidate
    \begin{equation}
        \label{eq: definition of v candidate}
        \tilde{\rv}_{\timestep}
        := \nabla\rf_\dims(\Param_\timestep) - P_{V_\timestep}\nabla\rf_\dims(\Param_\timestep)
        = \nabla\rf_\dims(\Param_\timestep) - \sum_{i < \dimV_\timestep} \langle \nabla \rf_\dims(\Param_\timestep), \rv_i\rangle \rv_i,
    \end{equation}
    where \(P_{V_\timestep}\) is the projection to \(V_\timestep\).
    If \(\tilde{\rv}_{\timestep} = 0\), then \(\nabla\rf_\dims(\Param_\timestep) \in
    V_\timestep\) and we thus have \(V_{\timestep+1} = V_\timestep\) by definition
    \eqref{eq: vector space of evaluation points}. In that
    case we already have a basis for \(V_{\timestep+1}\).

    For any \(\timestep\) where the dimension increases such that
    \(\dimV_{\timestep+1} = \dimV_\timestep + 1\), we define
    \begin{equation}
        \label{eq: definition of v}
        \rv_{\dimV_{\timestep+1}-1}
        =\rv_{\dimV_\timestep}
        := \frac{\tilde{\rv}_{\timestep}}{\|\tilde{\rv}_{\timestep}\|}.
    \end{equation}
    We thus have obtained an orthonormal basis \(\rv_{[0:\dimV_{\timestep+1})}\)
    of \(V_{\timestep+1}\) which is \(\filt_\timestep\) measurable.
    
    \paragraph*{Basis extensions} With this construction of basis elements of
    \(V_\timestep\) done, we \(\filt_{\timestep-1}\)-measurably select an
    arbitrary orthonormal basis \(\rw^{(\timestep)}_{[\dimV_\timestep:\dims)}\)
    of \(V_\timestep^\perp\) to obtain an \(\filt_{\timestep-1}\)-measurable
    orthonormal basis
    \[
        B_\timestep := (\rv_{[0:\dimV_\timestep)}, \rw^{(\timestep)}_{[\dimV_\timestep:\dims)})
    \]
    for every \(\timestep\in\nat\). The coordinate systems \(B_\timestep\) are
    thus previsible.
\end{definition}

Using this specialized coordinate system we state an extension of
Theorem~\ref{thm: asymptotically deterministic behavior}. This extension looks less friendly but is more
natural to prove. It implies Theorem~\ref{thm: asymptotically deterministic
behavior}, proves three additional claims and relaxes the assumptions on
the prefactors of the gradient span algorithm. The additional claims cannot
be stated separately as they are all proved in one laborious induction. And to
prove the claim we are most interested in, i.e.\ \ref{ind: information convergence},
we also require the other claims in the induction step.
Finally, with the help of Proposition~\ref{prop: wlog deterministic starting
points}, we can (and will) assume deterministic starting points in
Theorem~\ref{thm: asymptotically deterministic behavior variant}
without loss of generality.

\begin{theorem}[Predictable progress {[Extension of Theorem~\ref{thm: asymptotically deterministic behavior}]}]
	\label{thm: asymptotically deterministic behavior variant}
Let \(\kernel\) be a kernel valid in all dimensions
	(Definition~\ref{def: valid in all dimensions}) and  \(\rf_\dims \sim \normal(\mu, \kernel)\) be a sequence of scaled (non-stationary) isotropic Gaussian random
	functions (Definition~\ref{def: isotropic gaussian random
	function}) in $\dims$. 
  Assume that \(\mu\) and \(\kernel\)  are
	sufficiently smooth (Assumption~\ref{assmpt: smoothness}) and assume the
	covariance of \((\rf_\dims,\nabla\rf_\dims)\) is strictly positive definite.
	Let \(\gsa\) be a general gradient span algorithm (Definition~\ref{def:
	general gsa}), which is \emph{asymptotically} continuous and uses
    the most recent gradient \emph{asymptotically} (cf.~Assumption~\ref{assmpt: generalized assumptions}).

	Let \(\gsa\) be applied to \(\rf_\dims\) with starting points
	\(\param_0\in \real^\dims\) such that \(\Param_\timestep = \gsa(\rf_\dims,
	\param_0, \timestep)\). We assume that the deterministic initialization point
	\(\param_0\in \real^\dims\) is of constant length \(\|\param_0\|=\radius\)
    over the dimension \(\dims\).
	Using \(\gradients_\timestep := \bigl(\nabla \rf_\dims(\Param_k): k \le
	\timestep\bigr)\), we further define the modified random information
	vector
    \[
		\VarInfo_\timestep
		:= \Bigl(\rf_\dims(\Param_k): k \le \timestep\Bigr)
		\cup \Bigl(
			\langle v, w\rangle :
			v \in \rv_{[0:\dimV_{\timestep+1})}, w \in \gradients_\timestep
		\Bigr).
    \]
    The following inductive claims hold for all \(\timestep\in \nat\), where
    \ref{ind: full dimension} implies that the vector \(\VarInfo_\timestep\) is
    almost surely of constant length.
    \begin{enumerate}[label=(Ind-\Roman*)]
        \item\label{ind: information convergence}
        \textbf{Information convergence:}
        There exists some deterministic limiting information vector
        \(\VarLiminfo_\timestep=\VarLiminfo_\timestep(\gsa, \mu, \kernel)\),
        such that
        \[
            \VarInfo_\timestep
            \underset{\dims\to\infty}{\overset{p}\to}
            \VarLiminfo_\timestep.
        \]
        The limiting information is split into \(\VarLiminfo_\timestep =
        (\limf_k, \gamma_k^{(i)})_{k,i}\),
        where the limiting elements \(\limf_k\) of \(\rf_\dims(\Param_k)\) were already
        defined in Theorem~\ref{thm: asymptotically deterministic behavior}.
        We further define the limiting inner products of gradients by
        \begin{equation}
            \label{eq: definition gamma}
            \gamma^{(i)}_k := \lim_{\dims\to\infty}
            \langle \nabla\rf_\dims(X_k), \rv_i\rangle.
        \end{equation}

        \item\label{ind: full dimension} \textbf{(Asymptotic) Full rank}:
        If the most recent gradient is always used (and not just asymptotically),
        then the previsible vector space of evaluation points \(V_{\timestep+1}\) has almost
        surely full rank (assuming \(\param_0\neq 0\)). Specifically, for all \(m\le \timestep+1\)
        we have almost surely
        \[
            \dimV_m = m + \ind_{\param_0 \neq 0}.
        \]
        This always holds in the limit (even when the most recent gradient is only
        used asymptotically), which means that for all \(k\le \timestep\) the
        Gram-Schmidt candidate \(\tilde{\rv}_{k}\) defined in \eqref{eq:
        definition of v} is not zero in the limit
        \begin{equation}
            \label{eq: limiting full rank}
            \gamma_k^{(\dimV_k)} = \lim_{\dims\to\infty}\langle \nabla\rf_\dims(\Param_k), \rv_{\dimV_k}\rangle = \lim_{\dims\to\infty}\|\tilde{\rv}_{k}\| \neq 0.
        \end{equation}

        \item\label{ind: representation}
        \textbf{Representation:} For all \(k\le m\le \timestep +1\) there exist
        limiting representation vectors
        \[
            y_k = y_k(\gsa, \mu, \kernel)
            \quad\text{with}\quad 
            y_k
            = (y_k^{(0)}, \dots y^{(\dimV_m-1)})\in \real^{\dimV_m}
        \]
        of \(\Param_k\), such that for all \(i < \dimV_m\) we have
        \begin{equation}
            \label{eq: <X, v> converges to gamma}
            \langle \Param_k, \rv_i\rangle
            \underset{\dims\to\infty}{\overset{p}\to} y_k^{(i)}
            \quad\text{and}\quad
            \|\Param_k\|^2
            = \sum_{i=0}^{\dimV_m-1} \langle \Param_k, \rv_i\rangle^2
            \underset{\dims\to\infty}{\overset{p}\to}
            \|y_k\|^2.
        \end{equation}
        The asymptotic distances are then given for all \(k,l\le m\) by
        \begin{equation}
            \label{eq: limiting distances}
            \|\Param_k - \Param_l\|^2
            = \sum_{i=0}^{\dimV_m-1} \langle \Param_k - \Param_l, \rv_i\rangle^2
            \underset{\dims\to\infty}{\overset{p}\to} 
            \; \rho_{kl}^2
            \;:= \|y_k - y_l\|^2.
        \end{equation}

        \item\label{ind: evaluation points asymptotically different}
        \textbf{The evaluation points are asymptotically different}, i.e.
        \[
            \rho_{kl}^2>0 \quad \text{for all}\quad k,l\leq \timestep+1 \quad \text{with}\quad k\neq l.
        \]
    \end{enumerate}
\end{theorem}

Before we prove that Theorem~\ref{thm: asymptotically deterministic behavior} follows from 
Theorem~\ref{thm: asymptotically deterministic behavior variant} let
us state the `asymptotic continuity' and `use of the most recent gradient' assumptions on the prefactors.

\begin{assumption}[Assumptions on prefactors]
    \label{assmpt: generalized assumptions}
    Recall, that we defined the prefactors \(\lr_{\timestep}\) of a GSA
    to be functions of the information \(\Info_{\timestep-1}\) (Definition~\ref{def: general gsa}).
    We assume that
    \begin{itemize}
		\item \(\lr_\timestep\) is continuous in the point \(\liminfo_{\timestep-1}\) for all \(\timestep\), and  
		\item the most recent gradients are used at least in the asymptotic
		limit, i.e.
		\[		
			\hat{\lr}^{(g)}_{\timestep,\timestep-1}
			:=\lr^{(g)}_{\timestep,\timestep-1}(\liminfo_{\timestep-1}) \Bigl(= \lim_{\dims\to\infty}\lr^{(g)}_{\timestep,\timestep-1}(\Info_{\timestep-1})\Bigr) \neq 0.
		\]
	\end{itemize}
    By Lemma~\ref{lem: info is continuous function} it is apparent, 
    that we can equivalently assume the prefactors \(\lr_\timestep\) to be functions in
    \(\VarInfo_{\timestep-1}\) continuous in \(\VarLiminfo_{\timestep-1}\),
    which use the most recent gradient in the asymptotic limit \(\hat{\lr}_{\timestep,\timestep-1} = \lr_{\timestep,\timestep-1}(\VarLiminfo_{\timestep-1})\).
\end{assumption}

At first it might seem circular to make use of limiting elements in an assumption
which is necessary to prove the limiting elements exist.
But a closer inspection reveals that \(\lr_\timestep\) is only used
for the convergence of \(\Info_{k}\) to \(\liminfo_{k}\) for times \(k\ge \timestep\).
We can therefore interleave this assumption on \(\lr_\timestep\) with the
inductive existence proof of \(\liminfo_{\timestep-1}\).

The proof of Theorem~\ref{thm: asymptotically deterministic behavior} now
follows readily from Theorem~\ref{thm: asymptotically deterministic behavior
variant} via a couple of preliminary results:

\begin{proposition}\label{prop: wlog deterministic starting points}
    We can assume without loss of generality that the independent initialization
    \(\Param_0\) is a deterministic \(\param_0\in \real^\dims\) of length
    \(\|\param_0\|=\radius\) over all dimensions \(\dims\).

    \textbf{Stationary case:} Assume the random functions \(\rf_\dims\) are
    furthermore stationary isotropic and the algorithm \(\param_0\)-agnostic
    (Definition~\ref{def: general gsa}). Then the limiting information is
    independent of \(\radius\) and the random
    initialization \(\Param_0\) does not need to satisfy
    \(\|\Param_0\|=\radius\) almost surely.
\end{proposition}

The proof of this Proposition is accomplished via two lemmas:

\begin{lemma}[The `information' is invariant to linear isometries]
    \label{lem: information invariant}
    Let \(f\) be some function and \((x_0, \dots, x_n)\) evaluation points.
    We further define a change in coordinates
    \[
        g(y) := f \circ \phi
        \quad \text{and}\quad
        y_k := \phi^{-1}(x_k) \quad\text{for}\quad k\in \{0,\dots, n\}
    \]
    via a linear isometry \(\phi(x) = Ux\) for some orthonormal matrix \(U\). Then the information
    \[
        \info_\timestep
        := \Bigl(f(\param_k): k \le \timestep\Bigr)
        \cup \Bigl(
            \langle v, w\rangle :
            v,w \in (\param_0) \cup \gradients_\timestep
        \Bigr)
        \quad\text{with}\quad
        \gradients_\timestep := \bigl(\nabla f(\param_k): k \le \timestep\bigr)
    \]
    is exactly equal to the information
    \[
        \varinfo_\timestep
        := \Bigl(g(y_k): k \le \timestep\Bigr)
        \cup \Bigl(
            \langle v, w\rangle :
            v,w \in (y_0) \cup \tilde{\gradients}_\timestep
        \Bigr)
        \quad\text{with}\quad
        \tilde{\gradients}_\timestep := \bigl(\nabla g(y_k): k \le \timestep\bigr).
    \]
    If \(\phi\) is a general isometry of the form \(\phi(x) = Ux + b\), then we
    still have equality for the the reduced information, i.e.
    \[
        \info^{\setminus x_0}_\timestep = \varinfo^{\setminus y_0}_\timestep.
    \]
\end{lemma}

\begin{proof}
    We have by definition
    \[
        f(\param_k)
        = f\circ \phi(\phi^{-1}(\param_k))
        = g(y_k).
    \]
    Let us therefore turn to the inner products. Since \(\phi(x) = Ux\) or
    \(\phi(x) = Ux + b\), we have for the gradient
    \begin{equation}
        \label{eq: basis change gradient}
        \nabla g(y) = U^\transpose \nabla f(\phi(y)).
    \end{equation}
     this implies all the inner products are equal 
    \begin{align*}
        \langle \nabla g(y_k), \nabla g(y_l)\rangle
        &= \langle U^\transpose\nabla f(x_k), U^\transpose \nabla f(x_l)\rangle
        &=& \langle \nabla f(x_k), \nabla f(x_l)\rangle.
        \\
    \intertext{
        In the linear isometry case, where \(\phi(x) = Ux\), we have \(y_n = U^\transpose x_n\) and thus
    }
        \langle y_0, \nabla g(y_l)\rangle
        &= \langle U^\transpose x_0, U^\transpose \nabla f(x_l)\rangle
        &=& \langle x_0, \nabla f(x_l)\rangle
        \\
        \langle y_0, y_0\rangle
        &= \langle U^\transpose x_0, U^\transpose x_0\rangle
        &=& \langle x_0, x_0\rangle.
    \end{align*}
    \vspace{-4ex}
\end{proof}

\begin{lemma}[Linear isometry invariance of general gradient span algorithms]
    \label{lem: gsa linear isometry invariant}
    Let \(\gsa\) be a general gradient span algorithm (Definition~\ref{def:
    general gsa}), let \(f\) be a function, \(x_0\) a starting point.
    We define a change of basis
    \[
        g(y) := f \circ \phi
        \quad \text{and}\quad
        y_0 := \phi^{-1}(x_0).
    \]
    via a linear isometry \(\phi(x) = Ux\) with orthonormal matrix \(U\).
    Then the optimization paths
    \[
        x_n := \gsa(f, \param_0, \timestep)
        \quad \text{and}\quad
        y_n := \gsa(g, y_0, \timestep)
    \]
    are invariant, i.e. the simple basis change
    \[
        y_n = \phi^{-1}(x_n)
    \]
    is retained for all \(n\in \nat\).
    The same holds true for all isometries \(\phi\), if the algorithm is
    \(x_0\)-agnostic.
\end{lemma}
\begin{proof}
    We proceed by induction over \(n\), where the induction start is obvious.

    For the induction step \((\timestep-1)\to \timestep\) we use the induction
    claim \(y_k = \phi^{-1}(x_k)\) for all \(k\le \timestep-1\) to obtain that
    the information \(\info_{\timestep-1}\) is invariant (by Lemma~\ref{lem:
    information invariant}). This implies by definition of the gradient
    span algorithm and \eqref{eq: basis change gradient}
    \[
        y_\timestep
		= \lr^{(\param)}_{\timestep} y_0
		+ \sum_{k=0}^{\timestep-1}\lr_{\timestep,k}^{(g)}\nabla g(y_k)
		= U^\transpose\Bigl(\lr^{(\param)}_{\timestep} x_0
            + \sum_{k=0}^{\timestep-1}\lr_{\timestep,k}^{(g)}\nabla f(x_k)
        \Bigr)
        = \phi^{-1}(x_\timestep),
    \]
    where the invariance of the information was used implicitly as the \(\lr_\timestep\)
    are functions of \(\info_{\timestep-1}\).

    In the \(\param_0\)-agnostic case, the induction step is almost the same
    except we have for isometries \(\phi(x) = Ux + b\) that \(y_k = \phi^{-1}(x_k) = U^T (x_k-b)\)
    and thus
    \[
        y_\timestep
		= y_0
		+ \sum_{k=0}^{\timestep-1}\lr_{\timestep,k}^{(g)}\nabla g(y_k)
		= U^\transpose\Bigl( x_0 - b
            + \sum_{k=0}^{\timestep-1}\lr_{\timestep,k}^{(g)}\nabla f(x_k)
        \Bigr)
        = \phi^{-1}(x_\timestep),
    \]
    using the fact that \(\lr^{(x)}_\timestep = 1\). We similarly use that the
    reduced information \(\info^{\setminus x_0}_{\timestep-1}\) is retained for
    the prefactors.
\end{proof}

We are now ready to prove that we can assume without loss of generality that
the initialization is deterministic.

\begin{proof}[Proof of Proposition~\ref{prop: wlog deterministic starting points}]
    Since we have
    \[
        \Pr\bigl(\VarInfo_\timestep \in A\bigr)
        = \E\Bigl[\Pr\bigl(\VarInfo_\timestep \in A \mid \Param_0\bigr)\Bigr],
    \]
    it is sufficient to show that \(\Pr(\VarInfo_\timestep \in A \mid
    \Param_0=\param_0)\) is only dependent on \(\|\param_0\|=\radius\),
    which is constant over the distribution of \(\Param_0\).

    For any \(\param_0\), \(y_0\) with \(\|\param_0\|=\|y_0\|=\radius\), there
    exists a linear isometry \(\phi\) such that \(\param_0 = \phi^{-1}(y_0)\).

    Let \(\Info_\timestep = \Info_\timestep(\rf_\dims, y_0)\) be
    the information vector generated from running the gradient
    span algorithm on \(\rf_\dims\) with starting point \(y_0\) for \(\timestep\)
    steps. Since \(\rf_\dims \overset{(d)}= \rf_\dims \circ \phi\) in
    distribution, due to (non-stationary) isotropy (cf.~Definition~\ref{def:
    isotropic gaussian random function}), we have
    \[
        \Info_\timestep(\rf_\dims, y_0)
        \overset{(d)}=
        \Info_\timestep(\rf_\dims\circ \phi, y_0)
        = \Info_\timestep(\rf_\dims, x_0)
    \]
    where we have used Lemma~\ref{lem: information invariant} and
    Lemma~\ref{lem: gsa linear isometry invariant} for the last equation.
    With the note that \(\VarInfo_\timestep\) is a deterministic map of
    \(\Info_\timestep\) as it only requires Gram-Schmidt orthogonalization as
    outlined in Definition~\ref{def: previsible orthonormal
    coordinate system} we can conclude this proof
    \begin{align*}
        \Pr\Bigl( \VarInfo_\timestep(\rf_\dims, \Param_0) \in A \mid \Param_0= y_0\Bigr)
        &= \Pr\Bigl( \VarInfo_\timestep(\rf_\dims, y_0) \in A\Bigr)
        \\
        &= \Pr\Bigl( \VarInfo_\timestep(\rf_\dims, x_0) \in A\Bigr)
        \\
        &= \Pr\Bigl( \VarInfo_\timestep(\rf_\dims, X_0) \in A \mid \Param_0= x_0\Bigr).
    \end{align*}

    In the stationary case with \(\param_0\)-agnostic algorithm, we can make use
    of arbitrary isometries \(\phi\). In particular we can chose \(\phi(x) = x-x_0\)
    that maps \(x_0\) to zero. With the same arguments as above (using the
    \(x_0\)-agnostic version of Lemma~\ref{lem: information invariant} and
    Lemma~\ref{lem: gsa linear isometry invariant}) we get
    \[
        \Pr\Bigl( \VarInfo_\timestep(\rf_\dims, \Param_0) \in A \mid \Param_0= x_0\Bigr)
        = \Pr\Bigl( \VarInfo_\timestep(\rf_\dims, \Param_0) \in A \mid \Param_0= 0\Bigr).
    \]
    The distribution is thus completely independent of \(\param_0\). In particular,
    this forces the limiting values to be independent of \(\param_0\) and therefore
    also independent of \(\lambda\).
\end{proof}

\begin{lemma}
    \label{lem: info is continuous function}
    For any fixed \(\param_0\),
    \(\Info_\timestep\) is a continuous function of \(\VarInfo_\timestep\) for all \(\timestep\in \nat\).
\end{lemma}
\begin{proof}
    Let us first recall the definition of \(\Info_\timestep\) and
    \(\VarInfo_\timestep\). With \(\gradients_\timestep := \bigl(\nabla \rf_\dims(\Param_k): k \le \timestep\bigr)\)
    we have in a direct comparison:
    \begin{align*}
		\Info_\timestep
		&:= \Bigl(\rf_\dims(\Param_k): k \le \timestep\Bigr)
		\cup \Bigl(
			\langle v, w\rangle :
			v,w \in (\param_0) \cup \gradients_\timestep
		\Bigr)
        \\
		\VarInfo_\timestep
		&:= \Bigl(\rf_\dims(\Param_k): k \le \timestep\Bigr)
		\cup \Bigl(
			\langle v, w\rangle :
			v \in \rv_{[0:\dimV_{\timestep+1})}, w \in \gradients_\timestep
		\Bigr).
    \end{align*}
    As the identity is a continuous map, we simply map the function values
    \(\rf_\dims(\Param_k)\) from \(\VarInfo_\timestep\) to itself in
    \(\Info_\timestep\). We therefore only need to find a way to continuously
    construct the inner products of \(\Info_\timestep\) from
    \(\VarInfo_\timestep\). Since \(\nabla\rf_\dims(\Param_k)\) is contained in
    \(V_{\timestep+1}\) by its definition \eqref{eq: vector
    space of evaluation points} and \(\rv_{[0:\dimV_{\timestep+1})}\) is a basis
    of \(V_{\timestep+1}\) by construction (Definition~\ref{def: previsible
    orthonormal coordinate system}), we have for all \(k,l\le \timestep\)
    \begin{align}
        \nonumber
        \langle \nabla \rf_\dims(\Param_k), \nabla\rf_\dims(\Param_l)\rangle
        &= \Bigl\langle
            \sum_{i=1}^{\dimV_{\timestep+1}-1}\langle \nabla \rf_\dims(\Param_k), \rv_i\rangle \rv_i,
            \sum_{j=1}^{\dimV_{\timestep+1}-1}\langle \nabla \rf_\dims(\Param_l), \rv_j\rangle\rv_j
        \Bigr\rangle
        \\
        \label{eq: representation of gradient inner prod}
        &= \sum_{i=0}^{\dimV_{\timestep+1}-1}
        \underbrace{\langle \nabla\rf_\dims(\Param_k), \rv_i\rangle}_{\in \VarInfo_\timestep}
        \underbrace{\langle\nabla\rf_\dims(\Param_l), \rv_i\rangle}_{\in \VarInfo_\timestep}
    \end{align}
    Since \(\dimV_{\timestep +1}\) is almost surely constant by \ref{ind: full
    dimension}\footnote{
        \ref{ind: full dimension} and \ref{ind: evaluation points asymptotically
        different} have to be sacrificed if one wanted to get rid of the strict
        positive definiteness assumption in future work. This argument then has
        to be replaced using an upper bound on the \(\dimV_{\timestep+1}\) and
        Lemma~\ref{lem: gamma is triangular}.
    }, this covers most of the inner products of \(\Info_\timestep\).
    What is left are the inner products using \(\param_0\). If \(\param_0 = 0\),
    then all those inner products are zero and the zero map does the job. 

    In the following we therefore assume \(\param_0 \neq 0\). Because \(\param_0\)
    is deterministic, we do not need to construct \(\|\param_0\|^2 = \radius^2\)
    from \(\VarInfo_\timestep\) as a constant map does the job. Since
    \(\param_0\neq 0\) implies \(\rv_0 = \frac{\param_0}{\|\param_0\|}\) by
    construction (Definition~\ref{def: previsible orthonormal coordinate
    system}), we have for all \(k\le \timestep\)
    \[
        \langle \param_0, \nabla\rf_\dims(\Param_k)\rangle
        = \|\param_0\| \underbrace{\langle \rv_0, \nabla\rf_\dims(\Param_k)\rangle}_{\in \VarInfo_\timestep}.
    \]
    We have thus continuously constructed all inner products in \(\Info_\timestep\)
    form \(\VarInfo_\timestep\).
\end{proof}

Here is how the main theorem of the paper follows from Theorem \ref{thm: asymptotically deterministic behavior variant}.
\begin{proof}[Proof of Theorem~\ref{thm: asymptotically deterministic behavior}]
    By Proposition~\ref{prop: wlog deterministic starting points}, we can assume
    without loss of generality that the initial point is deterministic.
    Since the other assumptions of Theorem~\ref{thm: asymptotically deterministic
    behavior variant} are the same (or even more general), we only need to prove that
    for every \(\timestep\in\nat\) there exists some \(\liminfo_\timestep\) such
    that
    \[
        \Info_\timestep\underset{\dims\to\infty}{\overset{p}\to} \liminfo_\timestep.
    \]
    As \(\Info_\timestep\) is a continuous function of \(\VarInfo_\timestep\)
    by Lemma~\ref{lem: info is continuous function} this follows immediately
    from continuous mapping by the inductive claim \ref{ind: information convergence}
    of Theorem~\ref{thm: asymptotically deterministic behavior variant}.
\end{proof}
The rest of the section is an inductive proof of Theorem~\ref{thm: asymptotically deterministic behavior variant}.

\subsection{Proof of Theorem~\ref{thm: asymptotically deterministic behavior variant}}

The heart of the proof will
be a lengthy induction. Before, we want to address the additional assumptions
of
\begin{enumerate}
    \item representable limit points \ref{ind: representation} and 
    \item the claim that these points are different \ref{ind: evaluation points asymptotically different},
\end{enumerate}
which we introduced in Theorem~\ref{thm: asymptotically deterministic behavior
variant} but did not use in the proof of Theorem~\ref{thm: asymptotically
deterministic behavior}. Together with the strictly positive definite covariance of
\((\rf_\dims,\nabla\rf_\dims)\) these assumptions will allow us to argue for converging entries of
covariance matrices and invertible limiting covariance matrices, which are used
in the conditional expectation and conditional variance.

\subsubsection{Complexity reduction}

Before we start the induction, we will perform some complexity reductions.

\begin{enumerate}
    \item We show that \ref{ind: representation} follows from \ref{ind:
    information convergence} in Lemma~\ref{lem: conv inf -> representation}.
    
    \item We show that \ref{ind: evaluation points asymptotically different} follows from
    \ref{ind: full dimension} and \ref{ind: information convergence} in
    Lemma~\ref{lem: conv info, full rank -> different eval pts}.
    
    \item We reduce the work necessary to prove \ref{ind: information convergence}.
\end{enumerate}

In the actual induction we will therefore be able to focus on the claims
\ref{ind: information convergence} and \ref{ind: full dimension}.

\begin{lemma}\label{lem: conv inf -> representation}
    For all fixed \(\timestep\in\nat\), the convergence of information \ref{ind: information convergence} implies the
    representation \ref{ind: representation}.
\end{lemma}

\begin{proof}
    Assuming \ref{ind: information convergence} we have that \(\VarInfo_\timestep \to
    \VarLiminfo_\timestep\). By Lemma~\ref{lem: info is continuous function} this
    also implies \(\Info_\timestep \to \liminfo_\timestep\). Since the prefactors
    \(\lr_m\) are functions of \(\Info_{m-1}\) continuous in \(\liminfo_{m-1}\),
    this implies for all \(m\le \timestep+1\) that the prefactors converge
    \[
        \lr_m = \lr_m(\Info_{m-1})
        \underset{\dims\to\infty}{\overset{p}\to} \lr_m(\liminfo_{m-1})
        =: \hat{\lr}_m.
    \]
    Recall that by \eqref{eq: definition gamma} of \ref{ind: information
    convergence} we have for all \(k\le \timestep\) and all \(i <
    \dimV_{\timestep+1}\) 
    \[
        \langle \nabla\rf_\dims(X_k), \rv_i\rangle
        \underset{\dims\to\infty}{\overset{p}\to}
        \gamma^{(i)}_k
    \]
    for some \(\gamma^{(i)}_k\in \real\). For all \(m\le \timestep+1\) we therefore
    have by definition of \(\Param_m\) (Definition~\ref{def: general gsa})
    \begin{align}
        \nonumber
        \langle  \Param_m, \rv_i\rangle
        &= \lr^{(\param)}_m\langle \param_0, \rv_i\rangle
        + \sum_{k=0}^{m-1} \lr^{(g)}_{m,k} \langle \nabla\rf_\dims(\Param_k), \rv_i\rangle
        \\
        \label{eq: inner product convergence}
        \overset{p}&{\underset{\dims\to\infty}\to}
        \; y^{(i)}_m \quad
        := \hat{\lr}^{(\param)}_m\|\param_0\|\delta_{0i}
        + \sum_{k=0}^{m-1} \hat{\lr}^{(g)}_{m,k} \gamma^{(i)}_k,
    \end{align}
    where \(\delta_{ij}\) denotes the Kronecker delta.
    Since \(\Param_k\) is contained in the vector space of evaluation points \(V_m\)
    for all \(k\le m\) by definition \eqref{eq: vector space of evaluation points},
    its norms converges
    \[
        \|\Param_k\|^2
        = \sum_{i=0}^{\dimV_m-1} \langle \Param_k, \rv_i\rangle^2 
        \overset{p}{\underset{\dims\to\infty}\to}
        \sum_{i=0}^{\dimV_m-1} (y_k^{(i)})^2 = \|y_k\|^2,
    \]
    and likewise their distances for all \(k,l\le m\)
    \[
        \|\Param_k - \Param_l\|^2
        = \sum_{i=0}^{\dimV_m-1} \langle \Param_k - \Param_l, \rv_i\rangle^2
        \underset{\dims\to\infty}{\overset{p}\to} 
        \rho_{kl}^2
        := \|y_k - y_l\|^2.
    \]
    This proves the limiting representation \ref{ind: representation}.
\end{proof}

In the following we will prove, assuming \ref{ind: information convergence}
and \ref{ind: full dimension}, that the limiting distances \(\rho_{kl}\) are
greater zero, i.e. \ref{ind: evaluation points asymptotically different}.
The main ingredients are \ref{ind: full dimension} and the assumed asymptotic
use of the last gradient.

\begin{lemma}\label{lem: conv info, full rank -> different eval pts}
    For all fixed \(\timestep\in\nat\), the convergence of information \ref{ind:
    information convergence} together with the asymptotic full rank \ref{ind:
    full dimension} imply asymptotically different evaluation points \ref{ind:
    evaluation points asymptotically different}.
\end{lemma}

\begin{proof}
    We will proceed by induction over \(m\), where we assume the claim to
    be shown for all \(k,l\le m\le
    \timestep+1\). The induction start \(m=0\) is trivial since a single point
    is always distinct. For the induction step we assume to have the statement
    for \(m\), that is
    \[
        \rho_{kl} > 0
        \quad\text{for all}\quad
        k, l\le m
        \quad\text{with}\quad
        k\neq l.
    \]
    To show the statement for \(m+1\le \timestep+1\), we only need to check the
    distances to the point \(y_{m+1}\) as we have the others by induction.

    To show that \(y_{m+1}\) is distinct from any \(y_k\) with \(k\le m\), we use the fact
    that the most recent gradient is used asymptotically by assumption of the
    Theorem~\ref{thm: asymptotically deterministic behavior variant}
    (cf.~Assumption~\ref{assmpt: generalized assumptions}), i.e.
    \begin{equation}
        \label{eq: using the most recent gradient}
        \hat{\lr}^{(g)}_{m+1, m} = \lr^{(g)}_{m+1, m}(\liminfo_m) \neq 0.
    \end{equation}
    The claim \ref{ind: full dimension} ensures that \(V_{m+1}\) has a larger
    dimension than \(V_m\), that is \(\dimV_{m+1}=\dimV_m+1\).  The last basis
    element of \(V_{m+1}\) is thus given by \(\rv_{\dimV_{m+1}-1}=\rv_{\dimV_m}\).
    By \eqref{eq: definition of v candidate} this element is produced from the last
    gradient \(\nabla\rf_\dims(\Param_m)\), which cannot be used for \(\Param_k\) with
    \(k\le m\) but must be used for \(\Param_{m+1}\) asymptotically by \eqref{eq: using
    the most recent gradient}. Therefore \(\Param_{m+1}\) asymptotically contains a
    component of \(\rv_{\dimV_m}\), that no other \(\Param_k\) has, which translates
    to the inequality of the asymptotic representations \(y_{m+1}\) and \(y_k\).

    Formally, since \(\Param_0,\dots\Param_m\) is contained in \(V_m\) spanned by
    \(\rv_{[0:\dimV_m)}\), we have for all \(k\le m\)
    \begin{equation}
        \label{eq: limiting components <= m}
        y_k^{(\dimV_m)}
        \overset{\eqref{eq: inner product convergence}}=
        \lim_{\dims\to\infty} \langle \Param_k, \rv_{\dimV_m}\rangle = 0.
    \end{equation}
    For the asymptotic representation of the last evaluation point \(\Param_{m+1}\)
    on the other hand, we have
    \begin{equation}
        \label{eq: limiting component m+1}
        y_{m+1}^{(\dimV_m)}
        \overset{\eqref{eq: inner product convergence}}= 
        \hat{\lr}^{(x)}_{m+1}\|\param_0\|\underbrace{\delta_{0 \dimV_m}}_{=0} + \sum_{k=0}^{m}\hat{\lr}^{(g)}_{m+1,k}\gamma_k^{(\dimV_m)}
        = \hat{\lr}^{(g)}_{m+1,m} \gamma_m^{(\dimV_m)}.
    \end{equation}
    The last equation is due to \eqref{eq: gamma later are zero}. This follows from
    the fact that for all \(k < m\) the gradient \(\nabla\rf_\dims(\Param_k)\) is
    contained in \(V_m\) spanned by \(\rv_{[0:\dimV_m)}\). By definition of
    \(\gamma_k^{(i)}\) \eqref{eq: definition gamma} we therefore have
    \begin{equation}
        \label{eq: gamma later are zero}
        \gamma_k^{(\dimV_m)}
        \overset{\eqref{eq: definition gamma}}= \lim_{\dims\to\infty} \langle \rf_\dims(\Param_k), \rv_{\dimV_m}\rangle
        = 0.
    \end{equation}
    Putting \eqref{eq: limiting components <= m} and \eqref{eq: limiting component m+1} together
    we have
    \begin{align*}
        \rho_{(m+1)k}^2
        &= \|y_{m+1} - y_k\|^2
        \ge (y_{m+1}^{(\dimV_m)} - y_k^{(\dimV_m)})^2
        = (\hat{\lr}^{(g)}_{m+1,m}\gamma_m^{(\dimV_m)})^2
        > 0,
    \end{align*}
    where the last inequality is due to the asymptotic use of the most recent
    gradient \eqref{eq: using the most recent gradient} and the limiting full rank
    claim \eqref{eq: limiting full rank} of \ref{ind: full dimension}.
\end{proof}

In \ref{ind: information convergence} we claim convergence of \(\VarInfo_\timestep\),
where \(\VarInfo_\timestep\) contains inner products of the form
\[
    \langle \rv_i, \nabla\rf_\dims(\Param_k)\rangle
\]
for \(k\le \timestep\) and \(i< \dimV_{\timestep+1}\). The following lemma
essentially implies that the restriction on \(i\) was unnecessary. So
when we increase \(\timestep-1\) to \(\timestep\) in the induction step, we do
not have to revisit the inner products of old gradients and can focus solely on
\(\nabla\rf_\dims(\Param_\timestep)\).

\begin{lemma}
    \label{lem: gamma is triangular}

    For all \(k\in \nat\) and \(i\ge \dimV_{k+1}\) we have
    \[
        \langle \nabla\rf_\dims(\Param_k), \rv_i\rangle \underset{\dims\to\infty}{\overset{p}\to} 0 = \gamma_k^{(i)}.
    \]
\end{lemma}
\begin{proof}
    For all \(i\ge \dimV_{k+1}\) the basis vector \(\rv_i\) is constructed to be orthogonal to
    \(\nabla\rf_\dims(\Param_k)\) contained in \(V_{k+1}\) spanned by
    \(\rv_{[0:\dimV_{k+1})}\). This implies
    \[
        \lim_{\dims\to\infty}\langle \nabla\rf_\dims(\Param_k), \rv_i\rangle = 0 = \gamma_k^{(i)}
    \]
    where the last equation is simply the definition of \(\gamma_k^{(i)}\) of \eqref{eq: definition gamma}.
\end{proof}

\begin{remark}[Gamma are triangular]\label{rem: gamma is triangular}
    Lemma~\ref{lem: gamma is triangular} can be visualized using a triangular
    matrix as follows. With the definition
    \[
        \real^n := \{ (\param_i)_{i\in [0:\infty)} : \param_i = 0 \quad \forall i\ge n\},
    \]
    we can view \(\real^m\) as a subspace of \(\real^n\) for \(m\le n\). The vector
    \[
        \gamma_k := (\gamma_k^{(i)})_{i\in [0:\dimV_{k+1})} \in \real^{\dimV_{k+1}},
    \]
    is then (by Lemma~\ref{lem: gamma is triangular}) also a member of
    \(\real^m\) for \(m\ge \dimV_{k+1}\). Concatenating the vectors
    \[
        \gamma_{[0:\timestep]}
        = \begin{pmatrix}
            \gamma_0^{(0)} & \dots & \gamma_\timestep^{(0)}\\
            \vdots & & \vdots\\
            \gamma_0^{(\dimV_{\timestep+1}-1)}
            & \dots
            & \gamma_\timestep^{(\dimV_{\timestep+1}-1)}
        \end{pmatrix}
        = \begin{pmatrix}
            \gamma_0^{(0)} & \dots & \gamma_{\timestep-1}^{(0)} & \gamma_\timestep^{(0)}\\
            \gamma_0^{(1)} & \dots & \gamma_{\timestep-1}^{(1)} & \gamma_\timestep^{(1)}
            \\
            0 & 
            \\
            \vdots  & \ddots & & \vdots
            \\
            0 & \dots & 0 & \gamma_\timestep^{(\dimV_{\timestep+1}-1)}
        \end{pmatrix}
    \]
    therefore results in a upper triangular matrix above an offset
    diagonal, since we always have \(\dimV_k \le k+1\) due to the definition of
    \(V_k\) \eqref{eq: vector space of evaluation points}.

    Note that, for visualization purposes, we assumed \(\dimV_{\timestep+1}
    =\dimV_{\timestep} +1\) in the second representation. If the dimension stays
    constant at some times \(k\), then the zeros encroach above this diagonal.
\end{remark}

\subsubsection{Induction start with \texorpdfstring{\(\timestep=0\)}{\timestep=0}}

We start the induction by proving the claim for \(\timestep=0\). We have
structured the induction start similar to the induction step. We therefore
suggest the reader to familiarize themselves with this strategy here, as it is
easier to get lost in the details of the induction step.

By Lemma~\ref{lem: conv inf -> representation} and Lemma~\ref{lem: conv
info, full rank -> different eval pts} we only need to prove \ref{ind:
information convergence} and \ref{ind: full dimension}. For \ref{ind:
information convergence} we need to prove that
\[
    \VarInfo_0
    = \bigl(\rf_\dims(\param_0)\bigr)
    \cup
    \bigl(
        \langle \nabla\rf_\dims(\param_0), v\rangle:
        v \in \rv_{[0:\dimV_1)}
    \bigr)
    = \bigl(
        \rf_\dims(\param_0),
        \langle \nabla\rf_\dims(\param_0), \rv_{[0:\dimV_1)}\rangle
    \bigr)
\]
converges to a limiting \(\VarLiminfo_0 = (\limf_0, \gamma_0^{(0)}, \dots, \gamma_0^{(\dimV_1-1)})\)
in probability. Note that \(\dimV_1 \le 2\) as the vector space of
evaluation points is defined in \eqref{eq: vector space of evaluation points} to be previsible,
that is
\[
    V_1 = \Span\{\param_0, \nabla\rf_\dims(\param_0)\}.
\]
So we have \(\dimV_1 -1\le 1\) depending on whether the starting point
\(\param_0\) is zero. The vector
\[
    \gamma_0 = (\gamma_0^{(i)})_{i< \dimV_1} = (\gamma_0^{(0)}, \dots, \gamma_0^{(\dimV_1-1)})
\]
might therefore collapse to a single entry \(\gamma_0^{(0)}\). The approach we
will now take mirrors the approach in the induction step. Beyond the pedagogical
benefit, this order is also quite natural for the induction start when using the
notation we introduced.
\begin{enumerate}[label=\emph{Step \arabic*}]
    \item\label{it: step 1 induction start}
    First we prove that
    \[
        \bigl(\rf_\dims(\param_0)\bigr) \cup
        \bigl( \langle \nabla\rf_\dims(\param_0), v\rangle : v\in \rv_{[0:\red{\dimV_0})}\bigr)
    \]
    converges. We highlight that the limit \red{\(\dimV_0\)} is purposefully different from the
    range of \(\VarInfo_0\).

    \item\label{it: step 2 induction start}
    We will then prove \ref{ind: full dimension}, which ensures the dimension
    actually increases
    \[
        \dimV_1 = \dimV_0 + 1.
    \]
    
    \item\label{it: step 3 induction start}
    Finally we prove that \(\langle \nabla\rf_\dims(\param_0), \rv_{\dimV_0}\rangle = \langle \nabla\rf_\dims(\param_0), \rv_{\dimV_1-1}\rangle\) converges. This
    adds the missing index from the first step.
\end{enumerate}
While we could prove \ref{ind: full dimension} before \ref{ind: information convergence} in
the induction start, we will require the results of \ref{it: step 1 induction start} for
\ref{ind: full dimension} in the induction step. The element \(\rv_{\dimV_0}\) is not
only different from \(\rv_{[0:\dimV_0)}\) in the sense that the dimension increase needs
to be shown, it is also the first truly random basis element here. In the induction
step the difference will be between the previsible basis elements and the last
non-previsible element, which needs to be treated differently.

\paragraph*{\ref{it: step 1 induction start}}
Since our random function is (non-stationary) isotropic with \(\rf_\dims\sim \normal(\mu, \kernel)\), we have
\[
    \rf_\dims(\param_0)
    \sim \normal\Bigl(
        \mu\bigl(\tfrac{\|\param_0\|^2}2\bigr),
        \frac1\dims \kernel\bigl(\tfrac{\|\param_0\|^2}2,\tfrac{\|\param_0\|^2}2, \|\param_0\|^2 \bigr)
    \Bigr).
\]
this immediately implies convergence of the first component
\[
    \rf_\dims(\param_0)
    \overset{p}{\underset{\dims\to\infty}\to}
    \mu\bigl(\tfrac{\|\param_0\|^2}2\bigr)
    = \mu\bigl(\tfrac{\radius^2}2\bigr) =: \limf_0.
\]
Let us now turn to the convergence of the inner products. We consider two
cases.

\textbf{Case (\(\param_0\neq 0\)):} In this case we have \(\rv_0 =
\frac{\param_0}{\|\param_0\|}\) by Definition~\ref{def: previsible orthonormal
coordinate system}. Therefore \(\rv_0\) is deterministic and by an application
of Lemma~\ref{lem: cov of derivatives, non-stationary isotropy} we have
\[
    D_{\rv_0}\rf_\dims(\param_0) \sim \normal\Bigl(
        \mu'\bigl(\tfrac{\|\param_0\|^2}2\bigr)\|\param_0\|,
        \frac1\dims\bigl[(\kernel_{12} + \kernel_{13} + \kernel_{32} + \kernel_{33})\|\param_0\|^2
        + \kernel_3\bigr]
    \Bigr),
\]
using the notation
\[
    \kernel
    := \kernel\bigl(\tfrac{\|\param_0\|^2}2, \tfrac{\|\param_0\|^2}2, \|\param_0\|^2\bigr)
    = \kernel\bigl(\tfrac{\radius^2}2, \tfrac{\radius^2}2, \radius^2\bigr)
\]
to omit inputs to the kernel \(\kernel\).
But this immediately implies
\[
    \langle \nabla\rf_\dims(\param_0), \rv_0\rangle 
    = D_{\rv_0}\rf_\dims(\param_0)
    \overset{p}{\underset{\dims\to\infty}\to}
    \mu'\bigl(\tfrac{\|\param_0\|^2}2\bigr)\|\param_0\|
    =: \gamma_0^{(0)}.
\]
As \(\dimV_0 = \dim(V_0) = \dim(\Span(\param_0)) = 1\), we have covered all elements
in \(\rv_{[0:\dimV_0)}\).

\textbf{Case (\(\param_0 = 0\)):} In this case, \(\dimV_0 = \dim(V_0) = 0\)
and \(\rv_{[0:\dimV_0)}\) is empty. We therefore do not have to do anything.

\paragraph*{\ref{it: step 2 induction start}}

To prove the dimension increases, we consider the Gram-Schmidt candidate
\begin{align*}
    \tilde{\rv}_0
    &:= \nabla\rf_\dims(\param_0) - P_{V_0}\rf_\dims(\param_0)
    = \nabla\rf_\dims(\param_0) - \sum_{i<\dimV_0} \langle\nabla\rf_\dims(\Param_0), \rv_i\rangle \rv_i
    \\
    &= \sum_{i= \dimV_0}^{\dims-1} \langle\nabla\rf_\dims(\Param_0), \rw^{(0)}_i\rangle \rw^{(0)}_i,
\end{align*}
where \(\rw^{(0)}_{[\dimV_0:\dims)}\) is the basis defined to be orthogonal to
\(V_0\) (Definition~\ref{def: previsible orthonormal coordinate system}). Since
\(V_0\) is deterministic, this orthogonal basis is also deterministic. In the
induction step, both will only be previsible. Note that in the case \(\param_0 = 0\),
\(\tilde{\rv}\) is simply the entire gradient by definition of \(\dimV_0\).

Since the \(\rw^{(0)}_i\) are deterministic, and orthogonal to \(\param_0\) in
either case, we can apply Lemma~\ref{lem: cov of derivatives, non-stationary isotropy} to obtain
\[
    D_{\rw^{(0)}_i}\rf_\dims(\param_0) \overset{\iid}\sim \normal\Bigl(0, \frac{\kernel_3}\dims\Bigr).
\]
This implies that there exist \(Y_i\overset{\iid}{\sim}\normal(0,1)\) such that
\[
    D_{\rw^{(0)}_i}\rf_\dims(\param_0) = \sqrt{\frac{\kernel_3}\dims} Y_i.
\]
But since \(\dimV_0 \le 1\) and in particular \(\dimV_0\) is finite, this implies
with the law of large numbers
\begin{equation}
    \label{eq: LLN induction start}
    \|\tilde{\rv}_0\|^2
    = \sum_{i=\dimV_0}^{\dims-1} \langle \nabla\rf_\dims(\param_0), w_i\rangle^2
    = \kernel_3 \cdot\frac1\dims \sum_{i=\dimV_0}^{\dims-1} Y_i^2
    \overset{p}{\underset{\dims\to\infty}\to} \kernel_3.
\end{equation}
Since we have \(\kernel_3>0\) by Lemma~\ref{lem: kappa_3 positive},
\(\|\tilde{\rv}_0\|^2\) is almost surely strictly greater than zero, which
implies \(\dim(V_1) > \dim(V_0)\) almost surely. Additionally,
the dimension also increases in the limit, i.e.
\(\lim_{\dims\to\infty}\|\tilde{\rv}_0\|=\kernel_3>0\).
We have therefore shown all of \ref{ind: full dimension}.

\paragraph*{\ref{it: step 3 induction start}}

We have by definition of \(\rv_{\dimV_0}\) in \eqref{eq: definition of v} and
the definition of \(\tilde{\rv}_0\)
\[
    \langle \nabla\rf_\dims(\param_0), \rv_{\dimV_0}\rangle 
    = \bigl\langle \nabla\rf_\dims(\param_0), \tfrac{\tilde{\rv}_0}{\|\tilde{\rv}_0\|}\bigr\rangle
    = \|\tilde{\rv}_{0}\|
    \overset{p}{\underset{\dims\to\infty}\to} \sqrt{\kernel_3}
    =: \gamma_0^{(\dimV_0)}.
\]
This finishes the last step and therefore the induction start.

\subsubsection{Induction step \texorpdfstring{\((\timestep-1)\to \timestep\)}{}}

We now get to the main body of the proof. Before we start, let us recapitulate
the lemmas we proved for complexity reduction. By Lemma~\ref{lem: conv inf ->
representation} and Lemma~\ref{lem: conv info, full rank -> different eval pts}
it is sufficient to prove the statements \ref{ind: information convergence} and
\ref{ind: full dimension}, as the others follow.

With \(\gradients_\timestep
:= \bigl(\nabla \rf_\dims(\Param_k): k \le \timestep\bigr)\) the modified information \(\VarInfo_\timestep\)
of the information convergence \ref{ind: information convergence} was given by
\[
    \VarInfo_\timestep
    := \Bigl(\rf_\dims(\Param_k): k \le \timestep\Bigr)
    \cup \Bigl(
        \langle v, w\rangle :
        v \in \rv_{[0:\dimV_{\timestep+1})}, w \in \gradients_\timestep
    \Bigr).
\]
By induction we already have \(\VarInfo_{\timestep-1}\to \VarLiminfo_{\timestep-1}\).
And due to our discussion of \(\langle \rv_i, \nabla\rf_\dims(\Param_k)\rangle\) for
\(i\ge \dimV_{\timestep+1} \ge \dimV_{k+1}\) in Lemma~\ref{lem: gamma is triangular}
we therefore only need to prove
\begin{equation}
    \label{eq: new column is convergence claim} 
    \begin{pmatrix}
        \rf_\dims(\Param_\timestep)
        \\
        \bigl\langle \rv_{[0:\dimV_{\timestep+1})}, \nabla\rf_\dims(\Param_\timestep)\bigr\rangle
    \end{pmatrix}
    = \begin{pmatrix}
        \rf_\dims(\Param_\timestep)
        \\
        D_{\rv_0}\rf_\dims(\Param_\timestep)
        \\
        \vdots
        \\
        D_{\rv_{\dimV_{\timestep+1}-1}}\rf_\dims(\Param_\timestep)
    \end{pmatrix}
    \underset{\dims\to\infty}{\overset{p}\to} \begin{pmatrix}
        \limf_\timestep
        \\
        \gamma_\timestep
    \end{pmatrix},
\end{equation}
where we used the definition \(\gamma_\timestep = (\gamma_\timestep^{(i)})_{i\in
[0:\dimV_{\timestep+1})}\) from Remark~\ref{rem: gamma is triangular}. In this
remark we have also explained how \(\gamma_k \in \real^{\dimV_{k+1}}\) can
be viewed as a member of \(\real^{\dimV_{\timestep+1}}\) and laid out the
concatenated vectors \(\gamma_{[0:\timestep]}\) in triangular matrix form.
In the following we will arrange the random information into a matching
matrix form by the definition of an auxiliary function \(Z\).

This function has multiple purposes: First, it turns the direction vectors of the
directional derivatives into inputs for an application of the previsible sampling
result \citep[Corollary 2.12]{benningMeasureTheoryConditionally2026}, which we explained in Idea 3 of the
proof sketch (Section~\ref{sec:sketch}). Second, it structures the information
vector \(\VarInfo_\timestep\) into a better readable matrix form for human
digestion. Third, it allows us to rearrange the elements of the column vectors
we concatenate into row vectors. This effectively rearranges the covariance
matrix of \(Z\) into a more sparse block form.

With some abuse of notation we define \(Z\) for a varying number
of direction vectors \(w_1,\dots,w_m\in \real^\dims\) and evaluation points
\(\param_1,\dots, \param_k\in\real^\dims\) for \(m\le \dims\) and \(k\le \timestep\)
\[
    Z(w_1,\dots, w_m; \param_0,\dots, \param_k) := \begin{pmatrix}
        \rf_\dims(\param_0) & \dots & \rf_\dims(\param_k)\\
        D_{w_1}\rf_\dims(\param_0) & \cdots & D_{w_1}\rf_\dims(\param_k)
        \\
        \vdots & & \vdots
        \\
        D_{w_m}\rf_\dims(\param_0) & \cdots & D_{w_m}\rf_\dims(\param_k)\\
    \end{pmatrix}.
\]
The number of inputs is therefore not variable and we separate their
type by a semicolon. Formally one would have to consider slices of \(Z\),
but since projections are measurable this is not an issue.
Note that we are not interested in this matrix as an
operator. The two dimensional layout is only used for better readability and we
will generally view it as a vector. For this purpose we treat the matrix as
`row-major', i.e. whenever we treat it like a vector, we concatenate the rows.
This grouping is purposefully different from the column layout of the \(\gamma_k\)
as it will enable a block matrix sparsity in the covariance matrices later on.

By induction we have information convergence \ref{ind: information convergence}
for \(\timestep-1\), which can be expressed using \(Z\) as
\begin{equation}
    \label{eq: main induction assumption}
    Z(\rv_{[0:\dimV_\timestep)}; \Param_{[0:\timestep)})
    = \begin{pmatrix}
        \rf_\dims(\Param_0) & \cdots & \rf_\dims(\Param_{\timestep-1}) 
        \\
        D_{\rv_0}\rf_\dims(\Param_0) & \cdots & D_{\rv_0}\rf_\dims(\Param_{\timestep-1})
        \\
        \vdots & & \vdots
        \\
        D_{\rv_{\dimV_\timestep-1}}\rf_\dims(\Param_0) & \cdots & D_{\rv_{\dimV_\timestep-1}}\rf_\dims(\Param_{\timestep-1})
    \end{pmatrix}
    \overset{p}{\underset{\dims\to\infty}\to}
    \begin{pmatrix}
        \limf_{[0:\timestep)}\\
        \gamma_{[0:\timestep)}
    \end{pmatrix}.
\end{equation}
Observe that the induction step simply adds the additional column given in
\eqref{eq: new column is convergence claim} to the right of the matrix,
where we use Lemma~\ref{lem: gamma is triangular} to argue that we can arbitrarily
extend the number of rows.

We will now follow the same strategy as in the induction start:

\begin{enumerate}[label=\emph{Step \arabic*}]
    \item\label{it: step 1 induction step}
    First, we prove the convergence of the `new column'
    \[
        Z(\rv_{[0:\red{\dimV_{\timestep}})}; \Param_\timestep)
        = \begin{pmatrix}
            \rf_\dims(\Param_\timestep)\\
            \langle \rv_{[0:\red{\dimV_{\timestep}})}, \nabla\rf_\dims(\Param_\timestep)\rangle
        \end{pmatrix}
        \underset{\dims\to\infty}{\overset{p}\to}
        \begin{pmatrix}
            \limf_\timestep\\
            \gamma_\timestep^{([0:\dimV_\timestep))}
        \end{pmatrix}
    \]
    \item\label{it: step 2 induction step}
    We prove \ref{ind: full dimension}, which ensures asymptotically
    \[
        \dimV_{\timestep+1} = \dimV_\timestep + 1.
    \]
    \item\label{it: step 3 induction step}
    We finally prove convergence of the `new corner element'
    \[
        Z(\rv_{\dimV_\timestep}; \Param_\timestep)
        = \langle \rv_{\dimV_\timestep}, \nabla\rf_\dims(\Param_\timestep)\rangle
        \underset{\dims\to\infty}{\overset{p}\to}
        \gamma_\timestep^{(\dimV_\timestep)}.
    \]
\end{enumerate}
The reason for this split are twofold. First, we have that
\(\rv_{[0:\dimV_\timestep)}\) is previsible, that is
\(\filt_{\timestep-1}\)-measurable, while \(\rv_\timestep\) is not.
\(\rv_\timestep\) must therefore be treated differently. Second,
we need to construct \(\rv_{\dimV_\timestep}\) from \(\tilde{\rv}_\timestep\),
which will naturally prove \ref{ind: full dimension} before we get
to the convergence of the inner product in \ref{it: step 3 induction step}.

This strategy can get a bit lost, as we will spend the majority of our time
with \ref{it: step 1 induction step} before wrapping up \ref{it: step 2
induction step} and \ref{it: step 3 induction step} fairly quickly.
An additional reason for this strategy to get lost is, that we will not
just consider
\((
    \rf_\dims(\param_\timestep),
    \langle \rv_{[0:\dimV_\timestep)}, \nabla \rf_\dims(\Param_\timestep)\rangle
)\), but instead consider the conditional distribution of
\((
    \rf_\dims(\param_\timestep),
    \langle \Basis_\timestep, \nabla \rf_\dims(\Param_\timestep)\rangle
)\)
for the entire previsible basis \(\Basis_\timestep = (\rv_{[0:\dimV_\timestep)},
\rw^{(\timestep)}_{[\dimV_\timestep:\dims)})\). Later, we will aggregate the
directional derivatives in the directions \(\rw^{\timestep}_i\) into
\(\tilde{\rv}_\timestep\), which means that we work towards \ref{it: step 1
induction step} and \ref{it: step 2 induction step} simultaneously.
The first objective will be, to apply the previsible sampling result
\citep[Corollary 2.12]{benningMeasureTheoryConditionally2026}
such that we can treat the previsible inputs as deterministic.

\paragraph*{Getting rid of the random input}

In the proof sketch we outlined how we needed to treat the random input with
care (cf.~\ref{sec:sketch},  Idea~3). In particular we need to consider the
(basis; point) pairs
\[
    (\Basis_0; \Param_0),\cdots, (\Basis_\timestep;\Param_\timestep)
    := \Bigl(\rv_{[0:\dimV_0)}, \rw_{[\dimV_0:\dims)}^{(0)}; \Param_0\Bigr),
    \cdots,
    \Bigl(\rv_{[0:\dimV_\timestep)}, \rw_{[\dimV_\timestep:\dims)}^{(\timestep)}; \Param_\timestep\Bigr).
\]
Recall that we defined the basis \(\Basis_k\) in Definition~\ref{def: previsible
orthonormal coordinate system} to be previsible just like \(\Param_k\), i.e.\
measurable with regard \(\filt_{k-1}\). Moreover, since all basis
changes are previsible and invertible it is straightforward
to check inductively that there exists a measurable bijective map between
different basis representations of the derivative information up to \(\timestep\)
\[
    \bigl(\rf_\dims(\Param_k),\nabla\rf_\dims(\Param_k)\bigr)_{k<\timestep}
    \leftrightsquigarrow \bigl(Z(\Basis_k; \Param_k)\bigr)_{k<\timestep}.
\]
This implies that their generated sigma algebras are the same
\[
    \filt_{\timestep-1}
    \overset{\text{def.}}=
    \sigma\Bigl(\bigl(\rf_\dims(\Param_k),\nabla\rf_\dims(\Param_k)\bigr), k\le\timestep-1\Bigr)
    = \sigma\bigl(Z(\Basis_k; \Param_k), k\le\timestep-1\bigr).
\]
By Corollary 2.12 from \citet{benningMeasureTheoryConditionally2026} we then have, for the
canonical conditional Gaussian distribution (Theorem \ref{thm: conditional
gaussian distribution}) and all bounded measurable \(h\),
\[
    \E\Bigl[
        h\bigl(Z(\Basis_\timestep;\Param_\timestep)\bigr) \mid \filt_{\timestep-1}
    \Bigr]
    = F(\Basis_{[0:\timestep]}; \Param_{[0:\timestep]}),
\]
for the function \(F\) defined using the canonical conditional Gaussian distribution
(Theorem \ref{thm: conditional gaussian distribution}) with deterministic evaluation points \(\param_k\)
and basis \(\basis_k\) as
\[
    F(\basis_{[0,\timestep]}; \param_{[0:\timestep]})
    := \E\Bigl[
        h\bigl(Z(\basis_\timestep;\param_\timestep)\bigr) \mid 
        Z(\basis_0;\param_0), \dots, Z(\basis_{\timestep-1};\param_{\timestep-1})
    \Bigr].
\]
The above is abuse of notation: We explicitly need the canonical conditional
Gaussian distribution for Corollary 2.12 from
\citet{benningMeasureTheoryConditionally2026} to be
applicable. This notation for the conditional expectation is however only
defined up to zero sets. As we have no interest in any other conditional distribution
than the canonical one, we hope that this notation is more helpful than confusing.

In other words, Corollary 2.12 from \citet{benningMeasureTheoryConditionally2026}
justifies treating the inputs \((\Basis_k; \Param_k)\)
as deterministic when calculating the conditional distribution. But keeping
track of \(\timestep+1\) different basis \(\Basis_k\) for every evaluation
point \(\Param_k\) is very inconvenient. So our next goal is to reduce the
number of basis to one.
For this we optimistically define for a single basis \(\basis\)
\[
    G(\basis; \param_{[0:\timestep]})
    := \E\Bigl[
        h\bigl(Z(\basis;\param_\timestep)\bigr) \mid 
        Z(\basis;\param_0), \dots, Z(\basis;\param_{\timestep-1})
    \Bigr].
\]
We are now going to observe that \(F\) is constant in all but the most recent
basis \(\basis_\timestep\), i.e.
\begin{equation}
    \label{eq: basis reduction}
    F(\basis_{[0:\timestep]}; \param_{[0:\timestep]})
    = G(\basis_\timestep; \param_{[0:\timestep]}).
\end{equation}
That is because the sigma algebras generated by different basis representations
are identical as they can be bijectively translated into each other, i.e.
for any \(\basis\)
\[
    \sigma(Z(\basis_0;\param_0), \dots, Z(\basis_{\timestep-1};\param_{\timestep-1}))
    = \sigma(Z(\basis;\param_0), \dots, Z(\basis;\param_{\timestep-1})).
\]
In particular this is true for \(\basis = \basis_\timestep\) and thus we have
\eqref{eq: basis reduction}, i.e.
\begin{align*}
    &\E\Bigl[
        h\bigl(Z(\basis_\timestep;\param_\timestep)\bigr) \mid 
        Z(\basis_0;\param_0), \dots, Z(\basis_{\timestep-1};\param_{\timestep-1})
    \Bigr]
    \\
    &= \E\Bigl[
        h\bigl(Z(\basis_\timestep;\param_\timestep)\bigr) \mid 
        Z(\basis_\timestep;\param_0), \dots, Z(\basis_\timestep;\param_{\timestep-1})
    \Bigr].
\end{align*}
Since we only need to consider the most recent basis from now on, we
drop the index and write \(\rw_k := \rw_k^{(\timestep)}\) and summarize our
result as
\begin{equation}
    \label{eq: inputs as deterministic}
    \E\Bigl[
        h\bigl(Z(\rv_{[0:\dimV_\timestep)}, \rw_{[\dimV_\timestep:\dims)}; \Param_\timestep)\bigr) \mid \filt_{\timestep-1}
    \Bigr]
    = G( \rv_{[0:\dimV_\timestep)}, \rw_{[\dimV_\timestep:\dims)}; \Param_0,\dots, \Param_\timestep).
\end{equation}
Recall that we use the semicolon to separate the basis elements from the evaluation points,
which should be treated as concatenated vectors respectively.

In essence, by definition of \(G\) we can now treat our evaluation points
\(\Param_{[0:\timestep]}\) and coordinate system \(\Basis_\timestep =
(\rv_{[0:\dimV_\timestep)}, \rw_{[\dimV_\timestep:\dims)})\) as deterministic when
calculating the conditional distribution.

\paragraph*{The conditional distribution is known in the Gaussian case}

Since \(Z\) is a Gaussian random function, we have for non-random input
\((\basis, \param)\) that its conditional distribution
is also Gaussian (cf.~Theorem~\ref{thm: conditional gaussian distribution}).
Equation~\eqref{eq: inputs as deterministic} translates this result to the
random input and we can therefore conclude that
\[
    Z(\Basis_\timestep; \Param_\timestep) \mid \filt_{\timestep-1}
    \sim \normal\Bigl(
        \E[Z(\Basis_\timestep; \Param_\timestep) \mid \filt_{\timestep-1}],
        \Cov[Z(\Basis_\timestep; \Param_\timestep) \mid \filt_{\timestep-1}]
    \Bigr).
\]
In particular there exist independent \(Y_1,\dots, Y_\dims \sim \normal(0,1)\)
independent of \(\filt_{\timestep-1}\) such that we have in distribution
\begin{equation}
    \label{eq: decomposition in distribution}
    Z(\Basis_\timestep; \Param_\timestep)
    = \E[Z(\Basis_\timestep; \Param_\timestep) \mid \filt_{\timestep-1}]
    + \sqrt{\Cov[Z(\Basis_\timestep; \Param_\timestep) \mid \filt_{\timestep-1}]} \begin{pmatrix}
        Y_0
        \\
        \vdots
        \\
        Y_{\dims}
    \end{pmatrix},
\end{equation}
where we denote the cholesky decomposition of a matrix \(A\) by the squareroot \(\sqrt{A}\).
\begin{remark}
    As we only wish to prove convergence in probability to deterministic numbers,
    this distributional equality will be enough for our purposes, but if the
    covariance was invertible one could get a almost sure  equality (cf.~Remark~\ref{rem:
    decomposition}). Later, we will also see that the covariance is at least
    asymptotically invertible due to strict positive definiteness of \(Z\)
    and asymptotically different evaluation points \ref{ind: evaluation points asymptotically different}.
\end{remark}

\paragraph*{Calculating the conditional expectation and covariance}

Since the first two moments can be calculated by treating the inputs as deterministic
(cf.~Equation~\eqref{eq: inputs as deterministic}), we can calculate the
conditional expectation and variance using the well known formulas for
Gaussian conditionals (cf.~Theorem~\ref{thm:
conditional gaussian distribution})
\begin{align}
    \label{eq: cond expect}
    \E[Z(\Basis_\timestep; \Param_\timestep) \mid \filt_{\timestep-1}]
    &= \rmean_{\timestep}^\dims + \blue{\rcov^\dims_{[0:\timestep),\timestep}}^T [\magenta{\rcov^\dims_{[0:\timestep)}}]^{-1} (Z(\Basis_\timestep; \Param_{[0:\timestep)})- \rmean_{[0:\timestep)}^\dims)
    \\
    \label{eq: cond variance}
    \Cov[Z(\Basis_\timestep; \Param_\timestep) \mid \filt_{\timestep-1}]
    &= \tfrac1\dims\Bigl[\green{\rcov^\dims_\timestep}
    - \blue{\rcov^\dims_{[0:\timestep),\timestep}}^T
    [\magenta{\rcov^\dims_{[0:\timestep)}}]^{-1}
    \blue{\rcov^\dims_{[0:\timestep),\timestep}}\Bigr],
\end{align}
where we define the mixed covariance by
\begin{align*}
    \tfrac1\dims\blue{\rcov^\dims_{[0:\timestep),\timestep}}
    &:= \C_Z(\Basis_\timestep; \Param_{[0:\timestep)}, \Param_\timestep)
    &
    \C_Z(\basis; \param_{[0:\timestep)}, \param_\timestep)
    &:= \Cov(Z(\basis; \param_{[0:\timestep)}), Z(\basis; \param_\timestep)),
\intertext{
    the autocovariance matrices by
}
    \tfrac1\dims\magenta{\rcov^\dims_{[0:\timestep)}}
    &:= \C_Z(\Basis_\timestep; \Param_{[0:\timestep)})
    &
    \C_Z(\basis; \param_{[0:\timestep)})
    &:= \Cov[Z(b; \param_{[0:\timestep)})]
    \\
    \tfrac1\dims\green{\rcov^\dims_\timestep}
    &:= \C_Z(\Basis_\timestep; \Param_\timestep)
    &
    \C_Z(\basis; \param_\timestep)
    &:=\Cov[Z(\basis; \param_\timestep)]
\intertext{
    and the expectations by
}
    \rmean_\timestep^\dims
    &:= \mu_Z(\Basis_\timestep; \Param_\timestep)
    &
    \mu_Z(\basis; \param_\timestep)
    &:= \E[Z(\basis; \param_\timestep)]
    \\
    \rmean_{[0:\timestep)}^\dims
    &:= \mu_Z(\Basis_\timestep; \Param_\timestep)
    &
    \mu_Z(\basis; \param_{[0:\timestep)})
    &:= \E[Z(\basis; \param_{[0:\timestep)})]
\end{align*}
The detour over the functions \(\C_Z\) and \(\mu_Z\) was necessary
to define the unconditional covariance matrices properly, because
inputs may only be treated as deterministic in conditional
distributions. So in general, since \(\Basis_\timestep\) and
\(\Param_\timestep\) are random variables,
we have for the unconditional covariance
\[
    \C_Z(\Basis_\timestep; \Param_\timestep) \neq \Cov[Z(\Basis_\timestep; \Param_\timestep)].
\]
To ensure the inputs are treated as deterministic as Equation~\eqref{eq: inputs
as deterministic} demands, we therefore had to make sure the expectation
was already applied before we plugged in our random input.

Note that we have moved the dimensional scaling \(\frac1\dims\) of the covariances
(cf.~Lemma~\ref{lem: cov of derivatives, non-stationary isotropy}) outside of our definition of
\(\magenta{\rcov^\dims_{[0:\timestep)}}\), \(\blue{\rcov^\dims_{[0:\timestep),\timestep}}\)
and \(\green{\rcov^\dims_\timestep}\), as we are now going to prove their entries, and
the entries of \(\rmean_{[0:\timestep)}^\dims\) and \(\rmean_\timestep^\dims\),
converge. This will eventually allow us to prove that the conditional expectation and
covariance will converge, which leads to the stochastic convergence of \(Z\) we
want to obtain for \ref{ind: information convergence}. Moving the dimensional scaling out also made the
dimensional scaling of the conditional covariance in \eqref{eq: cond variance}
much more visible.

\paragraph*{Covariance matrix entries converge}

Recall that \(Z\) is made up of evaluations of \(\rf_\dims\) and
\(D_v\rf_\dims\) for directions \(v\). Thus the entries of its covariance matrices are
given by Lemma~\ref{lem: cov of derivatives, non-stationary isotropy}, which we restate here for
your convenience.

\covOfDerivatives*

Recall that we have \eqref{eq: <X, v> converges to gamma} of \ref{ind: representation}
for \(\timestep-1\) by induction assumption, i.e. we have for all \(k\le \timestep\)
and all \(i < \dimV_\timestep\)
\[
    \langle \Param_k, \rv_i\rangle
    \underset{\dims\to\infty}{\overset{p}\to} y_k^{(i)}
    \quad\text{and}\quad
    \|\Param_k\|^2
    = \sum_{i=0}^{\dimV_\timestep-1} \langle \Param_k, \rv_i\rangle^2
    \underset{\dims\to\infty}{\overset{p}\to}
    \|y_k\|^2.
\]
Since the \(\Param_k\) for \(k\le \timestep\) are contained in \(V_\timestep\)
orthogonal to \(\rw_{[\dimV_\timestep: \dims)}\) we also have for all \(i\ge
\dimV_\timestep\)
\begin{equation}
    \label{eq: param and w is orthogonal}
    \langle \Param_k, \rw_i\rangle = 0.
\end{equation}
Put together, we have that all the inner products \(\langle \Param_k, v\rangle\)
for \(k\le \timestep\) and \(v\in \Basis_\timestep\) converge.

The entries of 
\(\magenta{\rcov^\dims_{[0:\timestep)}}\),
\(\blue{\rcov^\dims_{[0:\timestep),\timestep}}\),
\(\green{\rcov^\dims_\timestep}\), \(\rmean_{[0:\timestep)}^\dims\) and
\(\rmean_\timestep^\dims\) calculated with
Lemma~\ref{lem: cov of derivatives, non-stationary isotropy} therefore all converge in probability
by the continuous mapping theorem, since \(\kernel\) and \(\mu\) are
sufficiently smooth by Assumption~\ref{assmpt: smoothness} used in
Theorem~\ref{thm: asymptotically deterministic behavior variant}.
Observe that it was very important to remove the dimensional scaling \(\frac1\dims\)
of \eqref{eq: cov df, f} and \eqref{eq: cov df, df}
from the covariance matrices \(\magenta{\rcov^\dims_{[0:\timestep)}}\),
\(\blue{\rcov^\dims_{[0:\timestep),\timestep}}\) and
\(\green{\rcov^\dims_\timestep}\), as their entries would otherwise all
converge to zero. As we want to invert \(\magenta{\rcov^\dims_{[0:\timestep)}}\)
this would have been very inconvenient.

We further note that the sizes of these covariance matrices change with the dimension,
because the number of directional derivatives increases with \(\dims\) which
increases the size of \(Z(\rv_{[0:\dimV_\timestep)}, \rw_{[\dimV_\timestep:\dims)}; \Param_{[0:\timestep)})\)
and therefore the size of its covariance matrix. The convergence of their
entries is therefore not yet sufficient for a limiting object to be well defined.

\paragraph*{Splitting the increasing matrices into block matrices of constant size}

This is the heart of the proof, which relies heavily on our custom
coordinate system \(\Basis_\timestep=(\rv_{[0:\dimV_\timestep)}, \rw_{[\dimV_\timestep:\dims)})\). To understand this,
let us focus on the covariance of derivatives given in \eqref{eq: cov df, df} of
Lemma~\ref{lem: cov of derivatives, non-stationary isotropy}, i.e.
\begin{equation}
    \tag{\ref{eq: cov df, df}}
    \Cov(D_v \rf_\dims(x), D_w\rf_\dims(y))   
    = \frac1\dims\Bigl[\underbrace{\begin{aligned}[t]
        &\kernel_{12} \langle x, v\rangle\langle y, w\rangle + \kernel_{13} \langle x, v\rangle\langle x, w\rangle
        \\
        & + \kernel_{32} \langle y, v\rangle\langle y, w\rangle + \kernel_{33}\langle y, v\rangle\langle y, w\rangle
        \\
    \end{aligned}}_{\text{(I)}}
    + \underbrace{\kernel_3 \langle v, w\rangle}_{\text{(II)}}\Bigr].
\end{equation}
Notice that for the \(\rw_i\) the part (I) is always zero by \eqref{eq: param
and w is orthogonal}. This is why we defined \(\rw_{[\dimV_\timestep:\timestep)}\)
to be orthogonal to \(V_\timestep\) in Definition~\ref{def: previsible
orthonormal coordinate system}. Since we also defined our basis to be an
orthonormal basis, (II) is only non-zero when covariances of the directional
derivatives in the same direction are taken.

As we treat our \(Z\) matrix \eqref{eq: main induction assumption} as row-major,
the directional derivatives are grouped by direction \(\rw_i\) in \(Z\), which therefore
results in the following block matrix structure.
\begin{align}
    \nonumber
    \tfrac1\dims\magenta{\rcov_{[0:\timestep)}^{\dims}}
    &=\C_Z(\rv_{[0:\dimV_\timestep)}, \rw_{[\dimV_\timestep:\dims)}; \Param_{[0:\timestep)})
    \\
    \nonumber
    &= \begin{pmatrix}
        \C_Z(\rv_{[0:\dimV_\timestep)}; \Param_{[0:\timestep)})
        \\
        & \C_{D_{\rw_{\dimV_\timestep}}\rf_\dims}(\Param_{[0:\timestep)})
        \\
        & & \ddots 
        \\
        & & & \C_{D_{\rw_{\dims-1}}\rf_\dims}(\Param_{[0:\timestep)})
    \end{pmatrix}
    \\
    &=: \frac1\dims\left(\begin{array}{c c c c}
        \cline{1-1}
        \multicolumn{1}{|c|}{}
        \\
        \multicolumn{1}{|c|}{\quad\magenta{\rcov^{\black{v},\dims}_{[0:\timestep)}}\quad} 
        \\
        \multicolumn{1}{|c|}{}
        \\
        \cline{1-2}
        &\multicolumn{1}{|c|}{\magenta{\rcov^{\black{w},\dims}_{[0:\timestep)}}}
        \\
        \cline{2-2}
        & & \ddots
        \\
        \cline{4-4}
        & & &\multicolumn{1}{|c|}{\magenta{\rcov^{\black{w},\dims}_{[0:\timestep)}}}
        \\
        \cline{4-4}
    \end{array}\right),
    \label{eq: block matrix autocovariance}
\end{align}
For \(\magenta{\rcov^{\black{w},\dims}_{[0:\timestep)}}\) to be well defined, we
require the blocks \(\C_{D_{\rw_{i}}\rf_\dims}(\Param_{[0:\timestep)})\) to not depend
on \(i\). But since (I) is zero and \(\langle \rw_i, \rw_i\rangle =1\) we have
by \eqref{eq: cov df, df} of Lemma~\ref{lem: cov of derivatives, non-stationary isotropy}
\begin{align*}
    &\C_{D_{\rw_i}\rf_\dims}(\Param_{[0:\timestep)})
    \\
    &= \frac1\dims \begin{pmatrix}
        \kernel_3\bigl(\tfrac{\|\Param_0\|^2}2, \tfrac{\|\Param_0\|^2}2, \langle \Param_0, \Param_0\rangle\bigr)
        &\cdots
        & \kernel_3\bigl(\tfrac{\|\Param_0\|^2}2, \tfrac{\|\Param_{\timestep-1}\|^2}2, \langle \Param_0, \Param_{\timestep-1}\rangle\bigr)
        \\
        \vdots & & \vdots
        \\
        \kernel_3\bigl(\tfrac{\|\Param_{\timestep-1}\|^2}2, \tfrac{\|\Param_0\|^2}2, \langle \Param_{\timestep-1}, \Param_0\rangle\bigr)
        &\cdots
        & \kernel_3\bigl(\tfrac{\|\Param_{\timestep-1}\|^2}2, \tfrac{\|\Param_{\timestep-1}\|^2}2, \langle \Param_{\timestep-1}, \Param_{\timestep-1}\rangle\bigr)
    \end{pmatrix}
    \\
    &=: \tfrac1\dims\magenta{\rcov^{\black{w},\dims}_{[0:\timestep)}}.
\end{align*}
In particular there is no dependence on \(\rw_i\) so
\(\magenta{\rcov^{\black{w},\dims}_{[0:\timestep)}}\) is well defined. Since
these block matrices are of constant size, they converge if
all their finitely many entries converge. But we already argued that the entries
converge (in the previous paragraph) and we therefore have\footnote{
    The increasing number of identical
    \(\magenta{\rcov^{\black{w},\dims}_{[0:\timestep)}}\) will drive the law of large numbers of
    \[
        \|P_{V_\timestep^\perp}\nabla\rf_\dims(\param_0)\|^2 = \sum_{i=\dimV_{\timestep}}^{\dims-1} (D_{\rw_i}\rf_\dims(\param_0))^2
    \]
    similar to the first step (cf.~Equation~\eqref{eq: LLN induction start}). At the moment they are still matrices but this will change when
    combined with the mixed covariances \(\blue{\rcov^{\black{w}, \dims}_{[0:\timestep),\timestep}}\)
    finally resulting in \eqref{eq: limiting residual variance}.
}
\[
    \magenta{\rcov^{\black{v},\dims}_{[0:\timestep)}}
    \underset{\dims\to\infty}{\overset{p}\to}
    \magenta{\Sigma^{\black{v},\infty}_{[0:\timestep)}}
    \in \real^{\timestep (\dimV_\timestep+1) \times \timestep (\dimV_\timestep+1)}
    \quad \text{and} \quad
    \magenta{\rcov^{\black{w},\dims}_{[0:\timestep)}}
    \underset{\dims\to\infty}{\overset{p}\to} \magenta{\Sigma^{\black{w},\infty}_{[0:\timestep)}}
    \in \real^{\timestep \times \timestep}.
\]
For the mixed covariance we similarly have
\begin{align}
    \nonumber
    \tfrac1\dims\blue{\rcov^\dims_{[0:\timestep),\timestep}}
    &=\C_Z(\rv_{[0:\dimV_\timestep)},\rw_{[\dimV_\timestep:\dims)};\Param_{[0:\timestep)}, \Param_\timestep)
    \\
    \nonumber
    &= \begin{pmatrix}
        \C_Z(\rv_{[0:\timestep)};\Param_{[0:\timestep)},\Param_\timestep)
        \\
        & \C_{D_{w_{\dimV_\timestep}}\rf_\dims}(\Param_{[0:\timestep)}, \Param_\timestep)
        \\
        & & \ddots
        \\
        & & & \C_{D_{w_{\dims-1}}\rf_\dims}(\Param_{[0:\timestep)},\Param_\timestep)
    \end{pmatrix}
    \\
    &=: \frac1\dims\left(\begin{array}{c c c c}
        \cline{1-1}
        \multicolumn{1}{|c|}{}
        \\
        \multicolumn{1}{|c|}{}
        \\
        \multicolumn{1}{|c|}{\;\blue{\rcov^{\black{v},\dims}_{[0:\timestep),\timestep}}\;} 
        \\
        \multicolumn{1}{|c|}{}
        \\
        \multicolumn{1}{|c|}{}
        \\
        \cline{1-2}
        &\multicolumn{1}{|c|}{}
        \\
        &\multicolumn{1}{|c|}{\blue{\rcov^{\black{w},\dims}_{[0:\timestep),\timestep}}}
        \\
        &\multicolumn{1}{|c|}{}
        \\
        \cline{2-2}
        & & \ddots
        \\
        \cline{4-4}
        & & &\multicolumn{1}{|c|}{}
        \\
        & & &\multicolumn{1}{|c|}{\blue{\rcov^{\black{w},\dims}_{[0:\timestep),\timestep}}}
        \\
        & & &\multicolumn{1}{|c|}{}
        \\
        \cline{4-4}
    \end{array}\right),
    \label{eq: block matrix mixed cov}
\end{align}
with a similar argument why \(\blue{\rcov^{\black{w},\dims}_{[0:\timestep),\timestep}}\) is well defined as for
\(\magenta{\rcov^{\black{w},\dims}_{[0:\timestep)}}\).
And again by the discussion of the previous paragraph establishing convergence
of the entries, we have that these block matrices of constant size converge
\[
    \blue{\rcov^{\black{v},\dims}_{[0:\timestep),\timestep}}
    \underset{\dims\to\infty}{\overset{p}\to}
    \blue{\Sigma^{\black{v},\infty}_{[0:\timestep),\timestep}}
    \in \real^{\timestep(\dimV_\timestep+1) \times (\dimV_\timestep+1)}
    \quad\text{and}\quad
    \blue{\rcov^{\black{w},\dims}_{[0:\timestep),\timestep}}
    \underset{\dims\to\infty}{\overset{p}\to}
    \blue{\Sigma^{\black{w},\infty}_{[0:\timestep),\timestep}}
    \in \real^{\timestep \times 1}.
\]
We can also split up the autocovariance \(\green{\rcov^{\dims}_\timestep}\) in a similar fashion with
\[
    \green{\rcov^{\black{v},\dims}_\timestep}
    \underset{\dims\to\infty}{\overset{p}\to}
    \green{\Sigma^{\black{v},\infty}_\timestep}
    \in \real^{(\dimV_\timestep +1) \times (\dimV_\timestep+1)}
    \quad\text{and}\quad
    \green{\rcov^{\black{w},\dims}_\timestep}
    \underset{\dims\to\infty}{\overset{p}\to}
    \green{\Sigma^{\black{w},\infty}_\timestep}
    \in \real^{1 \times 1}.
\]
Finally, the expectation functions can also be split into a block containing
the directional derivatives \(\rv_{[0:\dimV_\timestep)}\) spanning \(V_\timestep\)
and the expectations of directional derivatives in the \(\rw_i\) directions.
More specifically we have
\begin{equation}
    \label{eq: expectation v} 
    \mu_\timestep^{v,\dims} = \mu_Z(\rv_{[0:\dimV_\timestep)}; \Param_\timestep)
    \underset{\dims\to\infty}{\overset{p}\to}
    \mu_\timestep^{v, \infty}\in\real^{\dimV_\timestep +1}
    \quad\text{and}\quad
    \mu_{[0:\timestep)}^{v,\dims}
    \underset{\dims\to\infty}{\overset{p}\to}
    \mu_{[0:\timestep)}^{v, \infty}\in\real^{\timestep(\dimV_\timestep +1)},
\end{equation}
which converge since the inner products and norms used in \eqref{eq: expect df}
converge.
Since the directions \(\rw_i\) are selected orthogonal to all evaluation
points \(\Param_\timestep\in V_\timestep\), the inner product in the expectation
of the directional derivative \eqref{eq: expect df} is always zero
and we therefore have
\begin{equation}
    \label{eq: expectation w}
    \mu_\timestep^{w,\dims} = \mu_Z(\rw_{[\dimV_\timestep:\dims)}; \Param_\timestep)
    = 0 \in\real
    \quad\text{and}\quad
    \mu_{[0:\timestep)}^{w,\dims}
    = 0 \in\real^{\timestep}.
\end{equation}

\paragraph*{Convergence of the conditional expectation}

Let us take a step back and review what we have done and want to do. We have
argued that \(Z(\Basis_\timestep; \Param_{[0:\timestep)})\) is conditionally Gaussian and that it
is therefore enough to understand the conditional expectation and covariance
(cf.~\eqref{eq: decomposition in distribution}). We have then applied a
well known result about the conditional distribution of Gaussian random
variables to obtain explicit formulas for the conditional expectation \eqref{eq: cond expect}
and covariance \eqref{eq: cond variance} made up of unconditional covariance
matrices. We proved that their entries converged and split these covariance
matrices into block form, such that these blocks of constant size converge in
probability. What is left to do, is to put these results together to prove
convergence results about \(Z(\Basis_\timestep; \Param_{[0:\timestep)})\). We start with the
conditional expectation and then get to \(Z(\Basis_\timestep; \Param_{[0:\timestep)})\) itself
by a consideration of the conditional covariance.

So let us take a look at the conditional expectation.
Since the \(\rw_i\) are by definition orthonormal to the previsible running span
of evaluation points \(V_\timestep\) \eqref{eq: vector space of evaluation
points}, we have \(D_{\rw_k}\rf_\dims(\Param_{[0:\timestep)})=0\). By applying the
block matrix structure \eqref{eq: block matrix autocovariance}, \eqref{eq: block
matrix mixed cov} to the the representation of the conditional expectation
\eqref{eq: cond expect} we therefore get using \eqref{eq: expectation w}
\begin{equation}
    \label{eq: conditional expectation in direction w}
    \E[D_{\rw_i}\rf_\dims(X_\timestep)\mid \filt_{\timestep-1}]
    = \mu_\timestep^{w,\dims}
    + \blue{\rcov^{\black{w},\dims}_{[0:\timestep),\timestep}}^\transpose
    [\magenta{\rcov^{\black{w},\dims}_{[0:\timestep)}}]^{-1}
    (D_{\rw_k}\rf_\dims(\Param_{[0:\timestep)}) - \mu_{[0:\timestep)}^{w,\dims})
    = 0.
\end{equation}
The only interesting part of the vector \(\E[Z(\rv_{[0:\dimV_\timestep)},
\rw_{[\dimV_\timestep:\dims)}; \Param_\timestep) \mid \filt_{\timestep-1}]\) are therefore the first
\(\dimV_\timestep\) entries \(\E[Z(\rv_{[\dimV_\timestep:\timestep)}; \Param_\timestep) \mid
\filt_{\timestep-1}]\). For this we use the formula for the conditional expectation
\eqref{eq: cond expect} and our block matrix decompositions \eqref{eq: block
matrix autocovariance}, \eqref{eq: block matrix mixed cov} again to obtain 
\begin{equation}
    \label{eq: relevant part of cond expec}
    \E[Z(\rv_{[0:\dimV_\timestep)}; \Param_\timestep) \mid \filt_{\timestep-1}]
    = \mu_\timestep^{(v,\dims)}
    +\blue{\rcov^{\black{v},\dims}_{[0:\timestep),\timestep}}^\transpose
    [\magenta{\rcov^{\black{v},\dims}_{[0:\timestep)}}]^{-1}
    (Z(\rv_{[0:\dimV_\timestep)}; \Param_{[0:\timestep)})-\mu_{[0:\timestep)}^{(v,\dims)}).
\end{equation}
Since \(Z(\rv_{[0:\dimV_\timestep)}; \Param_{[0:\timestep)})\) converges in probability by the
induction assumptions \ref{ind: information convergence}  summarized in
\eqref{eq: main induction assumption}, and since we have spent the previous
paragraphs proving that the covariance matrices and expectations converge, it
almost seems like we are done. But while the conditional expectation \eqref{eq:
cond expect} can handle non-invertible matrices with a generalized inverse
(Theorem~\ref{thm: conditional gaussian distribution}), an application of
continuous mapping requires continuity and the matrix inverse is only continuous
at invertible matrices.

\begin{lemma}
    \label{lem: strict positive cov matrix}
    \(\magenta{\Sigma^{\black{v},\infty}_{[0:\timestep)}}\) is strictly positive definite and
    therefore invertible.
\end{lemma}
\begin{proof}
    The essential ingredients are that \((\rf_\dims, \nabla\rf_\dims)\) is strictly positive
    definite by assumption of Theorem~\ref{thm: asymptotically deterministic
    behavior variant} and that the evaluation points are asymptotically
    different \(\rho_{ij} >0\) by
    \ref{ind: evaluation points asymptotically different}. The technical
    complications are that
    \(\magenta{\Sigma^{\black{v},\infty}_{[0:\timestep)}}\) is only defined as
    the limit of covariance matrices (which are themselves not necessarily
    strictly positive definite themselves) and that this limit involves changing
    the domain of \((\rf_\dims, \nabla\rf_\dims)\) as we increase its dimension.

    Let us first address the problem of the changing domain.
    Since we are only interested in the block matrix of the derivatives
    in the directions contained in \(V_\timestep\) and the evaluation points
    \(\Param_{[0:\timestep)}\) are also contained in \(V_\timestep\) we can map
    them via an isometry to \(\real^{\dimV_\timestep}\) which retains all distances and inner
    products and therefore retains the covariance matrices
    \(\magenta{\rcov^{\black{v},\dims}_{[0:\timestep)}}\) (cf.\ Lemma~\ref{lem: cov of derivatives, non-stationary isotropy}).
    Note that the dimensional scaling \(\frac1\dims\) is not an issue here, as
    we have already removed the scaling from
    \(\magenta{\rcov^{\black{v},\dims}_{[0:\timestep)}}\).
    Therefore \(\magenta{\rcov^{\black{v},\dims}_{[0:\timestep)}}\) can be viewed as a sequence
    of covariance matrices of \((\rf_\dims,\nabla\rf_\dims)\) with \(\rf_\dims:\real^{\dimV_\timestep}\to \real\).
    This solves the problem of the changing domain.

    Finally, we need that the limiting matrix
    \(\magenta{\Sigma^{\black{v},\infty}_{[0:\timestep)}}\) is in fact a covariance matrix of
    \((\rf_\dims, \nabla\rf_\dims)\) and not just a limit of covariance
    matrices. For this we apply \ref{ind: representation}, which
    ensures that the limiting scalar products \eqref{eq: <X, v> converges to gamma}
    and distances \eqref{eq: limiting distances} which make up
    \(\magenta{\Sigma^{\black{v},\infty}_{[0:\timestep)}}\) (cf.~Lemma~\ref{lem:
    cov of derivatives, non-stationary isotropy}) are realizable by points
    \(y_0,\dots, y_{\timestep-1}\in\real^{\dimV_\timestep}\).
    \(\magenta{\Sigma^{\black{v},\infty}_{[0:\timestep)}}\) is therefore a
    covariance matrix of \((\rf_\dims, \nabla\rf_\dims)\) viewed as a random function with
    domain \(\real^{\dimV_\timestep}\). 
    Since the distances \(\rho_{ij}\) are positive by \ref{ind: evaluation
    points asymptotically different}, the asymptotic representations \(y_k\) are
    distinct and their covariance matrix
    \(\magenta{\Sigma^{\black{v},\infty}_{[0:\timestep)}}\) is therefore strictly
    positive definite by the strict positive definiteness of \((\rf_\dims, \nabla\rf_\dims)\).
\end{proof}
The matrix inverse is therefore continuous in
\(\magenta{\Sigma^{\black{v},\infty}_{[0:\timestep)}}\) and by the convergence
of the matrices in probability and the convergence of \(Z(\rv_{[0:\dimV_\timestep)};
\Param_{[0:\timestep)})\) by induction assumptions \ref{ind: information convergence} for
\(\timestep-1\) restated in \eqref{eq: main induction assumption}, the
conditional expectation \eqref{eq: relevant part of cond expec} converges in
probability
\begin{align}
    \nonumber
    &\E[Z(\rv_{[0:\dimV_\timestep)}; \Param_\timestep) \mid \filt_{\timestep-1}]
    \\
    \label{eq: converging conditional expectation}
    \underset{\dims\to\infty}&{\overset{p}\to}
    \mu_\timestep^{v, \infty} + \blue{\Sigma^{v,\infty}_{[0:\timestep),\timestep-1}}^T
    [\magenta{\Sigma^{v,\infty}_{[0:\timestep)}}]^{-1}
    \Bigl(
        \begin{pmatrix}
            \limf_{[0:\timestep)}\\
            \gamma_{[0:\timestep)}
        \end{pmatrix}
        -\mu_{[0:\timestep)}^{v, \infty}
    \Bigr)
    =: \begin{pmatrix}
        \limf_\timestep
        \\
        \gamma_\timestep^{([0:\dimV_\timestep))}
    \end{pmatrix}.
\end{align}
Together with the conditional expectation of the directional derivatives in the
directions \(\rw_i\) in \eqref{eq: conditional expectation in direction w}, we
have now fully determined the conditional expectation.

An analysis of the conditional variance will immediately give us an understanding
of the distribution and we can therefore put these considerations together
to obtain convergence of \(Z\) itself.

\paragraph*{\ref{it: step 1 induction step}: Convergence of \(Z(\rv_{[0:\dimV_\timestep)}; \Param_\timestep)\)}

Recall, we outlined at the start of the induction that we would first prove
convergence of \(Z(\rv_{[0:\red{\dimV_\timestep})}; \Param_\timestep)\) (i.e.\
\ref{it: step 1 induction step}) before proving \(V_{\timestep+1}\) to
asymptotically have full rank (\ref{it: step 2 induction step}) and the
convergence of the new corner element
\(D_{\rv_{\dimV_\timestep}}\rf_\dims(\Param_\timestep)\) (\ref{it: step 3 induction
step}).

In \eqref{eq: converging conditional expectation} we have found the limit of the
conditional expectation of \(Z(\rv_{[0:\dimV_\timestep)}; \Param_\timestep)\).
This limit is also going to be the limit of the new column
\(Z(\rv_{[0:\dimV_\timestep)}; \Param_\timestep)\) itself. 
To prove this we only need to control the variance. For this purpose we recall that the
unconditional variance is given by 
\[
    \green{\rcov^\dims_\timestep}
    = \begin{pmatrix}
        \green{\rcov^{\black{v},\dims}_\timestep}
        \\
        & &  \kernel_3\bigl(\tfrac{\|\Param_\timestep\|^2}2,\tfrac{\|\Param_\timestep\|^2}2, \|\Param_\timestep\|^2\bigr) \identity_{(\dims-\dimV_\timestep)\times (\dims-\dimV_\timestep)}
    \end{pmatrix}.
\]
The conditional covariance matrix is therefore given by
\begin{align}
    \nonumber
    &\Cov[
        Z(\rv_{[0:\dimV_\timestep)},\rw_{[\dimV_\timestep:\dims)}; \Param_\timestep)
        \mid \filt_{\timestep-1}
    ]
    \\
    \label{eq: conditional covariance}
    &= \frac1\dims \begin{bmatrix}
        \green{\rcov^{\black{v},\dims}_\timestep}
        - \blue{\rcov^{\black{v},\dims}_{[0:\timestep),\timestep}}^T
        [\magenta{\rcov^{\black{v},\dims}_{[0:\timestep)}}]^{-1}
        \blue{\rcov^{\black{v},\dims}_{[0:\timestep),\timestep}}
        \\
        & \sigma_{w,d}^2
        \identity_{(\dims-\dimV_\timestep)\times (\dims-\dimV_\timestep)}
    \end{bmatrix},
\end{align}
where the conditional variance in the direction of the \(\rw_i\) is given by
\begin{equation}
    \label{eq: residual variance}
    \sigma_{w,\dims}^2
    \coloneqq
        \green{\kernel_3\bigl(\tfrac{\|\Param_\timestep\|^2}2,\tfrac{\|\Param_\timestep\|^2}2, \|\Param_\timestep\|^2\bigr)}
        - \blue{\rcov^{\black{w},\dims}_{[0:\timestep),\timestep}}^T
        [\magenta{\rcov^{\black{w},\dims}_{[0:\timestep)}}]^{-1}
        \blue{\rcov^{\black{w},\dims}_{[0:\timestep),\timestep}}.
\end{equation}
We will need \(\sigma_{w,\dims}^2\) for \ref{it: step 2 induction step} and
\ref{it: step 3 induction step}, but for now it is not important.
Due to the diagonal structure we have for our new column
\(Z(\rv_{[0:\red{\dimV_\timestep})}; \Param_\timestep)\) by \eqref{eq:
decomposition in distribution}
\begin{align}
    \nonumber
    &Z(\rv_{[0:\red{\dimV_\timestep})}, \Param_\timestep)
    \\
    \nonumber
    &= \E[Z(\rv_{[0:\dimV_\timestep)}, \Param_\timestep) \mid \filt_{\timestep-1}]
    + \sqrt{\Cov[Z(\rv_{[0:\dimV_\timestep)}, \Param_\timestep) \mid \filt_{\timestep-1}]} \begin{pmatrix}
        Y_0\\ \vdots\\ Y_{\dimV_\timestep}
    \end{pmatrix}
    \\
    \nonumber
    &= \E[Z(\hat{v}_{[0:\dimV_\timestep)}, \Param_\timestep) \mid \filt_{\timestep-1}]
    + \frac1{\sqrt{\dims}} \sqrt{
        \green{\rcov^{\black{v},\dims}_\timestep}
        - \blue{\rcov^{\black{v},\dims}_{[0:\timestep),\timestep}}^T
        [\magenta{\rcov^{\black{v},\dims}_{[0:\timestep)}}]^{-1}
        \blue{\rcov^{\black{v},\dims}_{[0:\timestep),\timestep}}
        }\begin{pmatrix}
        Y_0\\ \vdots\\ Y_{\dimV_\timestep}
    \end{pmatrix}
    \\
    &\underset{\dims\to\infty}{\overset{p}\to}
    \begin{pmatrix}
        \limf_\timestep
        \\
        \gamma_\timestep^{([0:\dimV_\timestep))}
    \end{pmatrix}.
    \label{eq: convergence new col}
\end{align}
This is exactly the convergence of the `new column' required for \ref{it: step 1 induction step}.

\paragraph*{\ref{it: step 2 induction step}: The rank of \(V_{\timestep+1}\)}

The last two steps follow fairly quickly. Recall that \(V_{\timestep+1}\) is the
\emph{previsible} running span of evaluation points defined in \eqref{eq: vector
space of evaluation points}. Since it is previsible, it only includes gradients
up to time \(\timestep\) and the Gram-Schmidt candidate \eqref{eq: definition of
v candidate} of its most recent addition is therefore given by
\[
    \tilde{\rv}_\timestep := \nabla\rf_\dims(\Param_\timestep) - P_{V_\timestep} \nabla\rf_\dims(\Param_\timestep)
    = \sum_{k=\dimV_\timestep}^{\dims-1} \langle \rw_i, \nabla\rf_\dims(X_\timestep)\rangle \rw_i
    = \sum_{k=\dimV_\timestep}^{\dims-1} D_{\rw_i}\rf_\dims(X_\timestep) \rw_i,
\]
where \(P_{V_\timestep}\) is the projection onto \(V_\timestep\).

Now we naturally want to analyze the directional derivatives
\(D_{\rw_i}\rf_\dims(X_\timestep)\) which make up \(\tilde{\rv}_\timestep\).
By representation \eqref{eq: decomposition in distribution} and our
formula for the conditional covariance \eqref{eq: conditional covariance}
we have in distribution
\[
    D_{\rw_i}\rf_\dims(\Param_\timestep)
    = \underbrace{\E[D_{\rw_i}\rf_\dims(\Param_\timestep)\mid \filt_{\timestep-1}]}_{\overset{\eqref{eq: conditional expectation in direction w}}=0}
    + \sqrt{\tfrac1\dims \sigma_{w,\dims}} Y_{i+1}.
\]
We now already see where our law of large numbers is going to come from. But we
first need to take a closer look at the residual variance \(\sigma_{w,\dims}^2\) defined
in \eqref{eq: residual variance}. We get convergence in probability of the residual variance
to a value strictly greater zero
\begin{align}
    \label{eq: limiting residual variance}
    \sigma_{w,\dims}^2
    &= \bigl(
        \green{\kernel_3\bigl(\tfrac{\|\Param_\timestep\|^2}2,\tfrac{\|\Param_\timestep\|^2}2, \|\Param_\timestep\|^2\bigr)}
        - \blue{\rcov^{\black{w},\dims}_{[0:\timestep),\timestep}}^T
        [\magenta{\rcov^{\black{w},\dims}_{[0:\timestep)}}]^{-1}
        \blue{\rcov^{\black{w},\dims}_{[0:\timestep),\timestep}}
    \bigr)
    \\
    \nonumber
    \underset{\dims\to\infty}&{\overset{p}\to}
    \bigl(
        \green{\kernel_3\bigl(\tfrac{\|y_\timestep\|^2}2,\tfrac{\|y_\timestep\|^2}2, \|y_\timestep\|^2\bigr)}
        - \blue{\Sigma^{\black{w},\infty}_{[0:\timestep),\timestep}}^T
        [\magenta{\Sigma^{\black{w},\infty}_{[0:\timestep)}}]^{-1}
        \blue{\Sigma^{\black{w},\infty}_{[0:\timestep),\timestep}}
    \bigr)
    =:\sigma_{w,\infty}^2 > 0,
\end{align}
using the convergence of the block matrices and the following lemma.

\begin{lemma}\label{lem: strict postive definite w direction}
    \(\magenta{\Sigma^{\black{w},\infty}_{[0:\timestep)}}\) is strictly positive definite
    and \(\sigma_{w,\infty}^2>0\).
\end{lemma}
\begin{proof}
    We are going to show with a similar argument as in Lemma~\ref{lem: strict
    positive cov matrix} that the matrix
    \[
        \Sigma^{w,\infty}_{[0:\timestep]}
        := \begin{bmatrix}
            \magenta{\Sigma^{\black{w},\infty}_{[0:\timestep)}}     
            & \blue{\Sigma^{\black{w},\infty}_{[0:\timestep),\timestep}}
            \\
            \blue{\Sigma^{\black{w},\infty}_{[0:\timestep),\timestep}}^T
            & \green{\kernel_3\bigl(\tfrac{\|y_\timestep\|^2}2,\tfrac{\|y_\timestep\|^2}2, \|y_\timestep\|^2\bigr)}
        \end{bmatrix}
    \]
    is strictly positive definite. Before we do so, let us quickly argue why this
    finishes the proof. Since \(\Sigma^{w,\infty}_{[0:\timestep]}\) is then strictly positive 
    definite, it has a cholesky decomposition \(L\) such that
    \[
        \Sigma^{w,\infty}_{[0:\timestep]} = LL^T
        = \begin{bmatrix}
            L_\timestep & 0
            \\
            l^T & \sigma
        \end{bmatrix}
        \begin{bmatrix}
            L_\timestep^T & l
            \\
            l &\sigma
        \end{bmatrix},
    \]
    which implies that \(L_\timestep\) is the cholesky decomposition of \(\magenta{\Sigma^{\black{w},\infty}_{[0:\timestep)}}\),
    \(l = L_\timestep^{-1}\blue{\Sigma^{\black{w},\infty}_{[0:\timestep),\timestep}}\) and
    \[
        \sigma = \sqrt{ 
            \green{\kernel_3\bigl(\tfrac{\|y_\timestep\|^2}2,\tfrac{\|y_\timestep\|^2}2, \|y_\timestep\|^2\bigr)}
            - \blue{\Sigma^{\black{w},\infty}_{[0:\timestep),n}}^T
            [\magenta{\Sigma^{\black{w},\infty}_{[0:\timestep)}}]^{-1}
            \blue{\Sigma^{\black{w},\infty}_{[0:\timestep),n}}
        }
        = \sigma_{w,\infty}.
    \]
    Since we have
    \[
        \det(L) = \sigma_{w,\infty}\det(L_\timestep),
    \]
    strict positive definiteness of \(\Sigma^{w,\infty}_{[0:\timestep]}\) and therefore \(0\neq\det(\Sigma^{w,\infty}_{[0:\timestep]}) = \det(L)^2\)
    implies
    \[
        \sigma_{w,\infty}^2 > 0
        \quad \text{and} \quad 
        \det(L_\timestep)\neq 0.
    \]
    But \(\det(L_\timestep)\neq 0\) also implies that
    \(\magenta{\Sigma^{\black{w},\infty}_{[0:\timestep)}}\) has to be strictly
    positive definite, since \(L_\timestep\) is its cholesky decomposition.

    What is therefore left to prove is the strict positive definiteness of
    \(\Sigma^{w,\infty}_{[0:\timestep]}\). For this note that
    \[
       \rcov^{w,\dims}_{[0:\timestep]} 
        := \begin{bmatrix}
            \magenta{\rcov^{\black{w},\dims}_{[0:\timestep)}}     
            & \blue{\rcov^{\black{w},\dims}_{[0:\timestep),n}}
            \\
            \blue{\rcov^{\black{w},\dims}_{[0:\timestep),n}}^T
            & \green{\kernel_3\bigl(\tfrac{\|\Param_\timestep\|^2}2,\tfrac{\|\Param_\timestep\|^2}2, \|\Param_\timestep\|^2\bigr)}
        \end{bmatrix}
    \]
    is the plug-in covariance matrix of \(D_{\rw_i}\rf_\dims(\Param_{[0:\timestep]})\).
    Where we use the term "plug-in" covariance to say that we treat the
    evaluation points \(\Param_{[0:\timestep]}\) and the direction \(\rw_i\) as
    deterministic. Since \(\Param_{[0:\timestep]}\) are contained in \(V_\timestep\),
    we again map them isometrically to \(\real^{\dimV_\timestep}\). But this time
    we view \(\real^{\dimV_\timestep}\) as a subspace of
    \(\real^{\dimV_\timestep+1}\) and map the additional vector \(\rw_i\) to
    \(e_{\dimV_\timestep+1}\) such that \(\rcov^{w,\dims}_{[0:\timestep]}\)
    is a covariance matrix of \(\nabla\rf_\dims\) in
    \(\real^{\dimV_\timestep+1}\).
    
    Now we finish the proof with the same limiting argument as in
    Lemma~\ref{lem: strict positive cov matrix}.  Since
    \(\nabla\rf_\dims\) is strictly positive definite by
    assumption, we get that \(\Sigma_{[0:\timestep]}^{w,\infty}\) is strictly
    positive definite as the covariance matrix of
    \((\partial_{\dimV_\timestep+1}\rf_\dims)\) at the points
    \(y_{[0:\timestep]} = (y_0, \dots, y_\timestep) \subseteq
    \real^{\dimV_\timestep+1}\) of \ref{ind: representation}. These are the
    limiting representations and non of the points \(y_k\) are equal by
    \ref{ind: evaluation points asymptotically different}.
\end{proof}

Using the convergence of the residual variance \(\sigma_{w,\dims}^2\to
\sigma_{w,\infty}^2\) allows us to finally make our law of large
numbers argument
\begin{equation}
    \label{eq: v_n tilde}
    \|\tilde{\rv}_\timestep\|^2
    = \sum_{i=\dimV_\timestep}^{\dims-1} (D_{\rw_i}\rf_\dims(X_\timestep))^2
    = \frac{\sigma_{w,\dims}^2}\dims\sum_{i=\dimV_\timestep+1}^{\dims}  Y_i^2 
    \;\underset{\dims\to\infty}{\overset{p}\to} \;
    \sigma_{w,\infty}^2  > 0.
\end{equation}
This implies that \(V_\timestep\) has full rank asymptotically \ref{ind: full
dimension}.

Assuming the last gradient is always used (and not just in the asymptotic limit),
we can get \(\dimV_{\timestep+1} = \dimV_\timestep+1\) almost surely,
since \(\sigma_{w,\dims}^2> 0\) almost surely is sufficient for
\(\|\tilde{\rv}_\timestep\|^2>0\) almost surely. The use of the
most recent gradient ensures the points \(\Param_k\) are different by an inductive
argument similar to Lemma~\ref{lem: conv info, full rank -> different eval pts}.
This in turn ensures, using the strict positive definiteness of
\((\rf_\dims,\nabla\rf_\dims)\), that \(\sigma_{w,\dims}^2> 0\) almost surely with a similar
argument as in Lemma~\ref{lem: strict postive definite w direction}.

\paragraph*{\ref{it: step 3 induction step}: Convergence of the new corner element}

The last step follows immediately by definition of \(\rv_{\dimV_\timestep} =
\frac{\tilde{\rv}_\timestep}{\|\tilde{\rv}_\timestep\|}\) in
Definition~\ref{def: previsible orthonormal coordinate system}
and \eqref{eq: v_n tilde}
\[
    D_{\rv_{\dimV_\timestep}}\rf_\dims(X_\timestep)
    = \bigl\langle
        \nabla\rf_\dims(X_\timestep),
        \tfrac{\tilde{\rv}_\timestep}{\|\tilde{\rv}_\timestep\|}
    \bigr\rangle
    = \|\tilde{\rv}_\timestep\|
    \underset{\dims\to\infty}{\overset{p}\to}
    \sigma_{w,\infty} =: \gamma_{\timestep}^{(\dimV_\timestep)}
    > 0.
\]

 \section{Discussion and Outlook}\label{sec: outlook}

In this section we want to discuss limitations, possible generalizations and
extensions of our results. Let us begin by with the only structural assumption:

\begin{enumerate}[label={S\arabic*.},ref={S\arabic*}]
	\item \textbf{Isotropy.} This assumption is the key argument to get independent,
	identically distributed partial derivatives and thereby a law of large numbers
	of the gradient norm. Isotropic kernels are
	characterized by the property \(\C(x,y) = \C(Ux, Uy)\) for any orthogonal
	matrix \(U\) \citep{benningSchoenbergCharacterizationContinuous2025}.
	It is natural to ask how small the set of transformations \(U\) may be before
	our results break down. In place of rotation invariance, one might consider
	exchangeable directions for example (which is captured by the set of
	dimension permutations). In the first step this would likely yield
	exchangeable partial derivatives and and thereby a law of large numbers.
	However our proof approach would already break down in the second step as we
	are using an adapted coordinate system and not the standard basis. We are 
	doubtful whether these results even hold in a more general
	setting.
\end{enumerate}

Next we want to discuss the assumptions that may be removed or relaxed.
We sort them by our current perception of their difficulty in increasing order:

\begin{enumerate}[label={A\arabic*.},ref={A\arabic*}]
	\item\label{it: strict positive def removal}
	\textbf{Strict positive definiteness and use of the most recent gradient}.
	The strict positive definiteness was used in the proof of Theorem~\ref{thm:
	asymptotically deterministic behavior variant} to show that the limiting
	covariance matrices are invertible. Without this assumption
	one would either have to work with generalized matrix inverses and
	prove that these are still continuous in their entries or use an approach
	that avoids matrix inverses altogether such as describing the conditional
	distribution using characteristic functions.

	Note that we needed the evaluation points to be different \ref{ind:
	evaluation points asymptotically different} for
	strict positive definiteness to apply and we used the assumption
	that the most gradient was used for this purpose. This also
	ensured that the span of gradients \(V_\timestep\) had full rank
	\ref{ind: full dimension}. Removing the strict positive definiteness
	and the use of the most recent gradient by the algorithm would therefore also remove
	these corollary implications.

	\item \textbf{Finite time horizon.} We only show convergence of the iterates
	up to a fixed time horizon \(n\). This is structural to our qualitative approach
	as we need convergence of the previous step to show convergence of the next step.
	There are two potential ways to allow \(n\to \infty\) jointly with the
	dimension \(\dims\):
	\begin{enumerate}
		\item  \textbf{Quantitative convergence and stability results.}
		The proof would be similar to proving convergence of an
		ODE discretization. In every step errors are accumulated and passed through
		continuous functions (which would need to be Lipschitz or similar
		in a quantitative setting). This error accumulation would need to happen
		slowly enough that the overall error still converges even if the number
		of error accumulation steps grows to infinity. One can observe
		empirically that the variance does not increase with the number of steps
		\(n\) (cf.~Figure~\ref{fig: simulated gradient descent}), which means
		that this is likely possible.

		\item \textbf{Big \(n\) handover.} Since the iterates converge
		for every finite but possible very large time horizon \(n\), it is
		perhaps possible to prove that for very large \(n\) it is overwhelmingly
		likely that the iterates are captured by a local minimum and are therefore
		constant anyway. This approach is likely significantly harder: One would
		first need to prove that the optimizer reaches a certain level at time
		\(n\), which requires \textbf{explicit bounds on the \(\limf_n\)} and
		then combine this with topological results about GRFs to show
		that at a certain level the function only has convex regions close
		to minima. These topological results are only partially
		available for the special case of spin glasses
		\citep[e.g.][]{auffingerComplexityGaussianRandom2023}.
	\end{enumerate}

	\item \textbf{Gaussian random function.} Our argument is
	based on a law of large numbers applied to the norm of
	gradients. For this we needed that the \emph{squared} directional
	derivatives are uncorrelated. We obtained this result from the
	fact that the directional derivatives themselves are uncorrelated.
	In the Gaussian case they are thereby independent and their squares
	are therefore also uncorrelated. To remove the Gaussian assumption
	one would therefore have to obtain uncorrelated squares directly from
	isotropy. The difficulty then lies in pushing this argument through the steps,
	which may turn out to be straightforward or very difficult.

	\item\label{it: assumption gsa} \textbf{Gradient Span Algorithms.} In practice many algorithms leave
	the linear span of gradients with the use of preconditioning matrices.
	Example include Shampoo \citep{guptaShampooPreconditionedStochastic2018}
	or Adam \citep{kingmaAdamMethodStochastic2015}, which uses a diagonal
	preconditioning matrix of entry-wise learning rates.
	While it is likely that similar results hold for these type of algorithms
	based on their empirical behavior (cf.~Figure~\ref{fig: loss plots}),
	it will be very difficult to prove this theoretically. Observe that a key
	step in our proof is to show that the `learning rates' converge. More
	specifically we show that the coefficients of the evaluation locations with
	respect to the adapted span converge. This is necessary to ensure a
	deterministic behavior of the optimizer in the limit. For algorithms
	using preconditioners a similar result would need to be proven. However
	it is unclear in what sense a preconditioning matrix of increasing
	size may converge. Random matrix theory answers this question by stating
	that the distribution of eigenvalues converges. This approach loses the
	order of eigenvalues however, which is a problem here since they are
	highly correlated to the gradient realizations of the random function.
	Only after a sense in which preconditioning matrices may
	converge is established one can hope to extend our results to this
	generality.

	Perhaps the key lies in the fact that we are not considering arbitrary
	preconditioning matrices. Specifically, the authors conjecture in a
	different paper \citep[Appendix~E.1]{benningRandomFunctionDescent2024} that the
	purpose of the preconditioning matrix is to turn an anisotropic Random
	Function into an isotropic one and that preconditioning is unnecessary
	on isotropic functions.
\end{enumerate}

Perhaps even more interesting than the relaxation of assumptions would
be the following extensions:

\begin{enumerate}[label={E\arabic*.},ref={E\arabic*}]
	\item An analysis of \textbf{stochastic gradient descent} and similar
	stochastic algorithms which do not have access to the true cost but
	rather noisy evaluations of \(\rf_\dims\). It is again unclear in what
	sense one would obtain convergence, since we cannot expect convergence
	of the coordinates in the adapted span. Perhaps it is possible to prove
	convergence against a distribution of coordinates in the adapted span and
	thereby obtain convergence against a distribution of function values.
	Maybe ideas from Langevin dynamics can be reused here.
	
	\item \textbf{Bounds on the limiting function values \(\limf_n\)}
	for a given optimizer and random function distribution would be extremely
	useful for comparing optimizers. Perhaps it is even possible to find the
	`best optimizer' by optimizing over these limiting function values. This
	would be the most useful extension but also the most difficult. The
	optimizer that is one-step optimal is `Random Function Descent'
	\citep{benningRandomFunctionDescent2024}, but even the two-step optimal
	optimizer is very difficult to find as one has to find structure in
	in increasingly large covariance matrices.
\end{enumerate}

\section*{Acknowledgements}
\addcontentsline{toc}{section}{Acknowledgements}

This research was supported by the RTG 1953, funded by the German Research
Foundation (DFG). The authors would like to thank Yan Fyodorov for his
hospitality and helpful discussions at King's College. We thank him along with
Mark Sellke
and Antoine Maillard
for their time and valuable insights on spin glasses. 
\bibliography{references}

\appendix

\section{Random quadratic functions are isotropic}
\label{sec: random quadratic functions}

The goal of this section is twofold: First, we want to motivate that
the cost functions found in machine learning should be expected to
be non-stationary. Second, we want to demonstrate how random linear
models result in (isotropic) random quadratic functions. This suggests
that the isotropy assumption may be plausible.

Subject to strong simplifying assumptions, \(p\)-layer neural networks
may also be related to \(p\)-spin glasses \eqref{eq: p-spin glass}
\citep{choromanskaLossSurfacesMultilayer2015}, i.e.\ isotropic random functions
on the sphere. Much more realistic assumptions are sufficient to relate linear
models to random quadratic functions. More specifically, the random functions
considered by \citet{paquetteHaltingTimePredictable2022}, etc.\ are of the form
\[
	\rf_\dims(\param)
	= \frac1{2n}\|\dataMat(\param - \signal) + \noise \|^2
	= \frac1{2n}\sum_{i=1}^n(\dataMat_{i,\cdot}(\param - \signal) + \noise_i)^2
	= \frac1n \sum_{i=1}^n \rf_\dims^{(i)}(\param),
\]
where \(\dataMat\in\real^{n\times \dims}\) is a random data matrix,
\(\signal\in\real^\dims\) is a random vector representing the true signal and
\(\noise\) is noise. The last equation shows how \(\rf_\dims\)
can be decomposed into \(n\) stochastic losses \(\rf_\dims^{(i)}\), where \(n\) is
the amount of data.
\begin{restatable}[Quadratic function representation]{lemma}{quadFunRepresentation}
	Assume \(\dataMat\), \(\signal\) and \(\noise\) to be independent, where
	the noise \(\noise\) itself is a vector of iid variables with zero mean
	and variance \(\sigma_\noise^2\). Further assume the isotropic feature model
	of \citet{paquetteHaltingTimePredictable2022}, i.e. the entries of
	\(\dataMat\) are iid with zero mean and variance \(\sigma_{\dataMat}^2\).
	Then \(\rf_\dims\) can be represented as
	\[
		\rf_\dims(\param)
		= \rf_\dims^\infty(\param) + \frac1n \sum_{i=1}^n \underbrace{(\rf_\dims^{(i)}(\param) - \rf_\dims^\infty(\param))}_{=:\epsilon_i(\param)}
		\quad\text{with}\quad
		\rf_\dims^\infty(\param)
		= \tfrac{\sigma_{\dataMat}^2}2 \|\param-\signal\|^2 + \tfrac{\sigma_{\noise}^2}2,
	\]
	where \(\rf_\dims^\infty\) is the `infinite data limit' and \(\epsilon_i\) is the stochastic
	noise, which is identically distributed. Conditional on the true signal
	\(\signal\), the \(\epsilon_i\) are also independent with zero mean. The
	noise \(\epsilon_i\) is then also unconditionally uncorrelated and has zero mean.
\end{restatable}
\begin{proof}
	Before confirming it has the desired properties, we simply define the infinite
	data limit to be
	\begin{align*}
		\rf_\dims^\infty(\param)
		&:= \E[\rf_\dims^{(i)}(\param) \mid \signal]
		= \E\Bigl[
			\tfrac12(\param-\signal)^\transpose\dataMat_{i,\cdot}^\transpose\dataMat_{i,\cdot}(\param-\signal)
			+\dataMat_{i,\cdot}(\param-\signal)\noise_i 
			+ \tfrac12\noise_i^2 \mid \signal
		\Bigr]
		\\
		&= \tfrac12(\param-\signal)^\transpose \underbrace{\E\bigl[
			\dataMat_{i,\cdot}^\transpose\dataMat_{i,\cdot}
		\bigr]}_{\sigma_\dataMat^2 \identity}(\param-\signal)
		+ \tfrac12 \underbrace{\E\bigl[\noise_i^2 \bigr]}_{\sigma_\noise^2}
		\\
		&= \tfrac{\sigma_{\dataMat}^2}2 \|\param-\signal\|^2 + \tfrac{\sigma_{\noise}^2}2.
	\end{align*}
	By the independence of \(\dataMat\), \(\signal\), and \(\noise\) and the
	independence of the entries of \(\dataMat\), conditionally on \(\signal\),
	the \(\rf_\dims^{(i)}(\param)\) are independent. Since \(\rf_\dims^\infty\) is deterministic,
	conditional on \(\signal\) by definition, the \(\epsilon_i\) are therefore
	independent, conditionally on \(\signal\). That the conditional mean is
	zero follows from our definition of \(\rf_\dims^\infty\). The unconditional
	statements follow from the tower property of the conditional expectation.
\end{proof}

Since we do not consider stochastic optimization algorithms (like stochastic
gradient descent) but full gradient optimization in this paper, a comparison requires a comparison with the underlying infinite data limit
\(\rf_\dims^\infty\), which can be rewritten as
\[
	\rf_\dims^\infty(\param)
	= \underbrace{\tfrac{\sigma_{\noise}^2}2 + \tfrac{\sigma_{\dataMat}^2}2\bigl(\E[\|\signal\|^2]+\|\param\|^2\bigr)}_{\text{deterministic}}
	+ \sigma_{\dataMat}^2\langle \param, \signal\rangle
	+ \underbrace{\tfrac{\sigma_\dataMat^2}2\bigl(\|\signal\|^2-\E[\|\signal\|^2]\bigr)}_{\text{random constant}}.
\]
We are now going to assume that the signal \(\signal\) is centered (which represents
a translation of the coordinate system such that \(\E[\signal]\) is the origin). Let us assume 
assume, like \citet[Assumption 1]{paquetteHaltingTimePredictable2022}, that the entries \(\signal^{(i)}\)
are independent with variance \(\frac{R^2}\dims\). The law of large numbers then implies
\[
	\tfrac{\sigma_\dataMat^2}2\bigl(\|\signal\|^2-\E[\|\signal\|^2]\bigr)
	\underset{\dims\to\infty}{\overset{p}\to} 0.
\]
Hence, the random constant disappears in the high-dimensional limit. Since constants are also irrelevant for optimization there are two good
reasons why this random constant can be dropped  without loss of generality when
analyzing optimization on \(\rf_\dims^\infty\). But then, assuming \(\signal\)
to be Gaussian, the infinite data limit
\[
	\hat{\rf}_\dims^\infty(\param):= 
	\mu\bigl(\tfrac{\|\param\|^2}2\bigr)
	+ \sigma_{\dataMat}^2\langle \param, \signal\rangle,
	\quad\text{with}\quad
	\mu\bigl(\tfrac{\|\param\|^2}2\bigr)
	:=\tfrac{\sigma_{\noise}^2}2 + \tfrac{\sigma_{\dataMat}^2}2\bigl(\E[\|\signal\|^2]+\|\param\|^2\bigr),
\]
is a (non-stationary) isotropic GRF with covariance
\[
	\Cov(\hat{\rf}_\dims^\infty(x), \hat{\rf}_\dims^\infty(y))
	= \sigma_{\dataMat}^4 \tfrac{R^2}\dims \langle x, y\rangle
	=: \tfrac1\dims\kernel\bigl(\tfrac{\|x\|^2}2,\tfrac{\|y\|^2}2, \langle x,y\rangle\bigr).
\]
  Random
quadratic functions with isotropic features are therefore covered by the
(non-stationary) isotropy assumption of the present article.
The additional assumption we required over \citet{paquetteHaltingTimePredictable2022}
was that the mean of \(\signal\) had to be zero. By a change of coordinate
system this essentially implies the origin of the space necessarily has
to be the expectation of \(\signal\).

\begin{remark}
	The `correlated feature model' of
	\citet{paquetteHaltingTimePredictable2022} is essentially a change of space. 
	That is, (non-stationary) isotropy with respect to the Hilbertspace \(\real^\dims\)
	equipped with the inner product
	\[
		\langle x,y\rangle_{\Sigma} := \langle x, \Sigma y\rangle
	\]
	results in the correlated feature model. This geometric modification of
	(non-stationary) isotropy is well known in the stationary isotropic case
	under the term `geometric anisotropy'
	\citep[e.g.][p.~17]{steinInterpolationSpatialData1999}.
\end{remark}

 \section{Conditional Gaussian distributions}
\label{sec: conditional gaussian distribution}

Theorem~\ref{thm: conditional gaussian distribution} about conditional Gaussian
distributions is well known
\citep[e.g.][Prop.~3.13.]{eatonMultivariateStatisticsVector2007}, but for your
convenience we wrote down our favorite proof of this theorem capturing
the intuition behind the statement.

\begin{theorem}[Conditional Gaussian distribution]
    \label{thm: conditional gaussian distribution}
    Let \(X\sim\normal(\mu,\Sigma)\) be a multivariate Gaussian vector where
    the covariance matrix is a block matrix of the form
    \[
        \mu = \begin{bmatrix}
            \mu_1\\ \mu_2
        \end{bmatrix}
        \quad \text{and} \quad
        \Sigma = \begin{bmatrix}
            \Sigma_{11} & \Sigma_{12}
            \\
            \Sigma_{21} & \Sigma_{22}
        \end{bmatrix},
    \]
    then the conditional distribution of \(X_2\) given \(X_1\) is
    \[
        X_2\mid X_1 \sim \normal(\mu_{2\mid 1}, \Sigma_{2\mid 1}),
    \]
    with conditional mean and variance
    \begin{align*}
        \mu_{2\mid 1} &:= \mu_2 + \Sigma_{21}\Sigma_{11}^{-1}(X_1-\mu_1)
        \\
        \Sigma_{2\mid 1} &:= \Sigma_{22} -  \Sigma_{21}\Sigma_{11}^{-1}\Sigma_{12}.
    \end{align*}
    where \(\Sigma_{11}^{-1}\) may be a generalized inverse
    \citep[cf.][]{eatonMultivariateStatisticsVector2007}.
\end{theorem}
\begin{proof}
    For the general statement we point to the standard literature
    \citep[e.g.][]{eatonMultivariateStatisticsVector2007}.
    For this proof we will assume for simplicity that \(\Sigma\) is invertible.

    Let \(\bar{X} := X-\mu\) be the centered version of \(X\). Then there exists
    a unique lower triangular matrix \(L\) such that \(\Sigma = LL^T\) (i.e.\ the Cholesky
    Decomposition). This results in the following representation\footnote{
        It is straightforward to check that \(Y := L^{-1}(X-\mu)\)
        is a multivariate Gaussian vector of centered, uncorrelated entries with variance \(1\).
        They are therefore \(\iid\) standard normal since uncorrelated multivariate Gaussian
        random variables are independent. Sometimes this representation is even
        taken as the definition of a multivariate normal distribution.
        This is the only real place we use the invertibility of \(\Sigma\) in
        its entirety, the invertibility of \(\Sigma_{11}^{-1}\) is used later on
        because we do not want to get into the business of generalized inverses
        here.
    }
    \[
        X - \mu =:
        \begin{bmatrix}
            \bar{X}_1
            \\
            \bar{X}_2
        \end{bmatrix}
        = \begin{bmatrix}
            L_{11} & 0
            \\
            L_{21} & L_{22}
        \end{bmatrix}
        \begin{bmatrix}
            Y_1 \\ Y_2
        \end{bmatrix}
        = LY
    \]
    with independent standard normal \(Y_i\), i.e. \(Y\sim\normal(0, \identity)\).
    \(L_{11}\) is invertible since \(\Sigma_{11}\) and therefore the map from
    \(Y_1\) to \(X_1\) is bijective. Conditioning on \(X_1\) is therefore
    equivalent to conditioning on \(Y_1\). But we have
    \begin{equation}
        \label{eq: underlying explicit decomposition}
        X_2 = \mu_2 + \bar{X}_2
        = \underbrace{\mu_2 + L_{21}Y_1}_{\text{conditional expectation}} + \underbrace{L_{22} Y_2}_{\text{conditional distribution}}
    \end{equation}
    So it follows that
    \[
        X_2\mid X_1 \sim  \normal(\mu_{2\mid 1}, \Sigma_{2\mid 1})
    \]
    with
    \begin{align*}
        \mu_{2\mid 1} &:= \E[X_2 \mid X_1]
        &=& \mu_2 + L_{21}Y_1
        \\
        \Sigma_{2\mid 1} &:= \Cov[X_2 \mid X_1] = \E\Bigl[ \bigl(X_2 - \E[X_2 \mid X_1]\bigr)^2 \mid X_1\Bigr]
        &=& L_{22}L_{22}^T.
    \end{align*}
    What is left to do, is to find a representation for the \(L_{ij}\)
    using the block matrices of \(\Sigma\). For this we observe
    \begin{equation}
        \label{eq: sigma to L conversion}
        \setlength\arraycolsep{4pt}
        \begin{bmatrix}
            \Sigma_{11} & \Sigma_{12}\\
            \Sigma_{21} & \Sigma_{22}
        \end{bmatrix}
        = \Sigma
        = LL^T = \begin{bmatrix}
            L_{11}L_{11}^T & L_{11}L_{21}^T
            \\
            L_{21}L_{11}^T & L_{22}L_{22}^T + L_{21}L_{21}^T
        \end{bmatrix}.
    \end{equation}
    Using \(Y_1 = L_{11}^{-1}\bar{X}_1\) and the insertion of an identity matrix this implies
    \[
        L_{21}Y_1 = L_{21}(L_{11}^T L_{11}^{-T})(L_{11}^{-1}\bar{X}_1)
        \overset{\eqref{eq: sigma to L conversion}}= \Sigma_{21}\Sigma_{11}^{-1}(X_1-\mu_1).
    \]
    resulting in the desired conditional expectation \(\mu_{2\mid 1}\). The conditional variance follows alike
    \begin{align*}
        \Sigma_{2\mid 1} = L_{22}L_{22}^T
        \overset{\eqref{eq: sigma to L conversion}}&= \Sigma_{22} - L_{21}L_{21}^T
        \\
        &= \Sigma_{22} - \underbrace{L_{21}(L_{11}^T}_{\overset{\eqref{eq: sigma to L conversion}}=\Sigma_{21}}\underbrace{L_{11}^{-T}) (L_{11}^{-1}}_{\overset{\eqref{eq: sigma to L conversion}}=\Sigma_{11}^{-1}}\underbrace{L_{11})L_{21}^T}_{\overset{\eqref{eq: sigma to L conversion}}=\Sigma_{12}}.
    \end{align*}
\end{proof}

\begin{remark}[Decomposition]
    \label{rem: decomposition}
    \(X_2\mid X_1 \sim \normal(\mu_{2\mid 1}, \Sigma_{2\mid 1})\) always implies
    that there exists there exists \(Y_2\sim\normal(0,\identity)\) independent
    of \(X_1\) such that the following equality is true in distribution
    \begin{equation}
        \label{eq: explicit decomposition}
        X_2 = \mu_{2\mid 1} + \sqrt{\Sigma_{2\mid 1}} Y_2,
    \end{equation}
    where \(\sqrt{\Sigma}\) denotes the cholesky decomposition of \(\Sigma\).
    If the covariance matrix is moreover invertible, then \(Y_2\) can
    be determined constructively as in the proof of Theorem~\ref{thm:
    conditional gaussian distribution} such that equation holds always and not
    just in distribution (cf.~\eqref{eq: underlying explicit decomposition}).
    This might be true in the non-invertible case as well
    but would require deeper insights about singular matrices.
\end{remark}
 \section{Strict positive definite derivatives}
\label{sec: strictly pos definite derivatives}

While covariance matrices are always positive definite, they are not always
strict positive definite (i.e. invertible).
Recall, that a random function \(\rf\) and its covariance function \(\C_{\rf}\)
are \emph{strict} positive definite if the matrix \((\C_{\rf}(x_k,
x_l))_{k,l=1,\dots, n}\) is strict positive definite for any finite
\(x_1,\dots, x_n\). It is already known that, whenever \(\rf\) is stationary isotropic
and non-constant, \(\rf\) is strict positive definite if the
dimension satisfies \(\dims \ge 2\)
\citep[Theorem~3.1.5]{sasvariMultivariateCharacteristicCorrelation2013}.
But since we require the `jet' \(\jet[1]\rf = (\rf, \nabla\rf)\)
to be strict positive definite in Theorem \ref{thm: asymptotically deterministic behavior}, we prove a generalization of Theorem~3.1.6 in
\citet{sasvariMultivariateCharacteristicCorrelation2013}. This
generalization will be sufficient to prove that all stationary isotropic
covariance functions which are valid in all dimensions (cf. Definition \ref{def: valid in all dimensions}) are covered as we will see in Corollary~\ref{cor:
strict positive definite}. This result will be used to remove
the strict positive definiteness assumption in Corollary~\ref{cor:
asymptotically deterministic behavior}.

If \(\rg\) is a random function with multivariate output, then \(\C_\rg(x_k, x_j)\) is already a
matrix for fixed \(k,j\) and the collection over \(k,j\) is really a tensor. To
avoid introducing this machinery and explaining positive definiteness for
tensors, we will take the following equivalent statement for positive
definiteness of random functions as definition.

\begin{definition}[Strict positive definite random function]
	\label{def: strict positive definite random function}
    The covariance \(\C_\rg\) of a random function \(\rg: \real^\dims\to\real^m\)
    and the random function itself is called \emph{strict positive definite}
    if for all \(w_k\in \real^m\) and distinct \(x_1,\dots, x_n\in \real^\dims\) the
    equality
    \[
        0 = \Var\Bigl[\sum_{k=1}^n w_k^T\rg(x_k)\Bigr]
        \overset{\eqref{eq: equivalent formulation}}= \sum_{k,l=1}^n w_k^T\C_\rg(x_k, x_l) w_l
    \]
    implies \(w_k = 0\) for all \(k\).
\end{definition}
Note, that the second equality marked with \eqref{eq: equivalent formulation} is
always true and only represents an equivalent formulation, because after
centering \(\rg\) (without loss of generality, using \(\tilde{\rg} = \rg -
\E[\rg]\)), we have
\begin{equation}
    \label{eq: equivalent formulation}
    \sum_{k,l=1}^n w_k^T \C_\rg(x_k,x_l) w_l
    = \E\Bigl[\sum_{k,l=1}^n w_k^T \rg(x_k) \rg(x_l)^T w_l\Bigr]
    = \Var\Bigl[ \sum_{k=1}^n w_k^T \rg(x_k)\Bigr].
\end{equation}

\begin{restatable}[Strict positive definite derivatives]{theorem}{strictlyPosDefiniteDerivatives}
    \label{thm: strict positive definite derivatives}
    Let \(\rf\colon \real^\dims \to \real\) be a stationary random function, i.e.\
    its covariance has the form \(\C_\rf(x,y) = \ikernel(x-y)\).  Assume
    the positive definite function \(\ikernel\) is continuous such that the support of its spectral measure
    contains a non-empty open set. Then up to any order \(k\) up to which
    \(\rf\) is almost surely differentiable, the jet
    \[
        \jet\rf
        = (
			\rf, \nabla\rf,
			(\partial_{ij}\rf)_{i\le j},
			\dots,
			(\partial_{i_1\cdots i_k} \rf)_{i_1\le \dots\le i_k}
		)
    \]
    is strict positive definite.
\end{restatable}
Theorem~\ref{thm: strict positive definite derivatives}
is not more general than
\citet[Theorem~3.1.6]{sasvariMultivariateCharacteristicCorrelation2013} in the
sense that conditions are weakend, but it is more general in its implication. That is, 
\(\rg\) is strict positive definite and not just \(\rf\).

The following corollary shows that this fully covers the stationary isotropic
covariance functions valid in all dimensions (cf.\ Definition \ref{def: valid in
all dimensions}). We use this result about stationary isotropic random functions
in Corollary~\ref{cor: asymptotically deterministic behavior}
to omit the assumption that \((\rf, \nabla\rf)\) has to be strict
positive definite, which is needed for our general result (Theorem~\ref{thm: asymptotically deterministic behavior}).

\begin{restatable}{corollary}{corStrictPosDef}
    \label{cor: strict positive definite}
	Assume that \(\ikernel\) is a continuous stationary isotropic covariance
	kernel valid in all dimensions, i.e.\ defined on the space of square
	summable sequences \(\ell^2\) by Lemma 2.3 in
	\citet{benningSchoenbergCharacterizationContinuous2025}. Let
	\(\rf\sim\normal(\mu, \ikernel)\) be a Gaussian random function, which is
	not almost surely constant. Then the jet
    \[
        \jet \rf 
        = (\rf, \nabla\rf, (\partial_{ij}\rf)_{i\le j}, \dots, (\partial_{i_1\cdots i_k} \rf)_{i_1\le \dots\le i_k})
    \]
    is strict positive definite for any \(k\in \nat\).
\end{restatable}

\begin{proof}[Proof of Theorem \ref{thm: strict positive definite derivatives}]
    For any \(n\) finite and distinct \(x_1,\dots,x_n\in\real^\dims\) we need that
    \[
        0=\Var \Bigl[\sum_{j=1}^n w_j^T \jet\rf(x_j)\Bigr]
    \]
    implies \(w_i=0\) for all \(i=1,\dots, n\). Without loss of generality we
    assume \(\rf\) (and thus \(\jet\rf\)) to be centered. We can then rewrite this as
    \[
        \Var \Bigl[\sum_{j=1}^n w_j^T \jet\rf(x_j)\Bigr]
        = \Var\Bigl[\sum_{j=1}^n \sum_{l=0}^k \sum_{i_1\le \dots \le i_l} w_j^{(l, i_1,\dots, i_l)}\partial_{i_1\dots i_l}\rf(x_j)\Bigr],
    \]
    with appropriate indexing of the \(w_j\). Note that \(l\) ranges over the
    order of differentiation contained in \(\rg\) before we sum over all the
    partial derivatives in the inner sum.
    For \(k=1\) this is for example
    \[
        \Var\Bigl[\sum_{j=1}^n w_j^{(0)}\rf(x_j) + \sum_{i=1}^\dims w_j^{(1,i)}\partial_i\rf(x_j)\Bigr]
        = \Var\Bigl[\sum_{j=1}^n w_j^{(0)}\rf(x_j) + D_{w_j^{(1,\cdot)}}\rf(x_j)\Bigr] = 0.
    \]
    We now consider the linear differential operator
    \[
        T := \sum_{j=1}^n \sum_{l=0}^k \sum_{i_1\le \dots \le i_l}
        (-1)^lw_j^{(l, i_1,\dots, i_l)}\partial_{i_1\dots i_l}\delta_{x_j},
    \]
    where \(\delta_x(f) = f(x)\) is the dirac delta function and \(\partial_i
    \delta_x(f) = - \partial_i f(x)\) is its derivative in the sense of distributions. 
    Recall this means for test functions \(\phi\), that we have
    \[
        \langle \partial_i\delta_x, \phi\rangle
        = -\langle\delta_x, \partial_i\phi\rangle
        = -\partial_i\phi(x).
    \]
    The higher order derivatives in the sense of distributions are similarly
    defined. Using this operator we now obtain more succinctly
    \[
        0 = \Var\Bigl[\sum_{j=1}^n \sum_{l=0}^k \sum_{i_1\le \dots \le i_l} w_j^{(l, i_1,\dots, i_l)}\partial_{i_1\dots i_l}\rf(x_j)\Bigr]
        = \Var[T\rf].
    \]
    While \(T\rf\) is well defined, we now want to move this operator outside. For this
    we want to use the bilinearity of the covariance. But the covariance has two
    inputs and \(T \C_{\rf}\) is then not well defined because the differential
    operator \(T\) might be applied to the first or second input of \(\C_{\rf}\).
    To avoid this issue, we define with some abuse of notation \(T^t f(t) := Tf\)
    such that we can write \(T^t\C_{\rf}(t, s)\) when we mean \(T \C_{\rf}(\cdot, s)\).
    Note that \(T\) is a linear combination of basis elements
    \(\partial_{i_1\dots i_l}\delta_x\). So by bilinearity of the covariance it
    is sufficient to check, that we can move these basis elements out of
    the covariance, i.e.
    \begin{align*}
        \Cov(\partial_{i_1\dots i_l}\delta_x\rf, \partial_{j_1\dots j_{l'}}\delta_y\rf)
        &= \Cov(\partial_{i_1\dots i_l}\rf(x), \partial_{j_1\dots j_{l'}}\rf(y))
        \\
        &= (\partial_{i_1\dots i_l}\delta_x)^t\; (\partial_{j_1\dots j_{l'}}\delta_y)^s \C_{\rf}(t,s).
    \end{align*}
    Since \(T\) is a linear combinations of these, we get by the bilinearity of
    the covariance
    \[
        0
        = \Var[T \rf]
        = T^x T^y \C_{\rf}(x,y).
    \]
    In the remainder of the proof we will essentially show, that this
    variance can be represented as an integral over the absolute value
    of the fourier transform of \(T\) with respect to the spectral measure
    of \(\C_{\rf}\). This forces the fourier transform of \(T\) and therefore
    \(T\) to be zero.

    Using the spectral representation of \(\C_{\rf}\) \citep[e.g.][Theorem
    1.7.4]{sasvariMultivariateCharacteristicCorrelation2013} given by
    \[
        \C_{\rf}(x,y)
        = \int e^{\im \langle x-y, t\rangle} \spectMeasure(dt),
    \]
    we move the operator into the integral (by moving sums and derivatives
    into the integral)
    \[
        0 = T^x T^y \C_{\rf}(x,y)
        = \int T^x e^{\im\langle x,t\rangle} T^y e^{-\im\langle y,t\rangle}\spectMeasure(dt)
        = \int |T e^{\im\langle \cdot, t\rangle}| \spectMeasure(dt),
    \]
    where we use \(T\bar{f} = \overline{T f}\) for the last equation, which follows from
    \begin{itemize}
        \item \(\delta_x(\overline{f}) =\overline{\delta_x(f)}\)
        \item \(D_v\delta_x(\overline{f}) = D_v\overline{f(x)}= \overline{D_v f(x)} =
        \overline{D_v\delta_x(f)}\) because from \(|z| = |\overline{z}|\) follows
        \[
            \lim_{t\to 0}\frac{|\overline{f(x+tv)} - \overline{f(x)} - \overline{D_vf(x)} t|}{t}
            = 0,
        \]
        \item induction for higher order derivatives.
    \end{itemize}
    So we have that \(P(t):=Te^{\im\langle \cdot, t\rangle}\) must be zero \(\spectMeasure\)-almost everywhere.
    Since
    \[
        P(t)
        = \sum_{j=1}^n \sum_{l=0}^k \sum_{i_1\le \dots \le i_l} w_j^{(l, i_1,\dots, i_l)}\partial_{i_1\dots i_l}e^{\im\langle x_j, t\rangle}
    \]
    is continuous in \(t\), it can only be zero \(\spectMeasure\)-almost everywhere if it is zero on
    the support of \(\spectMeasure\). Since the support of the spectral measure
    \(\spectMeasure\) contains an open subset by assumption of the theorem, \(P\)
    must be zero on this open subset. But then \(P\) must be zero everywhere as an
    analytic function. As \(P(t)\) is the fourier transform of \(T\) in the
    sense of distributions, i.e.
    \[
        P(t)
        = \int \Bigl(\sum_{j=1}^n \sum_{l=0}^k \sum_{i_1\le \dots \le i_l} w_j^{(l, i_1,\dots, i_l)}\partial_{i_1\dots i_l}\delta_{x_j}\Bigr)(x)
        e^{\im\langle x, t\rangle}dx
        = \Ft[T](t),
    \]
    this requires \(T\) to be zero by linearity and invertibility of the Fourier transform. But since
    the \(\partial_{i_1\dots i_l}\delta_{x_j}\) are linear independent\footnote{
        To see the linear independence of \(\partial_{i_1\dots
        i_l}\delta_{x_j}\), consider that the finite \(x_k\) are distinct, so
        they have a minimal distance. Rescale a bump
        function with zero slope at the top, e.g.
        \[
            \phi(x) = \begin{cases}
                \exp\Bigl(-\frac{1}{1-\|x\|^2}\Bigr) & \|x\| < 1
                \\
                0 & \|x\| \ge 1
            \end{cases}
        \]
        such that it is centered on some \(x_j\) and it is zero at all other \(x_k\)
        (and zero at all derivatives). This implies
        \[
            0 = \langle T, \phi\rangle = w_j^{(0)} \phi(x_j)
        \]
        and thus \(w_j^{(0)}=0\).
        Then construct similar test functions to ensure the other prefactors have to
        be zero, by placing a non-zero slope at \(x_j\) while ensuring it is zero at
        all other \(x_k\).
    }
    for distinct \(x_j\) the only way \(T\) can be zero is, if all the \(w_j\)
    are zero. This finishes the proof.
\end{proof}

The general requirement, the existence of a non-empty open sets in the support
of the spectral measure, is satisfied for the stationary isotropic random
functions valid in all dimensions (Corollary~\ref{cor: strict positive
definite}). To prove this result we require the following lemma.
It proves that stationary isotropic random function in \(\ell^2\)
is either almost surely constant or the `Schoenberg measure' has positive measure
on \((0,\infty)\). Where the Schoenberg measure \(\schoenbergMeas\) refers to 
the measure in Schoenberg's characterization of stationary isotropic covariance
kernels on \(\ell^2\) given by
\begin{equation}
    \label{eq: schoenberg charact}    
    \C_\rf(x,y) = \ikernel(\tfrac{\|x-y\|^2}2) = \int_{[0,\infty)} \exp(-t^2 \tfrac{\|x-y\|^2}2)\schoenbergMeas(dt)
\end{equation}

\begin{lemma}[Constant random functions]
    \label{lem: constant random functions}
    Let \(\ikernel\) be a stationary isotropic covariance kernel valid in all
    dimensions and let \(\rf\sim \normal(\mu, \ikernel)\) be a
    continuous stationary isotropic random function.

    If the Schoenberg measure \(\schoenbergMeas\) of \(\ikernel\) has no mass on
    \((0,\infty)\), i.e.  \(\schoenbergMeas((0,\infty))=0\), then \(\rf\)
    is almost surely constant.
\end{lemma}
\begin{proof}
    By Schoenberg's characterization \eqref{eq: schoenberg charact} we have
    \[
        \ikernel(r)
        = \int_{[0,\infty)} \exp(-t^2r)\schoenbergMeas(dt)
        \overset{\schoenbergMeas((0,\infty)=0)}= \schoenbergMeas(\{0\}).
    \]
    This implies a constant covariance
    \[
        \Cov(\rf(x), \rf(y)) = \schoenbergMeas(\{0\}).
    \]
    Assuming \(\rf\) to be centered (without loss of generality)
    by switching to \(\tilde{\rf}_\dims = \rf - \mu\), this implies
    \[
        \E[(\rf(x) - \rf(y))^2] = \E[\rf(x)^2] - 2\E[\rf(x)\rf(y)] + \E[\rf(y)^2] = 0.
    \]
    Thus \(\rf(x) = \rf(y)\) almost surely for all \(x,y\). Via the union over a
    dense countable subset of \(\real^\dims\) and \as continuity of \(\rf\) we get
    \[
        \Pr(\rf \text{ \as constant})
        = \Pr\Bigl(\rf(x) = \rf(y) \quad \forall x,y \in \real^\dims\Bigr)
        = 1.
    \]
\end{proof}

Using this lemma, we can prove that all stationary isotropic covariance kernels
on \(\ell^2\) have strictly positive definite derivatives. 

\corStrictPosDef*
\begin{proof}[Proof of Corollary \ref{cor: strict positive definite}]
    Since \(\charfct(x) = e^{-\frac{\|x\|^2t^2}2}\) is the characteristic function of
    \(Y_t\sim\normal(0, t\identity_{\dims\times\dims})\), Schoenbergs
    characterization \eqref{eq: schoenberg charact} of the covariance  implies up to scaling
    \begin{align*}
        \C_{\rf}(x,\tilde{x})
        &= \ikernel\bigl(\tfrac{\|x-\tilde{x}\|^2}2\bigr)
        = \int_{[0,\infty)} e^{-\frac{t^2\|x-\tilde{x}\|^2}2} \schoenbergMeas(dt)
        \\
        &= \int_{[0,\infty)} \E[e^{\im\langle x-\tilde{x}, Y_t\rangle}] \schoenbergMeas(dt)
        \\
        &= \int e^{\im\langle x-\tilde{x}, y\rangle}\underbrace{
            \int_{(0,\infty)}  e^{-\frac{\|y\|^2}{2t}} \schoenbergMeas(dt)
        }_{=:\density_\spectMeasure(y)} dy
        + \schoenbergMeas(\{0\})
        \\
        &= \int e^{\im\langle x-\tilde{x}, y\rangle} \spectMeasure(dy)
    \end{align*}
    with spectral measure
    \[
        \spectMeasure(A) :=\int_A \density_\spectMeasure(y) dy + \tfrac1\dims\schoenbergMeas\{0\} \delta_0(A).
    \]
    Since \(\schoenbergMeas((0,\infty))\neq 0\) because \(\rf\) is not almost surely
    constant (Corollary~\ref{lem: constant random functions}),
    its density \(\density_\spectMeasure\) is continuous and positive
    \(\density_\spectMeasure(y) >0\) for all \(y\in \real^\dims\) and thus
    \(\support(\spectMeasure)= \real^\dims\). In particular Theorem~\ref{thm:
    strict positive definite derivatives} is applicable and finishes the proof.
\end{proof}

\section{Technical results}

Recall that \(\kernel_3>0\) was used in \eqref{eq: LLN induction start} to prove
that the gradient has positive length. This fact follows from strict positive
definiteness, which we are now going to prove.

\begin{lemma}
    \label{lem: kappa_3 positive} 
    Let \(\kernel\) be a (non-stationary) isotropic covariance kernel valid in all
    dimensions.  Let \(\rf\sim \normal(\mu, \kernel)\) and assume \((\rf,
    \nabla\rf)\) to be strict positive definite, then for any
    \(\param\in \real^\dims\)
    \[
        \kernel_3\bigl(\tfrac{\|\param\|^2}2, \tfrac{\|\param\|^2}2, \|\param\|^2\bigr)
        > 0.
    \]
\end{lemma}
\begin{proof}
    Since the norm \(\|\cdot\|^2\) is rotation invariant, we can assume without
    loss of generality \(x = \lambda e_1\) for the standard basis vector \(e_1
    \in \real^\dims\). Since \(\partial_2 \rf\) is strict positive definite, we know that
    \[
        0< \Var(\partial_2\rf(\param))
        = \Cov(\partial_2 \rf(\lambda e_1), \partial_2\rf(\lambda e_1))
        = \tfrac1\dims \kernel_3\bigl(\tfrac{\|\param\|^2}2, \tfrac{\|\param\|^2}2, \|\param\|^2\bigr),
    \]
    where we have used \eqref{eq: cov df, df} of Lemma~\ref{lem: cov of derivatives, non-stationary isotropy}
    and \(\|\lambda e_1\| = \|\param\|\) in the last equation.
\end{proof}

In the stationary isotropic case, we can say more. We do not only have \(\kernel_3
= -\ikernel'(0)>0\), but we have \(\ikernel'(r) <0\) for all \(r\ge 0\).

\begin{lemma}
    \label{lem: covariance derivative negative}
    Let \(\ikernel\) be a stationary isotropic covariance kernel
    valid in all dimensions.
    If \(\rf\sim\normal(\mu, \ikernel)\) is \emph{not} almost surely
    constant, then \(\ikernel'(r)<0\) for all \(r\ge 0\).
\end{lemma}
\begin{proof}
    By Corollary~\ref{lem: constant random functions} we have
    \(\schoenbergMeas((0,\infty)) > 0\) for the Schoenberg measure
    \(\schoenbergMeas\) in Schoenbergs's characterization \eqref{eq: schoenberg
    charact}.  Thus by the continuity of measures there exists
    \(a,b>0\) such that \(\schoenbergMeas([a,b]) > 0\). Then by
    \eqref{eq: schoenberg charact} we have
    \begin{align*}
        -\ikernel'(r)
        &= \int_{[0,\infty)} t^2 \exp(-t^2r)\schoenbergMeas(dt)
        \\
        &\ge \int_{[a,b]} t^2 \exp(-t^2r)\schoenbergMeas(dt)
        \\
        &\ge \schoenbergMeas([a,b])\inf_{s\in [a,b]}s^2 \exp(-s^2r) > 0.
    \end{align*}
\end{proof}
 
\end{document}